\documentclass[nohyperref]{article}
\usepackage{amsmath,amsfonts,bm}

\def\eqref#1{equation~\ref{#1}}
\def\1{\bm{1}}

\def\vx{{\bm{x}}}

\DeclareMathAlphabet{\mathsfit}{\encodingdefault}{\sfdefault}{m}{sl}
\SetMathAlphabet{\mathsfit}{bold}{\encodingdefault}{\sfdefault}{bx}{n}

\def\sR{{\mathbb{R}}}

\DeclareMathOperator*{\argmax}{arg\,max}

\DeclareMathOperator{\sign}{sign}

\usepackage{microtype}
\usepackage{graphicx}
\usepackage{subfigure}
\usepackage{booktabs} % for professional tables

\usepackage{hyperref}

\usepackage[accepted]{icml2022}
\usepackage{amsmath}
\usepackage{amssymb}
\usepackage{mathtools}
\usepackage{amsthm}

\usepackage[capitalize,noabbrev]{cleveref}

\newcommand{\httpsurl}[1]{\href{https://#1}{\nolinkurl{#1}}}
\usepackage[textsize=tiny]{todonotes}

\usepackage{microtype}
\usepackage{graphicx}
\usepackage{subfigure}
\usepackage{booktabs}
\usepackage{amsthm}

\usepackage{amsthm}
\newtheorem{mydef}{Condition}
\newtheorem{mydefinition}{Definition}

\usepackage{url}

\usepackage{algorithm}
\usepackage{algorithmic}
\usepackage{wrapfig}
\usepackage{multirow}
\usepackage{makecell}
\usepackage{tabu}
\usepackage{colortbl}
\usepackage{array}
\usepackage{xcolor}
\usepackage{soul} 
\definecolor{mydarkred}{rgb}{0.6,0,0}
\definecolor{mydarkgreen}{rgb}{0,0.6,0}
\definecolor{mydarkblue}{rgb}{0,0,0.6}

\icmltitlerunning{Fast and Reliable Evaluation of Adversarial Robustness with Minimum-Margin Attack}

\begin{document}

\twocolumn[
\icmltitle{Fast and Reliable Evaluation of Adversarial Robustness with \\ Minimum-Margin Attack}

% It is OKAY to include author information, even for blind
% submissions: the style file will automatically remove it for you
% unless you've provided the [accepted] option to the icml2022
% package.

% List of affiliations: The first argument should be a (short)
% identifier you will use later to specify author affiliations
% Academic affiliations should list Department, University, City, Region, Country
% Industry affiliations should list Company, City, Region, Country

% You can specify symbols, otherwise they are numbered in order.
% Ideally, you should not use this facility. Affiliations will be numbered
% in order of appearance and this is the preferred way.
\icmlsetsymbol{equal}{*}

\begin{icmlauthorlist}
\icmlauthor{Ruize Gao}{cuhk}
\icmlauthor{Jiongxiao Wang}{cuhk}
\icmlauthor{Kaiwen Zhou}{cuhk}
\icmlauthor{Feng Liu}{Mel}
\icmlauthor{Binghui Xie}{cuhk}
\icmlauthor{Gang Niu}{riken}
\icmlauthor{Bo Han}{bu}
\icmlauthor{James Cheng}{cuhk}
\end{icmlauthorlist}

\icmlaffiliation{cuhk}{Department of Computer Science and Engineering, The Chinese University of Hong Kong}
%\icmlaffiliation{fudan}{School of Mathematical Sciences, Fudan University}
\icmlaffiliation{bu}{Department of Computer Science, Hong Kong Baptist University}
%\icmlaffiliation{uts}{DeSI Lab, AAII, University of Technology Sydney}
\icmlaffiliation{Mel}{School of Mathematics and Statistics, The University of Melbourne}
\icmlaffiliation{riken}{RIKEN-AIP}

\icmlcorrespondingauthor{James Cheng}{jcheng@cse.cuhk.edu.hk}
\icmlcorrespondingauthor{Bo Han}{bhanml@comp.hkbu.edu.hk}

% You may provide any keywords that you
% find helpful for describing your paper; these are used to populate
% the "keywords" metadata in the PDF but will not be shown in the document
\icmlkeywords{Machine Learning, ICML}

\vskip 0.3in
]

% this must go after the closing bracket ] following \twocolumn[ ...

% This command actually creates the footnote in the first column
% listing the affiliations and the copyright notice.
% The command takes one argument, which is text to display at the start of the footnote.
% The \icmlEqualContribution command is standard text for equal contribution.
% Remove it (just {}) if you do not need this facility.

%\printAffiliationsAndNotice{}  % leave blank if no need to mention equal contribution
\printAffiliationsAndNotice{} % otherwise use the standard text.

\newcommand{\fix}{\marginpar{FIX}}
\newcommand{\new}{\marginpar{NEW}}

\begin{abstract}
The \emph{AutoAttack} (AA) has been the most reliable method to evaluate adversarial robustness when \emph{considerable} computational resources are available.
However, the high computational cost (e.g., \emph{100 times} more than that of the \emph{project gradient descent} (PGD-20) attack) makes AA \emph{infeasible} for practitioners with limited computational resources, and also hinders applications of AA in the \emph{adversarial training} (AT). In this paper, we propose a novel method, \emph{minimum-margin}~(MM) attack, to fast and reliably evaluate adversarial robustness. Compared with AA, our method achieves comparable performance but \emph{only costs 3\%} of the computational time in extensive experiments.
The reliability of our method lies in that we evaluate the quality of adversarial examples using the margin between two targets that can precisely identify the most adversarial example.
The computational efficiency of our method lies in an effective \emph{Sequential TArget Ranking Selection} (STARS) method, ensuring that the cost of the MM attack is independent of the number of classes. 
As a better benchmark, the MM attack opens a new way for evaluating adversarial robustness and provides a feasible and reliable way to generate high-quality adversarial examples in AT.
\end{abstract}

\begin{figure*}[tp]
    \begin{center}
        \subfigure[Reliability ranking]
        {\includegraphics[width=0.495\textwidth]{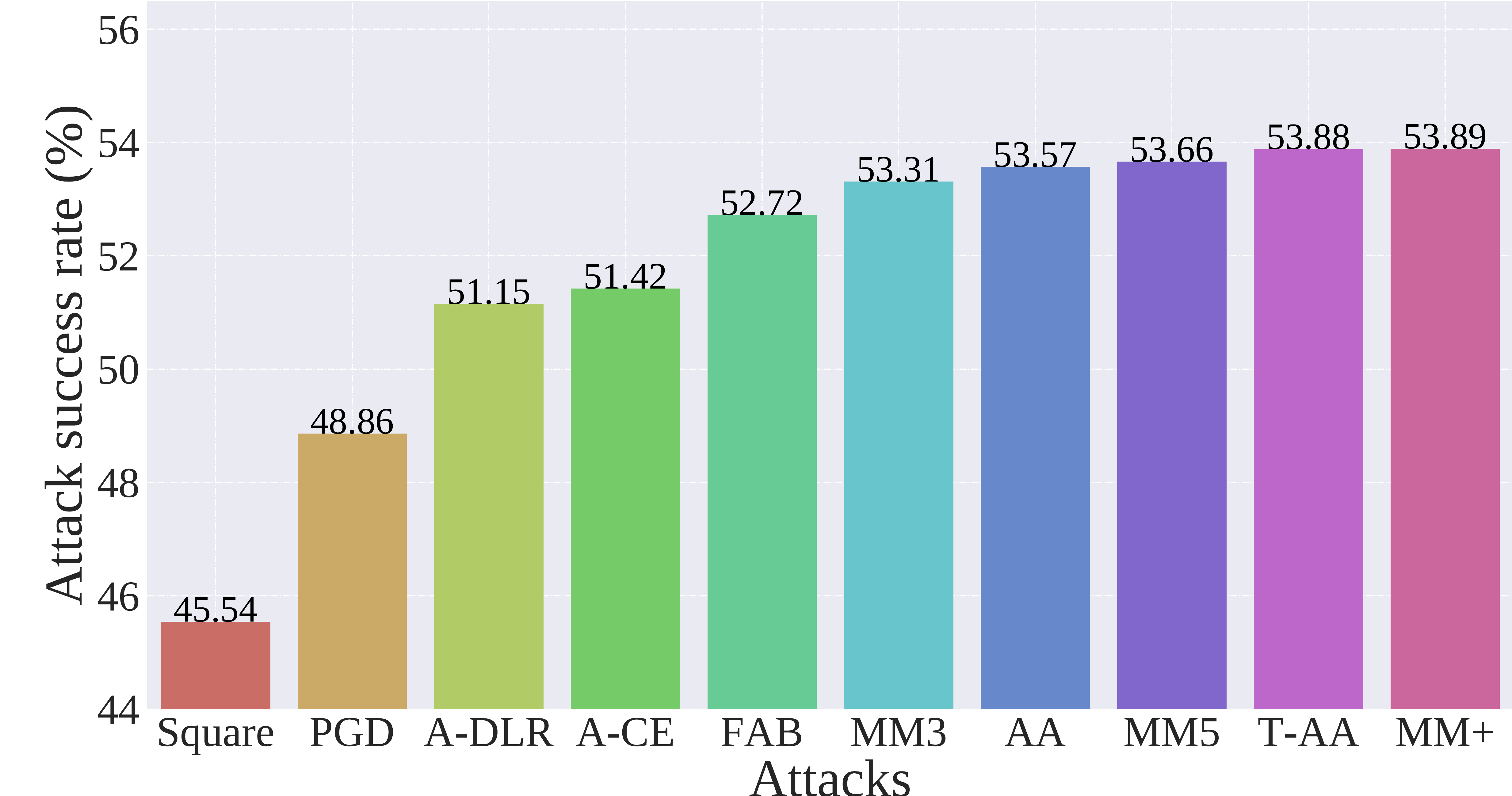}}
        \subfigure[Computational time]
        {\includegraphics[width=0.495\textwidth]{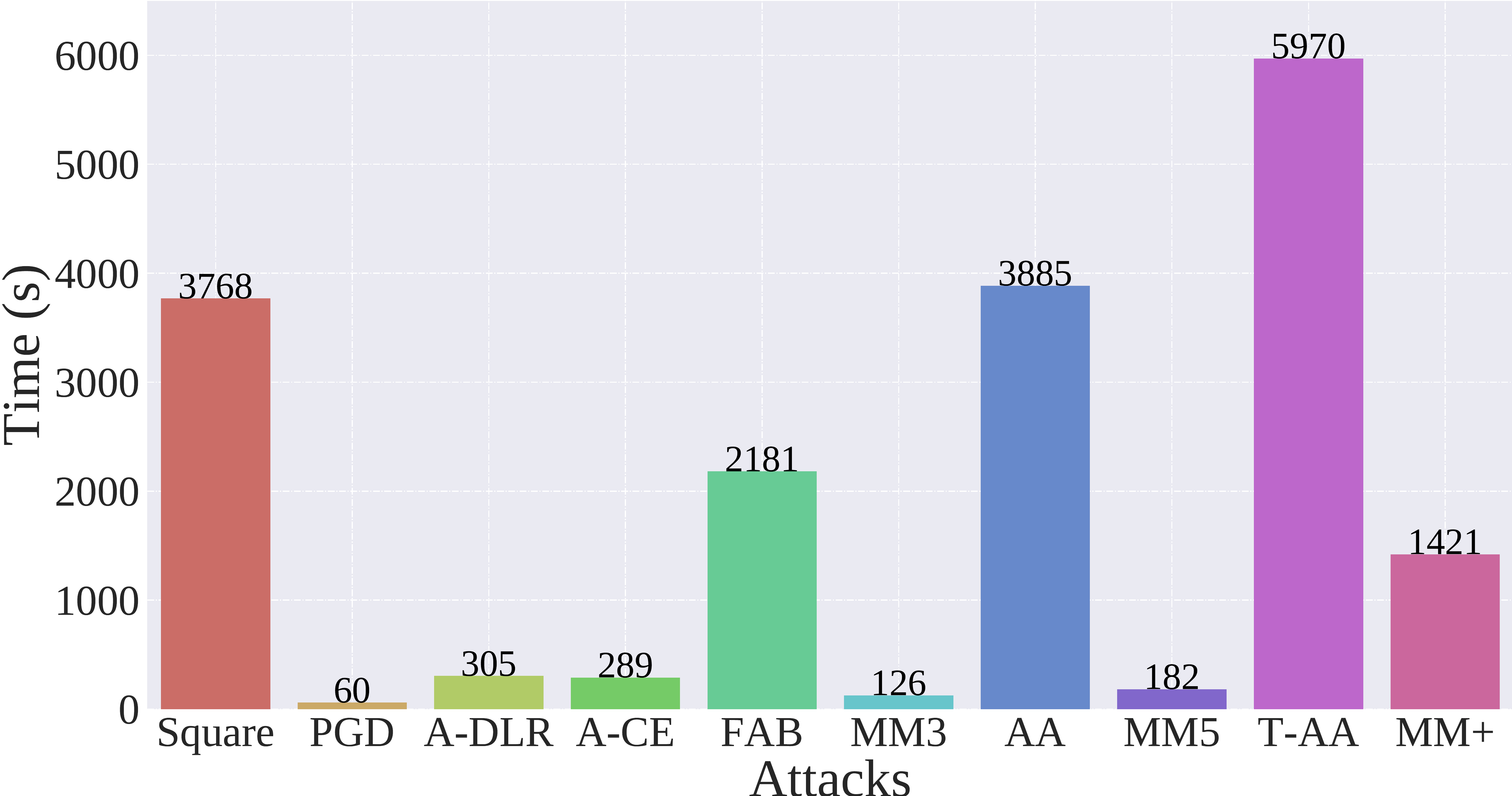}}
        \vspace{-1em}
        \caption{\footnotesize Comparison of reliability and computational cost among baselines and different versions of MM attack. MM3, MM5 and MM+ are three versions of our MM attack. The subfigure (a) shows the (sorted) attack success rates of different attacks. The higher the success rate, the stronger the attack. The subfigure (b) shows the computational cost of these attacks. The less time, the better the attack. AA is an ensemble of APGD-CE (A-CE for short), APGD-DLR (A-DLR for short), FAB and Square. T-AA considers each target for APGD-DLR and FAB in AA, and is thus more time-consuming. Compared with AA (or T-AA), our MM3 achieves comparable performance but only costs 3\% (or 2\%) of the computational time. The model structure we used is ResNet-18, which follows the adversarial training of \citep{Madry18PGD} on CIFAR-10.
        %In the subfigure (b), the gray shape is a hypothetical distribution of all adversarial variants maps within the bounded perturbation epsilon on a natural example; $\mathbb{P}_t$ and $\mathbb{P}_y$ are the predicted probability on a targeted false label $t$ and the true label $y$; The orange area ($\mathbb{P}_t > \mathbb{P}_y$) indicates that the adversarial variants inside can be misclassified, or to say attack successfully, while the blue area ($\mathbb{P}_t < \mathbb{P}_y$) indicates that the adversarial variants inside cannot attack successfully. 
        }
        \vspace{-1em}
    \label{fig:moti}
    \end{center}
    \vspace{-1em}
\end{figure*}

\newcommand{\cX}{\mathcal{X}}
\newcommand{\epsball}{\mathcal{B}_\epsilon}

\section{Introduction}
The \emph{deep neural network} (DNN) has attracted a large number of researchers from different disciplines such as computer science
\citep{goodfellow2016deep,castelvecchi2016deep,vaswani2017attention}, physics \citep{devries2018deep,huang2019applications,levine2019quantum}, biology \citep{maxmen2018deep,maxmen2018machine,webb2018deep} and medicine \citep{hao2015forniceal,esteva2017dermatologist}. The success of DNN mainly lies in its ability to learn useful high-level features from abundant data \citep{deng2014deep,lecun2015deep}. These learned features have been successfully used to address many difficult tasks. For example, DNNs can recognize images with high accuracy comparable to human beings \citep{lecun1998gradient,krizhevsky2012imagenet}. In addition, DNNs are also widely used for speech recognition \citep{hinton2012deep}, natural language processing \citep{andor2016globally}, and playing games \citep{mnih2013playing,silver2016mastering}. 

As the impacts of DNN increase fast, its reliability has been a key to deploy it in real-world applications \citep{huang2011adversarial,kurakin2016adversarial}. Recently, a growing body of research shows that DNNs are vulnerable to adversarial examples, i.e., test inputs that are modified slightly yet strategically to cause misclassification \citep{szegedy2013intriguing,nguyen2015deep,kurakin2016adversarial,carlini2017adversarial,finlayson2019adversarial,wang2019convergence,zhang2020dual,zhang2020principal,gao2020maximum,zhang2021adversarial}. The existence of such adversarial examples lowers the reliability of DNNs. Meanwhile, researchers have also been considering finding a reliable way to evaluate adversarial robustness of a DNN before deploying it in the real world. 

The high-level idea of evaluating adversarial robustness of a DNN is quite straightforward, i.e., generating adversarial examples and calculating the accuracy of the DNN on these examples (this kind of accuracy is also known as \emph{adversarial robust accuracy}). 
\citet{szegedy2013intriguing} first pointed out the existence of adversarial examples and used a less powerful box-constrained \emph{limited-memory Broyden-Fletcher-Goldfarb-Shanno} (L-BFGS) method to generate them. Based on the studies in \citep{szegedy2013intriguing}, \citet{goodfellow2014explaining} put forward the \emph{fast gradient sign method}~(FGSM). One common loss function they used is \emph{cross-entropy}~(CE) loss, and to maximize the loss function, FGSM uses its gradient to determine in which direction the pixel's intensity should be increased or decreased. \citet{Madry18PGD} introduced a simple refinement of the FGSM: \emph{projected gradient descent attack}~(PGD), where instead of taking a single step of size $\epsilon$ in the direction of the gradient sign, multiple smaller steps are taken. 

\textbf{Existing Evaluation Methods.} PGD was an effective method to evaluate adversarial robustness of a \emph{standard-trained DNN} \citep{Madry18PGD} since adversarial robust accuracy of a standard-trained DNN is always very low after using PGD. Nevertheless, the existence of adversarial examples has already inspired research on training a \emph{robust DNN} to defend against them, which means that a standard-trained DNN is not the only DNN we might meet and we need to evaluate adversarial robustness of a \emph{robust DNN} as well. Unfortunately, as observed by \citep{carlini2017towards,croce2020reliable}, PGD has limitations to reliably evaluate adversarial robustness of a robust DNN.

\citet{carlini2017towards} observed the phenomenon of gradient vanishing in the widely used CE loss for the potential failure of L-BFGS, FGSM and PGD, and replaced the CE loss with many possible choices. \citet{croce2020reliable} claimed that the fixed step size and the single attack used are the causes of poor evaluations, and they put forward an ensemble of diverse attacks (consisting of APGD-CE, APGD-DLR, FAB and Square) called \emph{AutoAttack}~(AA) to test adversarial robustness. Until now, AA \citep{croce2020reliable} has been regarded as the most reliable method and is widely used in the evaluation of adversarial robustness \citep{sehwag2021improving, rade2021helper, rebuffi2021fixing, andriushchenko2020understanding, gowal2020uncovering, sridhar2021robust, wong2020fast, robustness, carmon2019unlabeled, wang2019improving, wu2020adversarial, zhang2020geometry}.

However, though AA performs well in reliability, it needs a large amount of computational time. As shown in Figure~\ref{fig:moti}(b), for evaluating adversarial robustness of a ResNet-18 model on CIFAR-10 (following the adversarial training in \citep{Madry18PGD}), the computational cost of AA (or T-AA) is 65 times (or 100 times) more than PGD-20 used in \citep{Madry18PGD}, where T-AA is more time-consuming since it considers each target for APGD-DLR and FAB in AA. Worse still, in the worst case as analyzed in Appendix~\ref{app_bound}, the computational cost of AA (or T-AA) is even 109 times (or 440 times) more than PGD-20. 

\begin{figure*}[!t]
    \begin{center}
        {\includegraphics[width=0.65\textwidth]{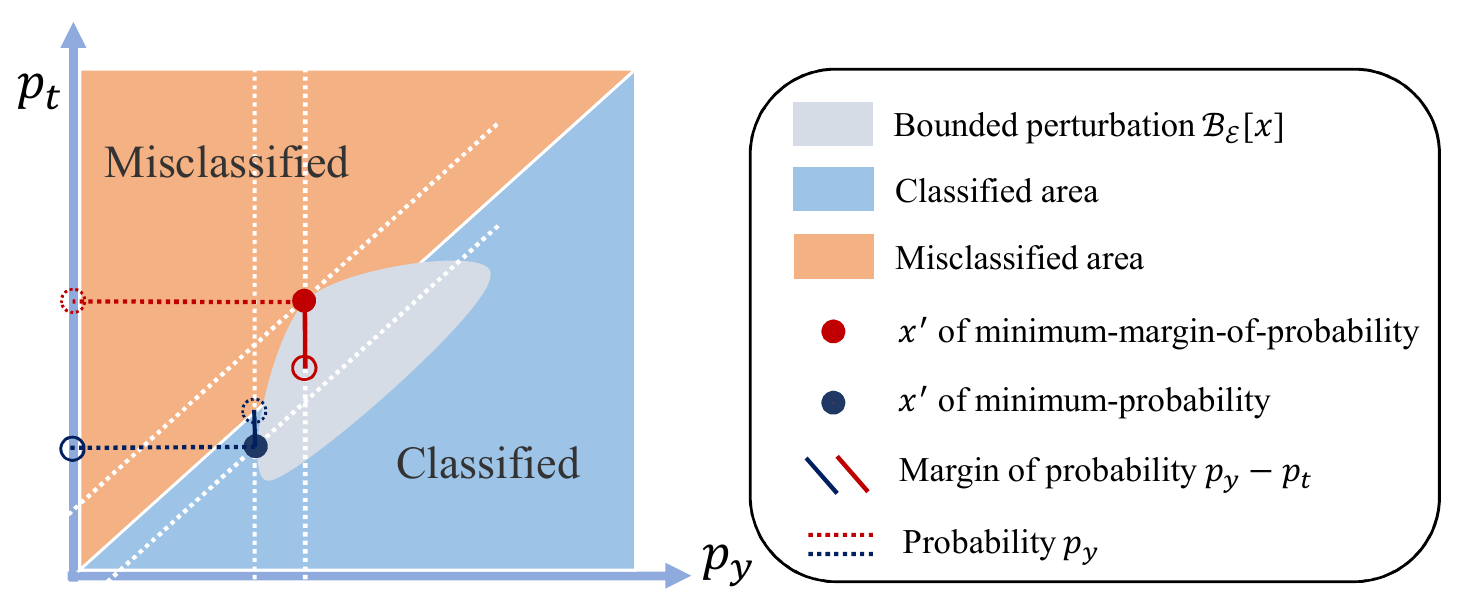}}
        \vspace{-1em}
        \caption{\footnotesize Minimum margin of probability. $p$ denotes the predicted probability, $p_y$ and $p_t$ are the predicted probability on the true label $y$ and a targeted false label $t$. %We discuss other rescaling functions in Section~\ref{sec:Rescaling}.
        The gray shape is the image of the adversarial variants $x^{\prime}$ within the bounded perturbation ball $\epsball[x]$ under the mapping of the network onto $(p_y, p_t)$; the orange area ($p_t > p_y$) indicates the region where the adversarial variants are misclassified, or to say a successful attack, while the blue area ($p_t < p_y$) indicates the region where the adversarial variants do not attack successfully.
        }
        \vspace{-1em}
    \label{fig:MMP}
    \end{center}
    \vspace{-1em}
\end{figure*}

\textbf{A Dilemma Between Reliability and Computational Efficiency.} The high computational cost makes AA infeasible when considerable computational resources are unavailable. Unfortunately, such scenarios are common in the real world, e.g., as recommended by \citep{rice2020overfitting}, practitioners need real-time evaluation at each epoch of the adversarial training process to find the robust model with “the best checkpoint”, and in this case, such high computational cost is unacceptable. Similarly, since a large number of adversarial examples need to be generated at each epoch 
during \emph{adversarial training} (AT), such high computational cost hinders applications of AA in AT. In consideration of the high reliability but low computational efficiency of AA, and the high computational efficiency but low reliability of PGD, we seem to encounter a \emph{dilemma}: we \emph{have to} give up one factor (reliability or computational efficiency) when evaluating the adversarial robustness.  \citep{croce2020reliable}.

\textbf{Our Reliable and Fast Solution.} In this paper, we aim to achieve reliability and computational efficiency simultaneously. For reliability, we evaluate the quality of adversarial examples using the margin between two targets for precisely identifying the most adversarial example. For computational efficiency, we propose an effective \emph{Sequential TArget Ranking Selection} (STARS) method to ensure that the cost of the MM attack is independent of the number of classes. 
%The reliability of our method lies in that we evaluate the quality of adversarial examples using the margin between two targets that can precisely identify the most adversarial example. The computational efficiency of our method lies in an effective \emph{Sequential TArget Ranking Selection} (STARS) method, ensuring that the cost of the MM attack is independent of the number of classes. 

\textbf{Reliability.} 
To achieve reliability, we investigate the reasons behind the failure of PGD. We identify that CE loss, which is based on the probability of the true label $p_y$, is not an appropriate measure to the quality of adversarial examples. In Figure \ref{fig:MMP}, we provide a simple demonstration to this issue, in which we consider one targeted false label $t$. As we can see, it is much more reasonable to measure the quality of adversarial examples in terms of the \emph{margin of probability} $p_y-p_t$. The most adversarial example in Figure~\ref{fig:MMP} then corresponds to the one with the \emph{minimum margin of probability} instead of the \emph{minimum probability} $p_y$. Detailed study of the rationality of minimum-magrin is provided in Section~\ref{sec:Ration}. Since the search space $\epsball$ of high dimensional images is large (grey area), previous studies use gradient descent methods to generate the adversarial example that maximizes the loss function \citep{goodfellow2014explaining,carlini2017towards,Madry18PGD}.

Though it looks promising to generate adversarial examples via minimizing the margin of probability, we find that there are still two issues: (a) The probability $p$ is a kind of rescaling method to the logits $z$. \citet{croce2020reliable} heuristically rescaled the logits $z$ using their proposed \emph{difference of logits ratio} (DLR) (defined at Eq.(\ref{Eq_DLR})). We investigate the performance of different rescaling methods in Section~\ref{sec:Ration}. We numerically find that the method using natural logits ${z}_y-{z}_t$ (the meaning of margin) with no rescaling performs the best; (b) For the problem of multi-class and untargeted attacks, the \emph{margin} is then ${z}_y-\max_{i\neq y}{z}_i$. However, $-({z}_y-\max_{i\neq y}{z}_i)$ is not an appropriate loss function, because the $\max$ function only considers one target at the current step while those unconsidered targets may lead to more adversarial examples. Hence, the reliable method is to minimize ${z}_y-{z}_t$ for each $t \neq y$ and take the most adversarial one \citep{croce2020reliable}.

\textbf{Computational Efficiency.} 
Although running the attack for each false target is reliable, the computational cost depends on the number of classes. For datasets with a large number of classes, e.g., CIFAR-100 (100 classes) and Imagenet (1,000 classes), the computational cost will increase accordingly. To achieve computational efficiency, we propose a STARS method to make the computational time independent of the number of classes. STARS consists of two strategies that pre-selecting targets and ranking sequential attack. 
% According to the ranking of predicted probability, 
STARS method only selects a few highest targets and runs a sequential attack. Experiments show that, benefited from STARS, computational time can be saved 76.36\% on CIFAR-10, 98.51\% on CIFAR-100 and 77.78\% on SVHN. 

%MM attack can save 76.36\% of the computational time on CIFAR-10, 98.51\% on CIFAR-100 and 77.78\% on SVHN. 

%We find that, for a natural example, a target class with larger predicted probability has a greater chance of conducting a successful attack. This finding motivates us to perform the following sequential attack based on the ranking of the predicted probability: we first consider the false target with the largest predicted probability. If the attack succeeds, we stop other attacks on other targets; otherwise,  we continue to consider the next target in the ranking. Experiments in Section \ref{sec:STP} show that using only few selected targets is enough to achieve comparable results compared with selecting all the targets.

%预先选择的目标顺序攻击，激励我们少数目标足够重要，节省时间。

By taking all the above factors into consideration, we propose a novel method, the \textit{minimum-margin}~(MM) attack. Its detailed realization is provided in Section~\ref{sec:realization}. We present extensive experimental results in Section \ref{sec:exp}, which verify that our MM attack can fast and reliably evaluate adversarial robustness. In particular, MM attack achieves comparable performance but \emph{only costs 3\%} of the computational time compared with the current benchmark AA.

The main contributions of our work are as follows: our conceptual contribution lies in using margin to identify the “most adversarial example”; our technical contribution on reliability lies in using adaptive step size and searching the “most adversarial example” with minimum margin; third, our technical contribution on computational efficiency lies in using STARS method to achieve the targets pre-selected and ranking sequential strategy. Furthermore, as a better benchmark compared with AA, our proposed MM attack provides a new direction of evaluating adversarial robustness and contributes a feasible and reliable method to generate high-quality adversarial examples in adversarial training.

\section{Preliminary}

\textbf{Neural Networks.}
A neural network is a function $f_\theta:\sR^n\rightarrow [0,1]^K$, where $\theta$ is the parameters contained in $f_\theta$ and $K$ is normally the number of classes. The output of the network is computed using the softmax function, which ensures that the output is a valid probability vector. Namely, given an input $x\in\sR^n$, $f_\theta(x)=[p_1,\dots,p_K]=p$ where $\sum_{i=1}^K p_i = 1$ and $p_i$ is the probability that input $x$ belongs to class $i$. Before the softmax function, the output of the network is called logits $z$, i.e., $p=\textnormal{softmax}(z)$. The classifier assigns the label $y=\argmax_{i}{f(x)}_i$.

\textbf{Projected Gradient Descent Attack (PGD).}
\citet{Madry18PGD} proposed the \emph{projected gradient descent} (PGD) attack to generate adversarial examples to mislead a well-trained classifier $f_\theta$. Specifically, they start with setting $x_{(0)}=x$, and then in each iteration:
% \citet{Madry18PGD} introduced a simple refinement of FGSM, which is the \emph{projected gradient descent} (PGD) attack. Instead of taking a single step of size $\epsilon$ in the direction of the gradient sign, multiple smaller steps are taken in PGD (the result is clipped by the same $\epsilon$). Specifically, we start with setting $x^{0}=x$, and then in each iteration:
\begin{align*}
x^{\prime}_{(t+1)} = \Pi_{\mathcal{B}_\epsilon[x_{(0)}]}(x'_{(t)}+\alpha \sign(\nabla_{x'_{(t)}}\ell(f_{\theta}(x'_{(t)}),y)), 
% \label{Equ_PGD}
\end{align*}
$t = 0, 1, 2,\ldots$, where 
\begin{align*}
% \label{eqn:perturbation_ball}
\epsball[x] = \{x^{\prime} \mid d_{\infty}(x,x')\le\epsilon\},
\end{align*}
is the closed ball of radius $\epsilon>0$ centered at $x$; the  $x_{(0)}$ refers to the starting point which corresponds to the natural example $x$ (or the natural example perturbed by a small Gaussian or uniformly random noise that $x+\delta$); $x^{(t)}$ is the adversarial example at step $t$; $\Pi_{\mathcal{B}_\epsilon[x^{(0)}]}(\cdot)$ is the projection function that projects the adversarial variant back to the $\epsilon$-ball centered at $x^{(0)}$ if necessary; the $L_{\infty}$ distance metric is $d_{\infty}(x,x^{\prime})=\|x-x^{\prime}\|_\infty$; and $\ell$ is \emph{cross entropy}~(CE) loss:
\begin{align}
\label{Equ_CEloss}
\textnormal{CE}(x, y) = -\log(p_y) = -z_y+\log{\left(\sum_{j=1}^{K}e^{z_j}\right)},
\end{align}
where $p_i=e^{z_i}/\sum_{j=1}^{K}e^{z_j}, i=1,...,K$, and $z$ is the logits of the model outputs. 
% In this paper, we focus on the $\ell_{\infty}$ distance metric. 
% PGD is found to produce superior results than FGSM and used to be the most popular untargeted attack method. 

\textbf{Carlini and Wagner attack (CW).}
\citet{carlini2017towards} observed the phenomenon of gradient vanishing in the widely used CE loss for potential failure. The gradient w.r.t $x$ in Eq.~(\ref{Equ_CEloss}) is given by
\begin{align}
\label{Equ_CEgrad}
\nabla_{x}\textnormal{CE}(x, y) = (-1+p_y)\nabla_{x}z_y+\sum_{i\neq j}p_i\nabla_{x}z_i.
\end{align}
If $p_y \approx 1$ and consequently $p_i \approx 0$ for $i \neq y$, then $\nabla_{x}\textnormal{CE}(x, y) \approx 0$ in Eq.~(\ref{Equ_CEgrad}). This gradient vanishing issue can lead to the failure of attacks. Motivated by this phenomenon, \citet{carlini2017towards} replaced the CE loss with several possible choices. Among these choices, the widely used one for the untargeted attack is 
\begin{align}
\label{Equ_CWloss}
\textnormal{CW}(x, y) = -z_y(x^{\prime})+\max_{i\neq y}z_i(x^{\prime}).
\end{align}

\textbf{AutoAttack (AA).} 
\citet{croce2020reliable} claimed that the fixed step size and the lack of diversity in attack methods are the main reasons for the limitations of previous studies. Motivated by the \emph{line search} method \citep{grippo1986nonmonotone}, they put forward \emph{auto PGD} (APGD) attack. They showed that using adaptive step size significantly improves the adversarial evaluation compared with using fixed step size. For the loss function at Eq.~(\ref{Equ_CWloss}), they claim that scale invariance w.r.t. $z$ is necessary, and they proposed an alternative loss:
\begin{align}
\label{Eq_DLR}
\textnormal{DLR}(x,y) = -\frac{z_y(x^{\prime})-\max_{i\neq y}z_i(x^{\prime})}{z_{\pi_1}(x^{\prime})-z_{\pi_3}(x^{\prime})},
\end{align}
where $\pi$ is the permutation of the components of $z$ in decreasing order. For inducing the misclassification into a target class $t$, they propose another alternative loss function:
\begin{align}
\label{Eq-T-DLR}
\textnormal{Targeted-DLR}(x,y) = - \frac{z_y(x^{\prime}) - z_t(x^{\prime})}{z_{\pi_1}(x^{\prime})-\frac{1}{2}\cdot(z_{\pi_3}(x^{\prime})+z_{\pi_4}(x^{\prime}))}.
\end{align}
For the lack of diversity, they claimed that diverse attacks are beneficial for reliability, and then they put forward an ensemble of various parameter-free attacks called \emph{AutoAttack} (AA) to test adversarial robustness, where AA contains APGD-CE, APGD-DLR, FAB and Square. \emph{Targeted AutoAttack} (T-AA) replaces APGD-DLR with the targeted APGD-DLR, and replaces FAB with targeted FAB.

%As we mentioned before, although AA is currently the most reliable method to evaluate adversarial robustness, it requires a large amount of computational time.
\begin{table*}[!t]
\setlength{\tabcolsep}{3.5mm}
\scriptsize
\renewcommand\arraystretch{1.5}
\centering
\caption{Attack success rate (\%) of different loss functions. Following \citep{Madry18PGD}, the attack setting is PGD-20.}
\label{table:exp_dataset1}
\begin{tabular}{c|c|c c|c c|c c|c c}
\toprule[1.5pt]
Attack & Loss function & CIFAR-10 & Diff. & CIFAR-100 & Diff. & SVHN & Diff. & Tiny-Imagenet & Diff. \\
\midrule[0.6pt]
\midrule[0.6pt]
PGD & $-\log(p_y)$ & 48.27 & -3.09 & 73.60 & -2.92  & 41.89 & -5.67  & 78.78 & -4.19\\
\midrule    
CW & $-z_y{(x')} + \max_{i \neq y}{z_i{(x')}}$ & 49.13 & -2.23 & 74.55 & -1.97  & 45.08 & -2.48  & 81.19 & -1.78\\
\midrule    
MM & $-{z}_y+{z}_t$ & 51.36 & 0.00 & 76.52 & 0.00  & 47.56 & 0.00  & 82.97 & 0.00\\
\bottomrule[1.5pt]
\end{tabular}
\vskip1ex%
\vskip -0ex%
\vspace{-1em}
\end{table*}

\section{The Realization of MM Attack}
\label{sec:realization}
In this section, we discuss the realization of our MM Attack. In Section~\ref{sec:Ration}, we verify the rationality of using \emph{minimum margin} as the loss function and discuss the influence of different logits rescaling methods on robustness. In Section~\ref{sec:STP}, we propose an effective STARS method to improve computational efficiency. In Section~\ref{sec:alg_MM}, we provide the detailed descriptions of MM attack.

\subsection{The Rationality of Minimum Margin}

\label{sec:Ration}
% \textbf{The comparison with PGD and CW.} 
%To explain the rationality of minimum-margin,
%We define the robustness of the classifier $f$ to the example $x$.

To understand the rationality of minimum margin, we first look into the situation where no adversarial attack can succeed. Then we show that the formulation of \emph{minimum margin} is naturally derived from such a situation. We say that a classifier $f$ is \emph{completely robust} if ${\forall} x^{\prime} \in \epsball[x]$, $\argmax_{i}{f(x')}_i = \argmax_{i}{f(x)}_i$. The following condition is necessary and sufficient to the complete robustness:
\begin{mydef}%[The complete robustness of the classifier $f$ to the example $x$]
Given a natural example $x$ with its true label $y$, the $K$-class classifier $f$ satisfies
\begin{align}
\label{rob:def}
{\forall} x^{\prime} \in \epsball[x], z_y{(x')} - \max_{i \neq y}{z_i{(x')}} \geq 0,
\end{align}
where $\epsball[x] = \{x^{\prime} \mid d_{\infty}(x,x')\le\epsilon\}$; $z_y{(x')} = {f(x')}_y$; $z_i{(x')} = {f(x')}_i$.
\end{mydef}

According to this condition, to reliably evaluate the complete robustness, the adversarial attacks should find the adversarial examples with minimum $z_y{(x')} - \max_{i \neq y}{z_i{(x')}}$, i.e., the most non-robust data point. Hence, we define the most adversarial example:

\begin{mydefinition}[The most adversarial example]%[The complete robustness of the classifier $f$ to the example $x$]
Given a natural example $x$ with its true label $y$, the most adversarial example $x^\ast$ within $\epsball[x]$ is defined as:
\begin{align}
\label{mostadv:def}
{\forall} x^{\prime} \in \epsball[x],
x^\ast = \argmax_{x^{\prime}}{-(z_y{(x')} - \max_{i \neq y}{z_i{(x')}})},
\end{align}
where $\epsball[x] = \{x^{\prime} \mid d_{\infty}(x,x')\le\epsilon\}$ is the closed ball of radius $\epsilon>0$ centered at $x$; $z_y{(x')} = {f(x')}_y$; $z_i{(x')} = {f(x')}_i$.

\end{mydefinition}

Equation~\ref{mostadv:def} indicates that we should replace the CE loss in Eq. ({\ref{Equ_CEloss}}) with $-(z_y{(x')} - \max_{i \neq y}{z_i{(x')}})$ as the loss function in adversarial attacks.
%However, such reliable evaluation should consider any $x^{\prime}$ in $\epsball[x]$. As the search space $\epsball$ of high dimensional images is so large, it is impractical to count each one in $\epsball[x]$. 
However, as mentioned before, $-({z}_y-\max_{i\neq y}{z}_i)$ in Eq.~(\ref{Equ_CWloss}) is not an appropriate loss function since it only focuses on the current step. The reliable method is to minimize ${z}_y-{z}_t$ for each target $t \neq y$ and take the most adversarial one. To verify the rationality of minimum margin, we conduct experiments on different datasets with $-({z}_y-{z}_t)$ being the loss function (MM). In Table~\ref{table:exp_dataset1}, MM performs a more reliable evaluation 
than PGD and CW. Although \citet{gowal2019alternative} proposed such a surrogate multi-target loss on the basis of PGD, due to the limitations of PGD, they found that in some examples, the attack's performance is worse than CW. The relatively poor performance \citep{gowal2019alternative} failed to attract attention of later researchers. Nowadays, AutoAttack is widely regarded as the most authoritative evaluation of adversarial robustness. In the current mainstream of the field, researchers are recommended to use DLR loss to achieve better attack success rate and to use an ensemble of various attacks.

%as the search space $\epsball$ of high dimensional images is so large (e.g., of CIFAR-10, a natural example of $3072$ pixels has $2^{3072}$ adversarial variants just on the boundary of $\epsball[x]$), it is impractical to count each one in $\epsball[x]$. Adversarial attacks \citep{goodfellow2014explaining,Madry18PGD,carlini2017towards} for evaluation are dedicated to maximaze searching for the most adversarial example $x^{\ast}$ to test adversarial robustness.

\textbf{The comparison with Targeted-DLR.} Targeted-DLR heuristically rescales the logits $z$ in Eq.~(\ref{Eq-T-DLR}) \citep{croce2020reliable}. 
For the logit of true target $z_y$ and a false target $z_t$, the relative magnitude of $z_y$ and $z_t$ is constant under different rescaling methods. However, though the rescaling methods preserve the sign of ${z}_y-{z}_t$, they may lead to different adversarial variants. %From the optimization perspective, there is not a kind of 
% Clearly, there is no perfect rescaling method that can perform the best in any situation but
% inappropriate rescaling methods will hinder the searching for the most adversarial example. 
Below, we conduct experiments to investigate the difference of using different rescaling methods.

The experimental setting follows \citep{Madry18PGD}, and we replace the CE loss with seven different logits rescaling methods. In Table~\ref{table:normform_new}, the successful set denotes the number of examples that can be attacked successfully in the test set of CIFAR-10 ($10,000$ examples). As shown in Table~\ref{table:normform_new}, inappropriate logits rescaling methods (e.g., Targeted-DLR in AA) reduce the reliability, and the method using natural logits (no rescaling) performs the best among the seven methods. The results motivate us to use it as the loss function. Note we do not deny that there could be a better rescaling method, but still, a reasonably good result can be obtained by no rescaling (natural logits in Table~\ref{table:normform_new}). 

We also investigate the difference among different successful sets. The non-empty difference sets $A \cup B_i - A$ and $A \cup B_i - B_i$ in Table~\ref{table:normform_new} suggest that diverse logits rescalings can be considered when considerable computational resources are available, we analyze it in Appendix~\ref{app:diverse}. 
%Hence, for searching the most adversarial example $x^{\ast}$, we propose margin $z_y-z_t$ in place of the loss function $-\log(p_y)$ in PGD.
\begin{table*}[!h]
\setlength{\tabcolsep}{4.7mm}
%\small
%\footnotesize
\scriptsize
\renewcommand\arraystretch{1.7}
\centering
\caption{The successful attack set of different rescaling methods on CIFAR-10. We compare the degree of overlap of the successful set between 6 different rescaling methods (ID: $B_i, i = 1,2,...,6$) and no rescaling methods (ID: $A$). $A \cup B_i - A$ denotes the number of instances can be attacked successfully by method $B_i$ but failed by method $A$. $A \cup B_i - B_i$ denotes the number of instances can be attacked successfully by method $A$ but failed by method $B_i$. For 6 different rescaling methods, larger $A \cup B_i - A$ and smaller $A \cup B_i - B_i$ mean better rescaling methods.  Following \citep{Madry18PGD}, the attack setting is PGD-20.}
\label{table:normform_new}
\begin{tabular}{c|c|c|c|c|c|c|c}
\toprule[1.5pt]
ID & Rescaling method & Formulation & Set size & Ranking & diff. & $A \cup B_i - A$ & $A \cup B_i - B_i$\\
\midrule[0.6pt]
\midrule[0.6pt]
$A$ & Natural logits & $- (z_y - z_t)$ & \textbf{5219} & 1 & 0 & / & /\\
\midrule    
$B_1$ & Softmax & $-\frac{{e}^{z_y}-{e}^{z_t}}{\sum_{i=0}^{K}{e}^{z_i}}$ & 5172 & 2 & -47 & 4 & 51\\
\midrule    
$B_2$ & Max & $- \frac{z_y - z_t}{z_{y}}$ & 5165 & =3 & -54 & 5 & 59\\
\midrule    
$B_3$ & Sum & $- \frac{z_y - z_t}{z_{y}+z_{t}}$ & 5165 & =3 & -54 & 5 & 59\\
\midrule    
$B_4$ & Min-Max & $- \frac{z_y - z_t}{z_{\pi_1}-z_{\pi_{10}}}$ & 5121 & 5 & -98 & 3 & 101\\
\midrule    
$B_5$ & DLR & $- \frac{z_y - z_t}{z_{\pi_1}-\frac{1}{2}\cdot(z_{\pi_3}+z_{\pi_4})}$ & 5078 & 6 & -141 & 2 & 143\\
\midrule    
$B_6$ & Sigmoid & $-(\frac{{e}^{z_y}}{1+{e}^{z_y}} - \frac{{e}^{z_t}}{1+{e}^{z_t}})$ & 4820 & 7 & -399 & 1 & 400\\
\bottomrule[1.5pt]
\end{tabular}
\vskip1ex%
\vskip -0ex%
\vspace{-0.5em}
\end{table*}\

\begin{algorithm}[!t]
\footnotesize
\caption{MM Attack}
\label{alg:Alg__MM}
\begin{algorithmic}[1]
\STATE \textbf{Input:} natural data $x$, true label $y$, set of false labels $C$, model $f$, loss function $\ell_{MM}$, maximum number of PGD steps $N$, perturbation bound $\epsilon$, initial step size $\alpha$, the number of classes $K$, targets selection number $K_s$, checkpoints set $W$;
\STATE \textbf{Output:} adversarial data $x'$;
\WHILE{$K_s > 0$} 
\STATE$x'_{0} \gets x$; 
\STATE $x'_{max} \gets x$; 
\STATE $f_{max} \gets f(x'_{0})$; \STATE $c = \argmax_{i \in C} {f(x)}_i$;
\FOR{$k=0$ \textbf{to} $N-1$}
\STATE $x'_{k+1} \gets \Pi_{\mathcal{B}_{\epsilon}[x]}\big(x'_{k}+\alpha sign(\nabla_{x'_k}\ell_{MM}(f(x'_{k}),y,c)\big)$;
\IF{$f(x'_{k+1}) > f_{max}$} 
\STATE $x'_{max} \gets x'_{k+1}$; \STATE $f_{max} \gets f(x'_{k+1})$;
\ENDIF
\IF{$k \in W$ \textbf{and} (Condition 2 \textbf{or} Condition 3)} 
\STATE $\alpha \gets \alpha/2$; 
\STATE $x'_{k+1} \gets x'_{max}$;
\ENDIF
\ENDFOR
\STATE $C \gets C \setminus \{c\}$;
\IF{$\argmax_{i \in C} {f(x')}_i \neq y$} 
\STATE $K_s \gets 0$;
\ENDIF
\STATE $K_s \gets K_s - 1$;
\ENDWHILE
%\vspace{1mm}
\end{algorithmic}
\end{algorithm}
\subsection{Sequential TArget Ranking Selection (STARS)}
\label{sec:STP}
As mentioned in the introduction, for multi-class and untargeted attacks, the reliable method is to minimize the loss function $-(z_y-z_t)$ for each target $t\neq y$, and then take the most adversarial one \citep{gowal2019alternative, croce2020reliable}. Since the computational cost of such a solution depends on the number of classes, it is unacceptable for practitioners with limited computational resources. Here, we propose a fast solution STARS which saves a large amount of running time with little-to-no performance lost. STARS consists of two strategies: pre-selecting targets and ranking sequential attack.
% We call our solution the \emph{Sequential TArget Ranking Selection} (STARS), which consists of two strategies.
\begin{figure*}[!t]
    \begin{center}
        \subfigure[Acc on CIFAR-10]
        {\includegraphics[width=0.329\textwidth]{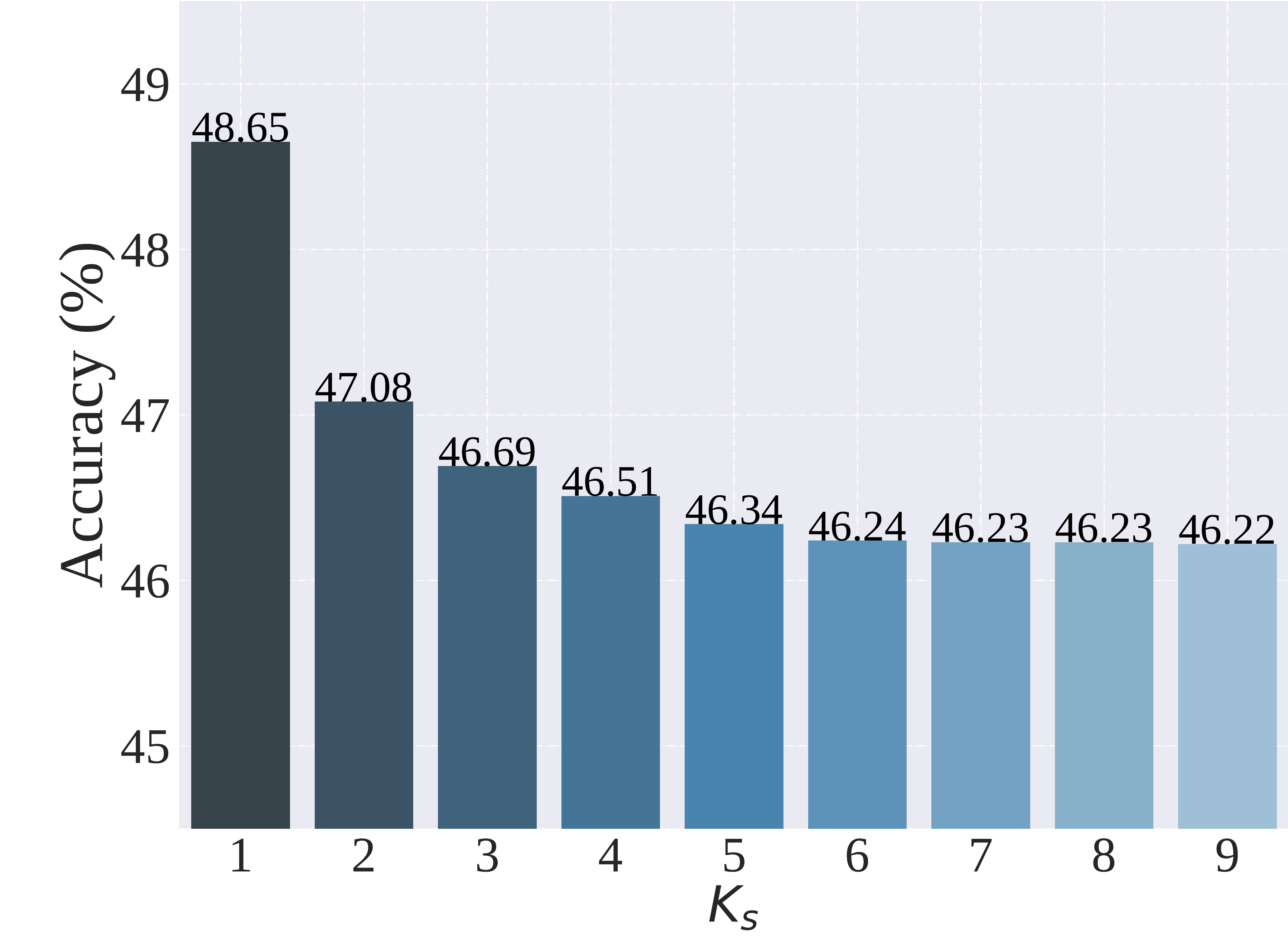}}
        \subfigure[Acc on CIFAR-100]
        {\includegraphics[width=0.329\textwidth]{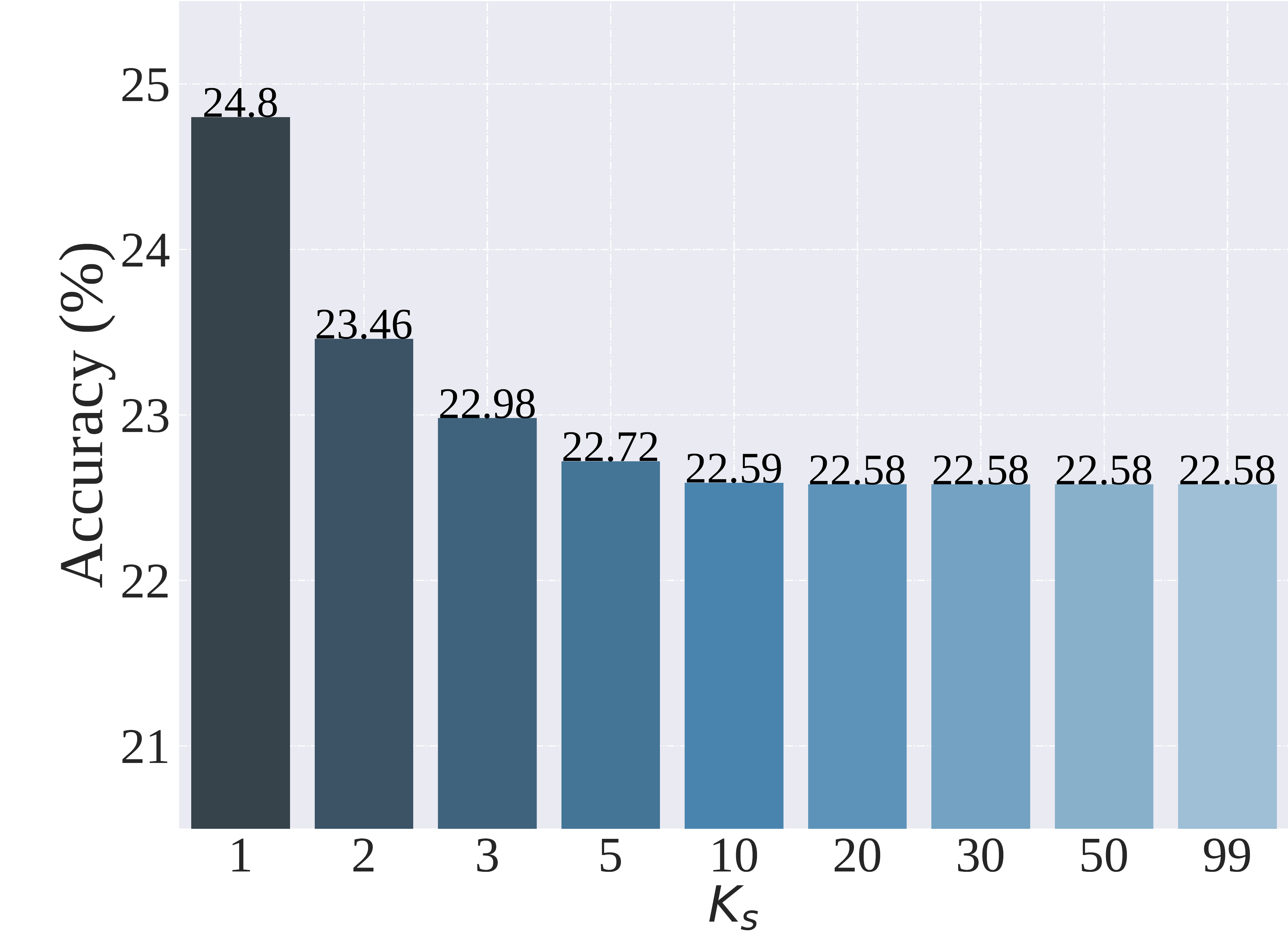}}
        \subfigure[Acc on SVHN]
        {\includegraphics[width=0.329\textwidth]{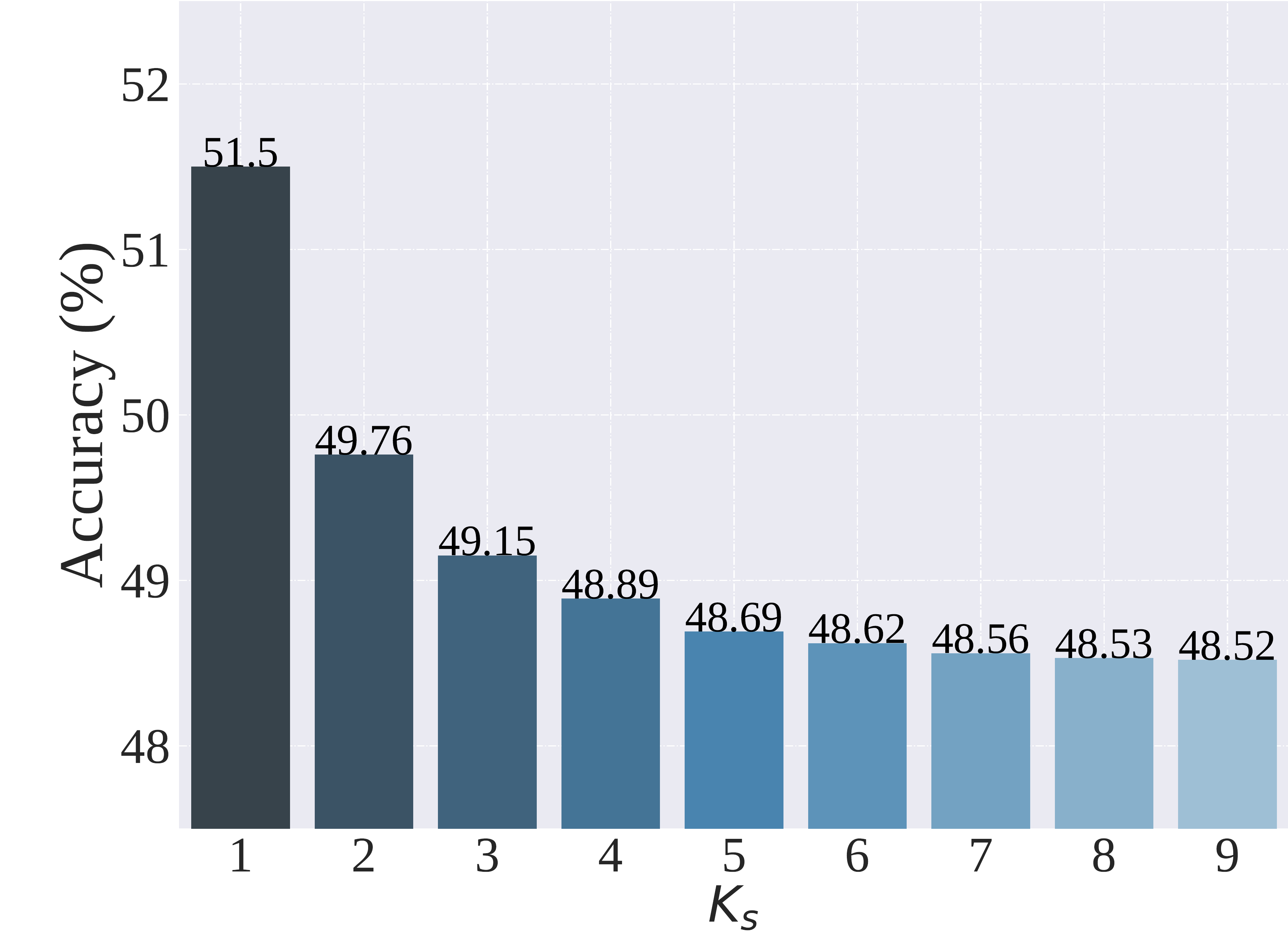}}
        \subfigure[Time on CIFAR-10]
        {\includegraphics[width=0.329\textwidth]{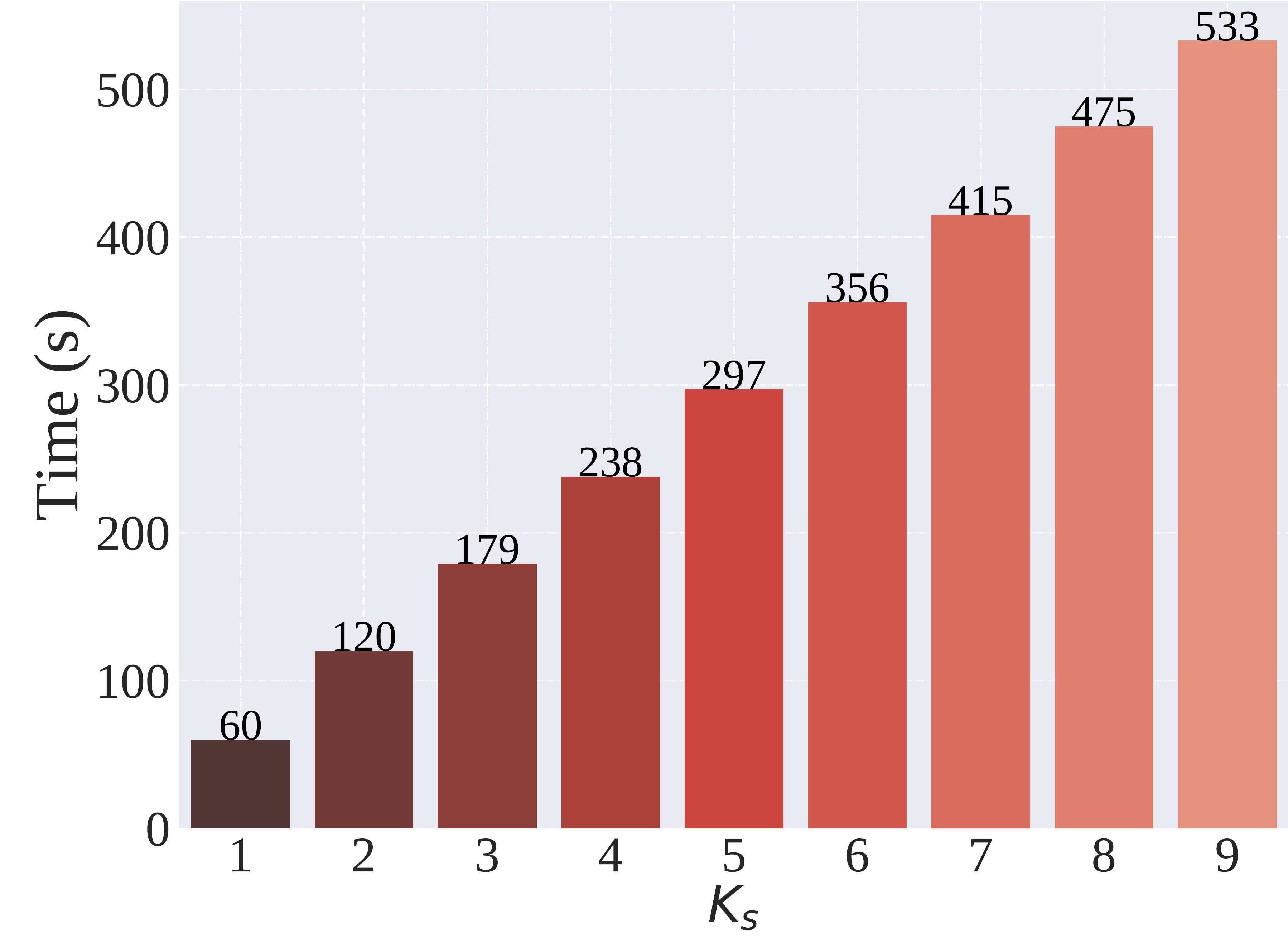}}
        \subfigure[Time on CIFAR-100]
        {\includegraphics[width=0.329\textwidth]{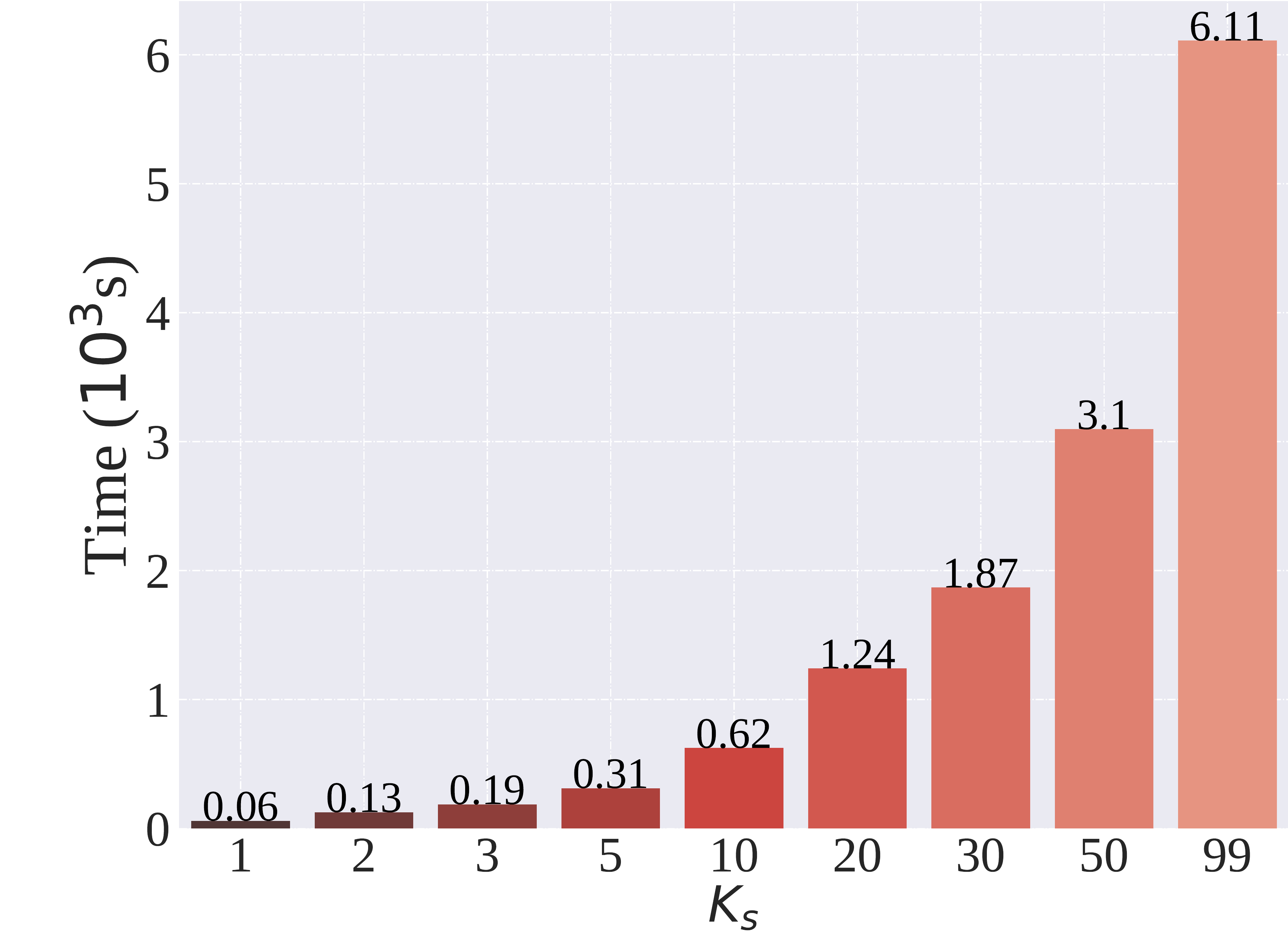}}
        \subfigure[Time on SVHN]
        {\includegraphics[width=0.329\textwidth]{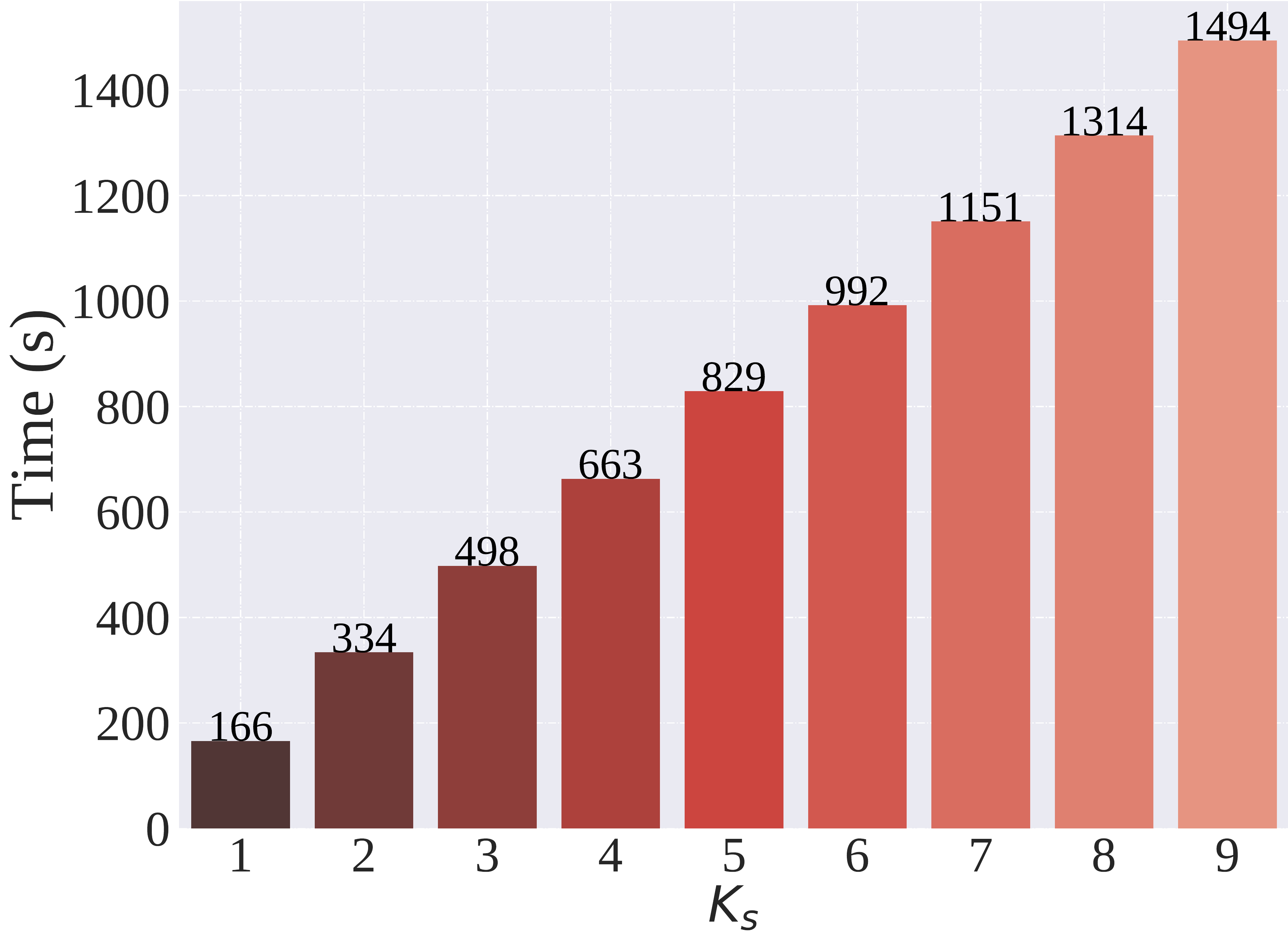}}
        \vspace{-1em}
        \caption{\footnotesize Reliability and computational time of different selection number. Regarding the performance for all targets as benchmarks, as shown in the subfigure (a), (b) and (c), pre-selecting 3 (or 5) targets  (the first strategy in STARS) can achieve 99.13\% (or 99.78\%) reliability on CIFAR-10, 99.49\% (or 99.82\%) reliability on CIFAR-100; 98.78\% (or 99.67\%) reliability on SVHN; as shown in the subfigure (d), (e) and (f), pre-selecting 3 (or 5) targets only costs 33.58\% (or 55.72\%) computational time on CIFAR-10, 3.11\% (or 5.07\%) computational time on CIFAR-100; 33.33\% (or 55.49\%) computational time on SVHN. 
        }
        % \vspace{-2em}
    \label{fig:Acc_Time}
    \end{center}
    % \vspace{-1em}
\end{figure*}

% The first strategy is \emph{pre-selecting targets}.
\textbf{Pre-selecting-Targets Strategy.}
Given a natural input $x$ and a $K$-class classifier $f$, denoting the predicted probability as ${f(x)}_i$ for a false label $i$, a natural intuition is that the false target $i$ with a higher value of ${f(x)}_i$ is more likely to lead to a successful attack. To verify this intuition, in Figure~\ref{fig:Acc_Time}, we compare the performance between only selecting the false targets with $K_s$ highest predicted probabilities and selecting all the $K-1$ false targets. The results show that only selecting $K_s$ (e.g., $K_s = 3$) highest targets achieves comparable performance. Note that the strategy is mentioned in \citep{gowal2019alternative} but has a poor performance limited by the used attack with fixed step size. For the ranking of predicted probability ${f(x)}_i$, we also investigate the difference of replacing the natural input $x$ with adversarial examples in Appendix~\ref{app:ranking}, which shows that the replacement only has limited improvements. For the sake of efficiency, we recommend to use the ranking of ${f(x)}_i$ of the natural input $x$, which does not need extra computation. 

\begin{figure*}[!h]
    \begin{center}
        \subfigure[Evaluation on CIFAR-10]
        {\includegraphics[width=0.495\textwidth]{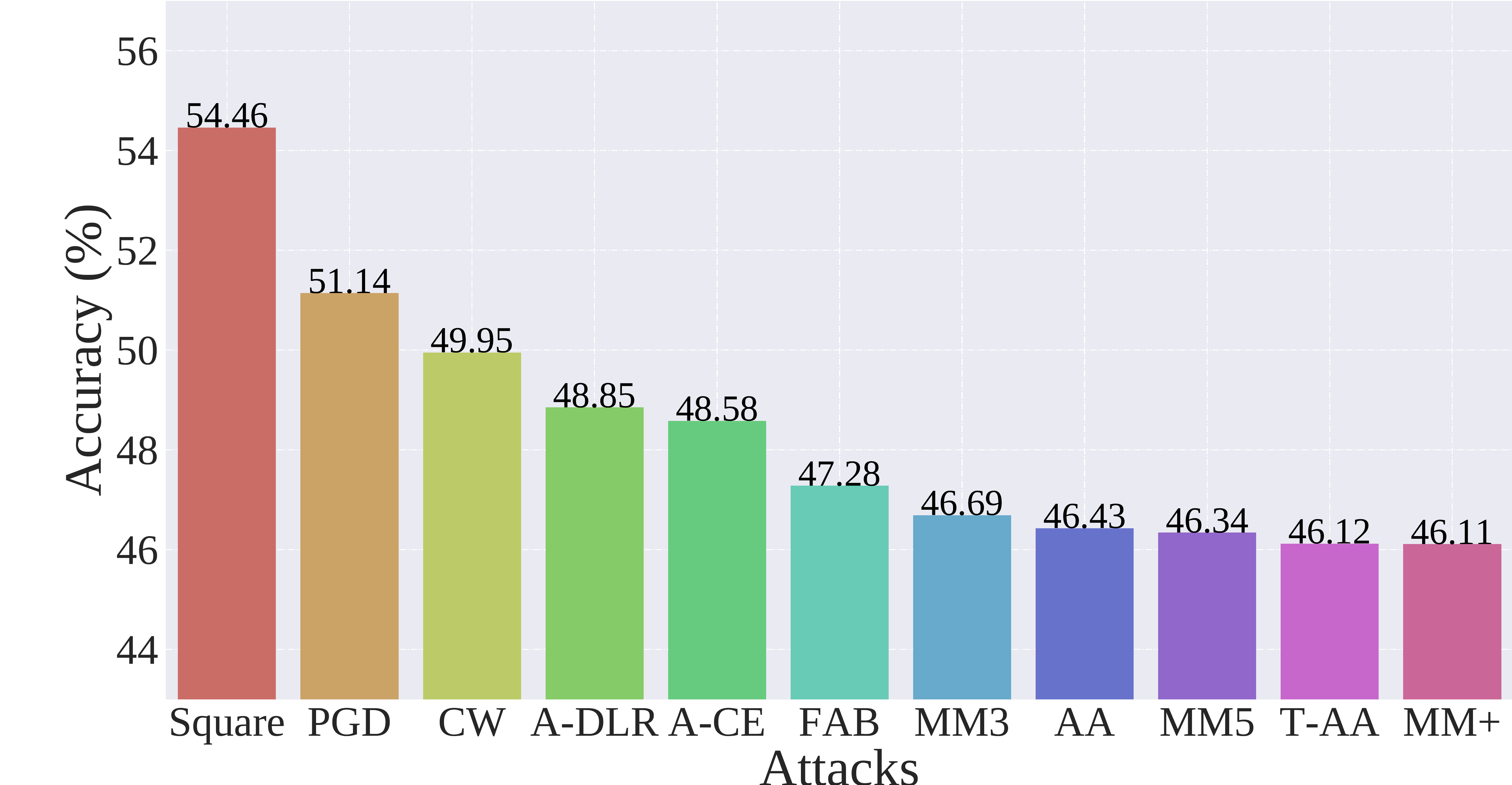}}
        \subfigure[Computational time on CIFAR-10]
        {\includegraphics[width=0.495\textwidth]{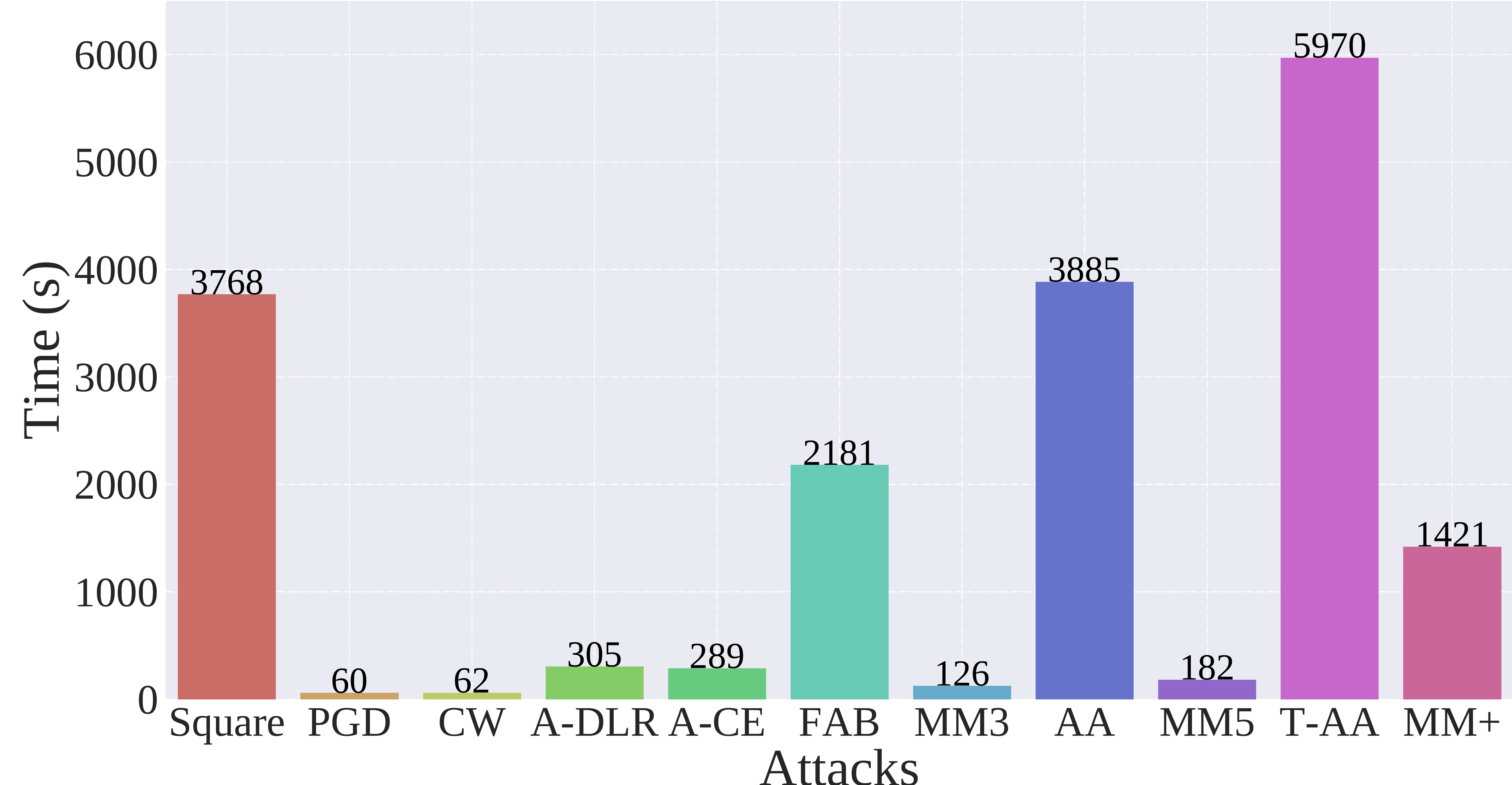}}
        \subfigure[Evaluation on CIFAR-100]
        {\includegraphics[width=0.495\textwidth]{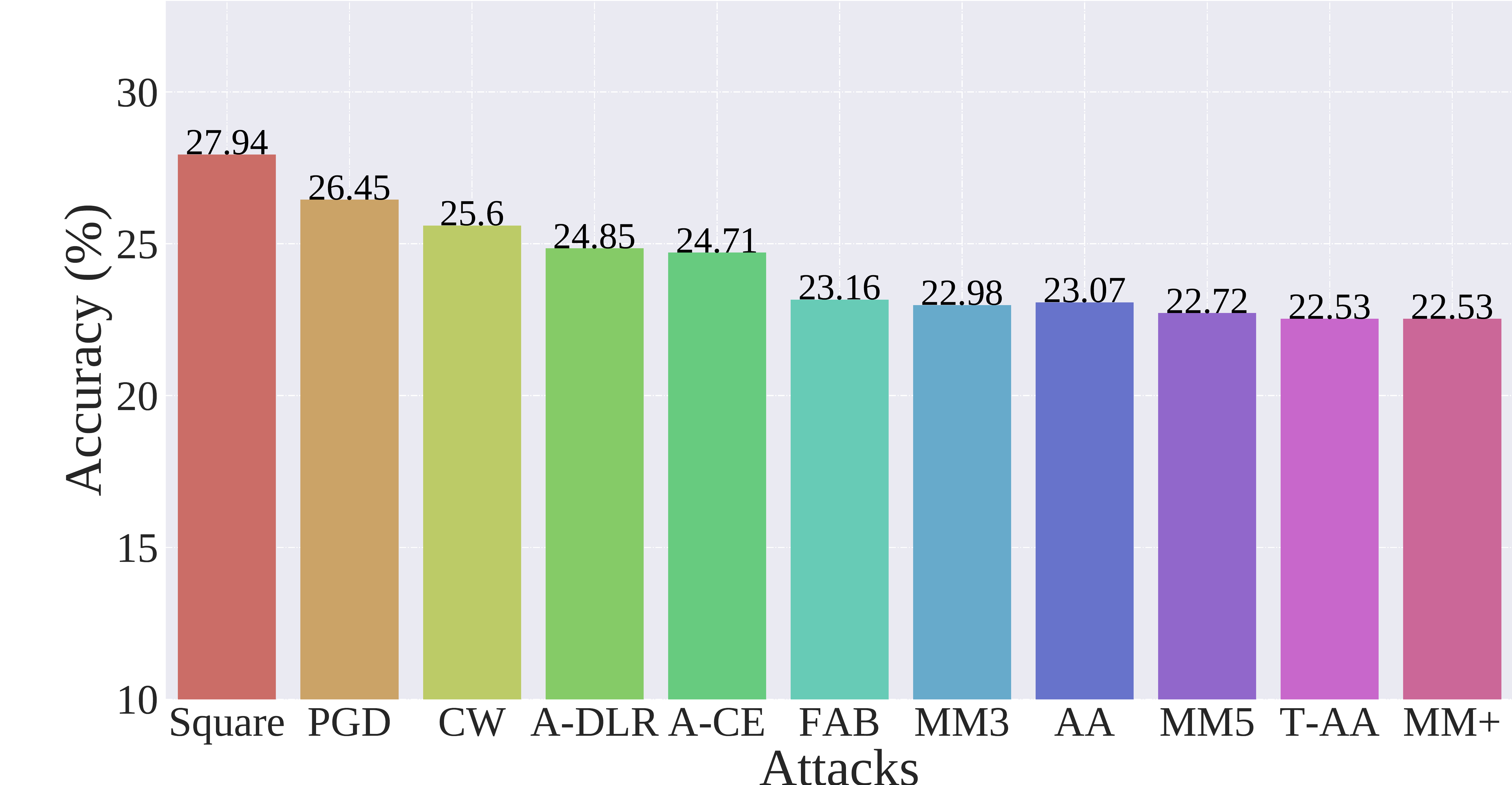}}
        \subfigure[Computational time on CIFAR-100]
        {\includegraphics[width=0.495\textwidth]{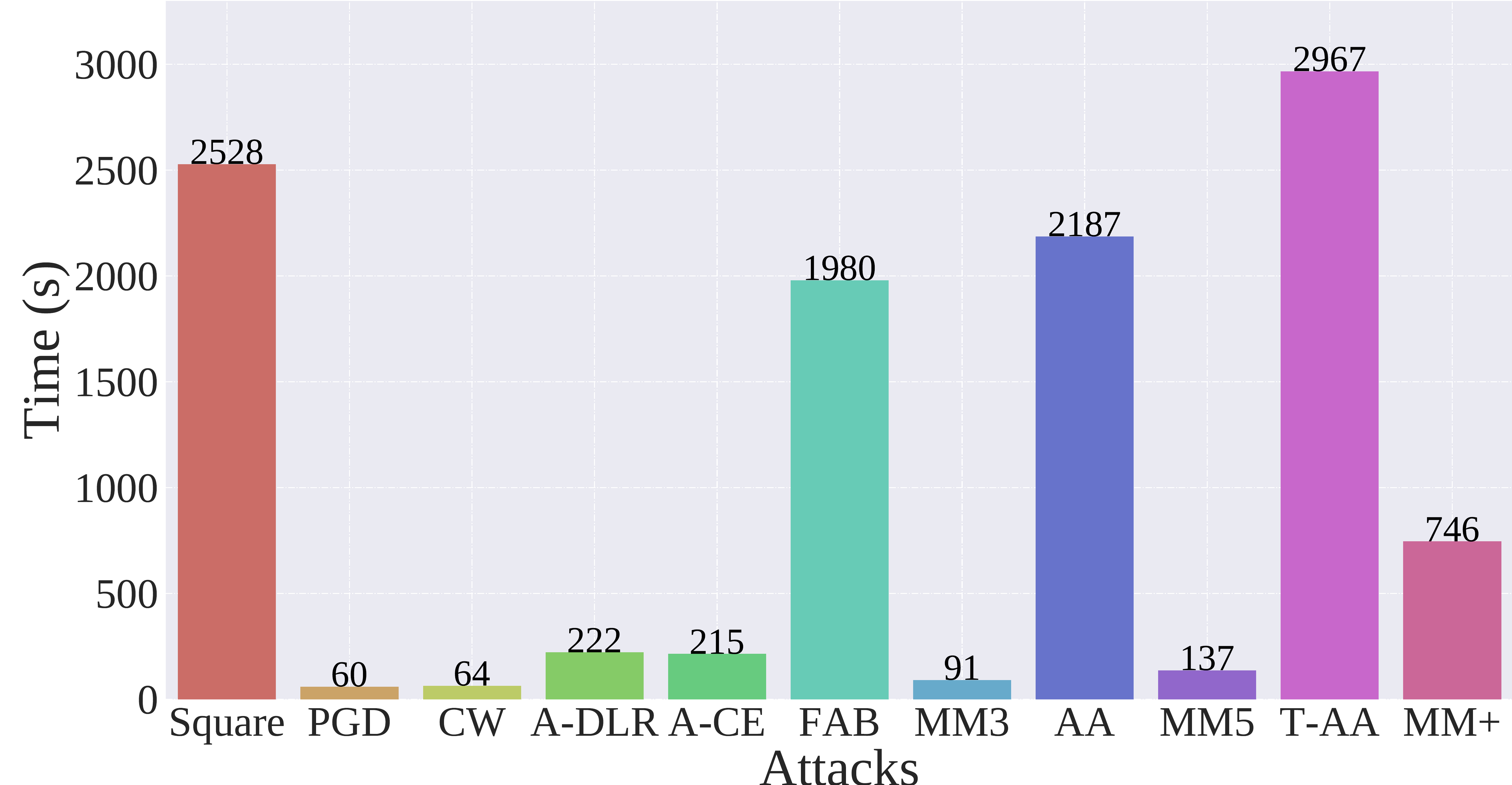}}
        \subfigure[Evaluation on SVHN]
        {\includegraphics[width=0.495\textwidth]{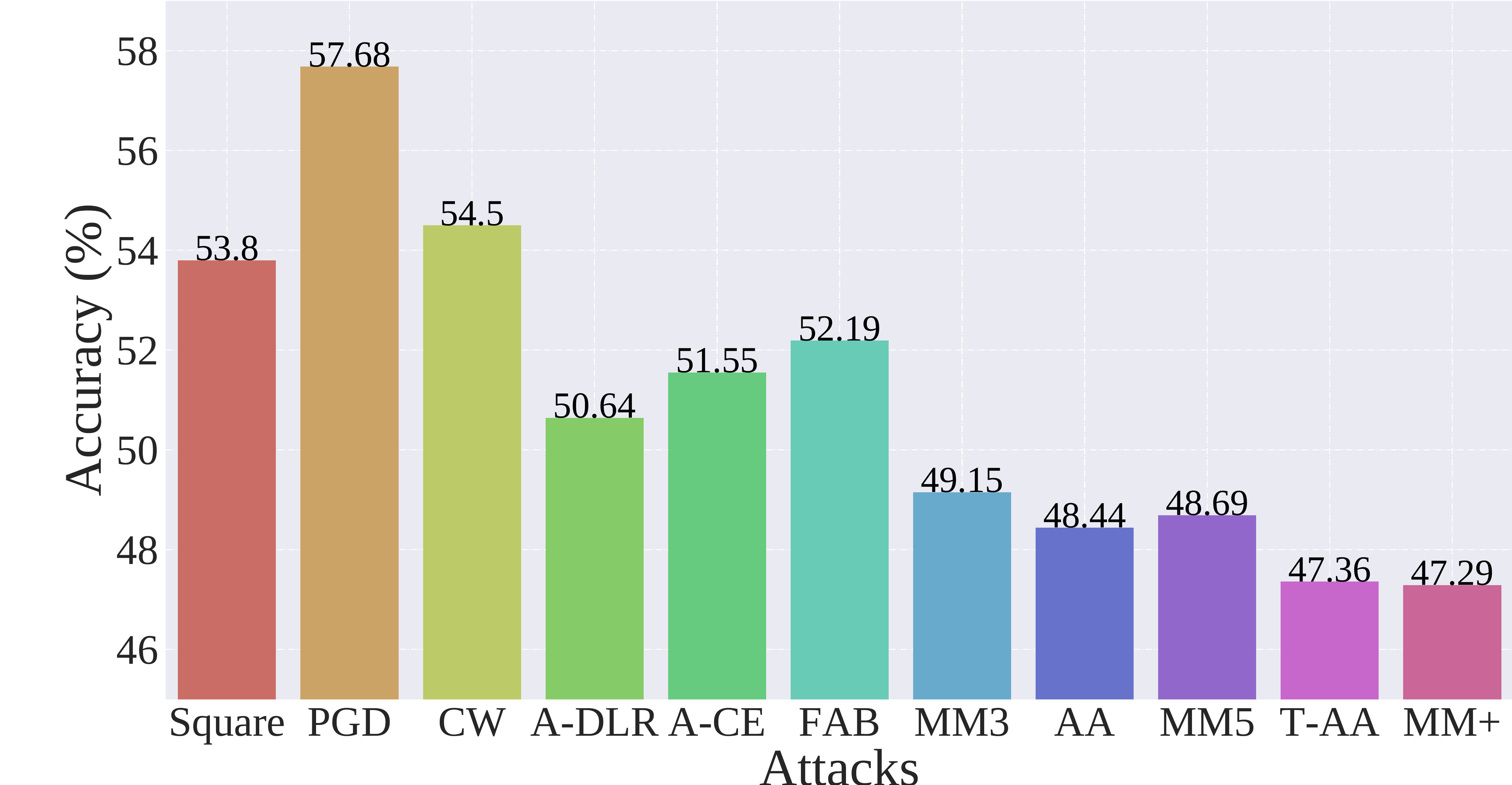}}
        \subfigure[Computational time on SVHN]
        {\includegraphics[width=0.495\textwidth]{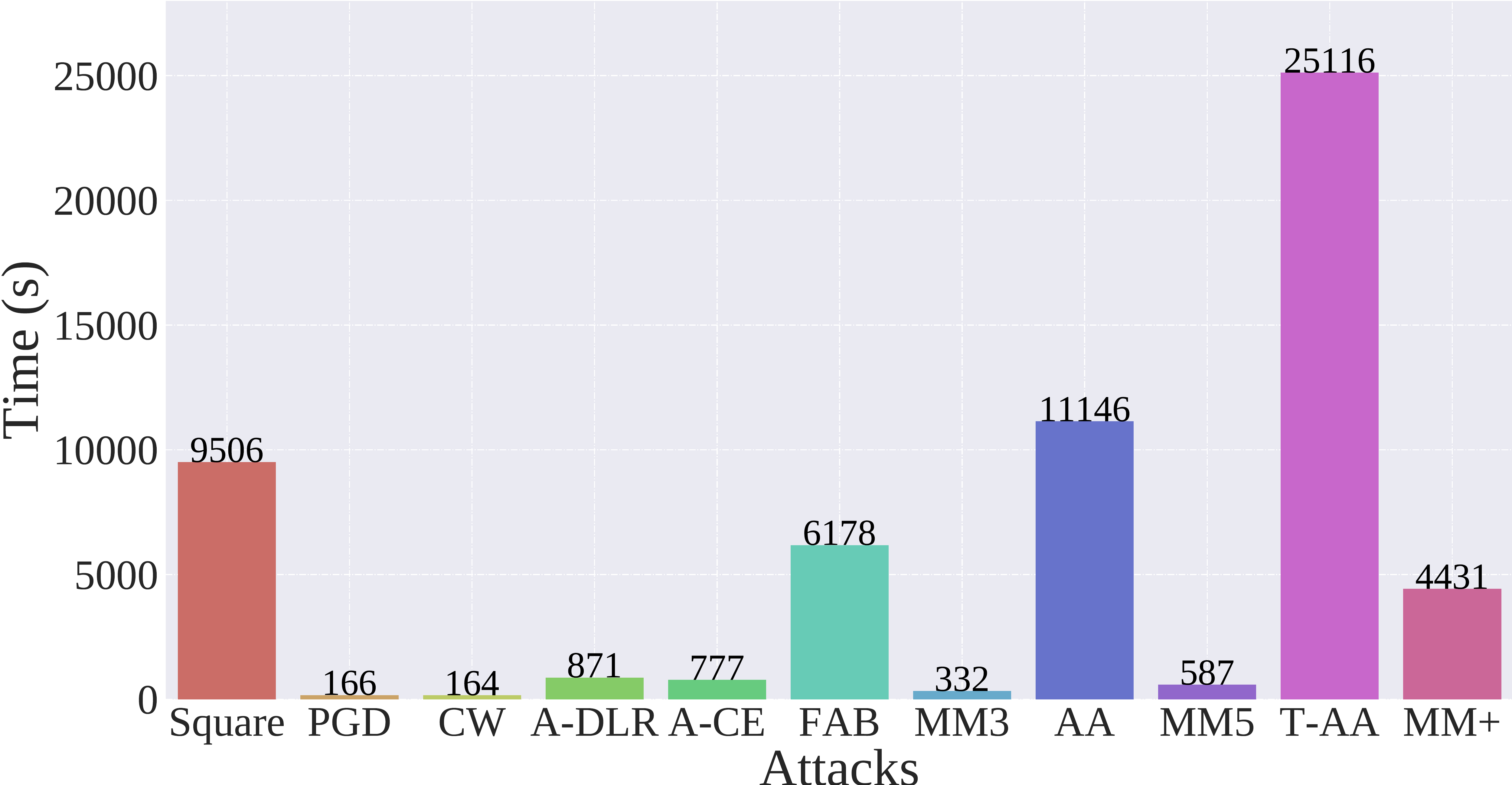}}
        \vspace{-1em}
        \caption{\footnotesize Comparison of reliability and computational cost. We compare three versions of our MM attack (MM3, MM5 and MM+ mentioned in Section~\ref{exp:baselines}) with 8 baselines. In subfigure (a), (c) and (e), the Y-axis is the accuracy of the attacked model, which means that the lower the accuracy, the stronger the attack (or to say the better evaluation). In subfigure (b), (d) and (f), the Y-axis is computational time, which means the less the time, the higher the computational efficiency. 
        %In the subfigure (b), the gray shape is a hypothetical distribution of all adversarial variants maps within the bounded perturbation epsilon on a natural example; $\mathbb{P}_t$ and $\mathbb{P}_y$ are the predicted probability on a targeted false label $t$ and the true label $y$; The orange area ($\mathbb{P}_t > \mathbb{P}_y$) indicates that the adversarial variants inside can be misclassified, or to say attack successfully, while the blue area ($\mathbb{P}_t < \mathbb{P}_y$) indicates that the adversarial variants inside cannot attack successfully.
        }
        \vspace{-1em}
    \label{main_exp}
    \end{center}
    % \vspace{-0.5em}
\end{figure*}

% The second strategy is ranking sequential attack.
\textbf{Ranking-Sequential-Attack Strategy.}
This strategy aims to perform a sequential attack based on the ranking of predicted probability on false targets. First, consider the false target $i$ with the highest predicted probability ${f(x)}_i$; if the attack succeeds, then we need to terminate attacks on other targets; otherwise, we need to continue considering the false target with the second highest predicted probability. The sequential attack strategy is simple and effective for saving computational time in the \emph{multi-target problem}.
% The strategy can save computational time without sacrificing reliability.

\subsection{Minimum-Margin Attack}
\label{sec:alg_MM}
With the above strategies, we summarize our scheme of MM attack in Algorithm~\ref{alg:Alg__MM}. We follow the setting of the adaptive step size selection in \citep{croce2020reliable} and specify checkpoints $W=\{w_{0}=0,...,w_{n}\}$ at which the MM attack decides whether to halve the current step size. The two conditions in Algorithm~\ref{alg:Alg__MM} are:
\begin{align*}
\label{Condition1}
\textbf{Condition 2. } &\sum_{i = w_{j-1}}^{w_{j}-1}1_{f(x'_{i+1})>f(x'_{i})}<\beta\cdot(w_j-w_{j-1})\text{.}\\
\textbf{Condition 3. }&\alpha^{w_{j-1}} \equiv \alpha^{w_j}~\text{and}~f_{max}^{w_{j-1}} \equiv f_{max}^{w_j}\text{.}
\end{align*}

$\sum_{i = w_{j-1}}^{w_{j}-1}1_{f(x'_{i+1})>f(x'_{i})}$ counts how many cases since $w_{j-1}$ the update step has been successful in increasing $f$, to determine whether $\sum_{i = w_{j-1}}^{w_{j}-1}1_{f(x'_{i+1})>f(x'_{i})}$ is less than a fraction $\beta$ of the total update steps $w_j-w_{j-1}$. The $\beta$ is a hyper-parameter in (0,1), and following \citep{croce2020reliable}, we use $\beta = 0.75$. $\alpha$ is the initial step size in Algorithm~\ref{alg:Alg__MM}.

According to different choices of $K_s$, we denote MM3 (or MM5) as the MM attack with $K_s=3$ (or $K_s=5$). Our experiments show that with the help of STARS method, MM3 attack saves 76.36\% of the computational time on CIFAR-10, 98.51\% on CIFAR-100 and 77.78\% on SVHN.

For one thing, we are \emph{the first} to propose searching the most adversarial example with \emph{minimum margin}, combining the loss function of MM attack with adaptive step size to achieve the proposal is \emph{new}. Although some of components exist in previous research \citep{carlini2017towards,gowal2019alternative,croce2020reliable}, \emph{the existence is not the most important and doesn't degrade our novelty about how best to combine and use existing methods}, and we provide clear evaluation and analysis why similar methods fail in previous works but can succeed in MM attack. For another, our STARS method consists of two novel strategies. Apart from the pre-selecting-targets strategy, we are the first to propose the sequential attack strategy which is \emph{simple and effective} for saving computational time.
\begin{table*}[!t]
\setlength{\tabcolsep}{4.2mm}
%\footnotesize
\scriptsize
\renewcommand\arraystretch{1.5}
\centering
\caption{Test accuracy (\%) of adversarial training. Robust test accuracy (\%) of MM3-F10 (\emph{ours}), MM5-F20 (\emph{ours}), MM3 (\emph{ours}) and baselines (PGD, CW) on CIFAR-10. \emph{Diff.} represents the difference between the current result and the optimal result in the row. The model structure of all methods is ResNet-18. Bold values represent the highest accuracy in each row.
}
\label{table:exp_AT}
\begin{tabular}{c|c c|c c|c c|c c|c c}
\toprule[1.5pt]
Methods & PGD & Diff. & CW & Diff. & MM3-F10 & Diff. & MM3-F20 & Diff. & MM3 & Diff.\\
\midrule[0.6pt]
\midrule[0.6pt]
PGD (Test) & 51.14 & -4.10 & 51.47 & -3.77 & 54.96 & -0.28 & \textbf{55.24} & 0.00 & 55.04 & -0.20 \\
\midrule
CW (Test) & 49.95 & -1.89 & \textbf{53.26} & 0.00 & 51.18 & -2.08 & 51.16 & -2.10 & 51.84 & -1.42 \\
\midrule
A-CE (Test) & 48.58 & -3.92 & 48.16 & -4.34 & 51.55 & -0.95 & \textbf{52.50} & 0.00 & 52.22 & -0.28 \\
\midrule
A-DLR (Test) &48.85 & -1.44 & \textbf{52.76} & 0.00 & 49.78 & -2.98 & 49.88 & -2.88 & 50.29 & -2.47 \\
\midrule
FAB (Test) & 47.28 & -1.22 & 47.13 & -1.37 & 47.83 & -0.67 & 48.28 & -0.22 & \textbf{48.50} & -0.00 \\
\midrule
Square (Test) & 54.46 & -0.66 & \textbf{55.32} & 0.00 & 54.80 & -0.52 & 54.83 & -0.49 & 55.12 & -0.20 \\
\midrule
AA (Test) & 46.43 & -1.85 & 46.36 & -1.92 & 47.62 & -0.66 & 47.84 & -0.44 & \textbf{48.28} & -0.00 \\
\midrule
T-AA (Test) & 46.12 & -0.97 & 45.26 & -1.83 & 46.39 & -0.70 & 46.73 & -0.36 & \textbf{47.09} & -0.00 \\
\midrule
MM3 (Test) & 46.69 & -1.17 & 46.77 & -1.09 & 47.20 & -0.66 & 47.48 & -0.38 & \textbf{47.86} & -0.00 \\
\midrule
MM9 (Test) & 46.21 & -0.95 & 45.36 & -1.80 & 46.49 & -0.67 & 46.82 & -0.34 & \textbf{47.16} & -0.00 \\
\midrule
MM+ (Test) & 46.12 & -0.90 & 45.22 & -1.80 & 46.39 & -0.63 & 46.68 & -0.34 & \textbf{47.02} & -0.00 \\
\bottomrule[1.5pt]
\end{tabular}
\vskip1ex%
\vskip -0ex%
\vspace{-1em}
\end{table*}\
%As shown in Figure~\ref{fig:Acc_Time}, a target class with more predicted probability has a greater chance of attacking successfully. The finding
%The method by pre-selecting partial targeted classes to simplify MMP attack is called MMP attack(s). 
%By taking care of the above factors, we propose a novel MMP attack

\section{Experiments}
\label{sec:exp}
\textbf{Datasets.}
We conducted experiments on CIFAR-10, SVHN and CIFAR-100. We consider $L_{\infty}$-norm bounded perturbation that $||\tilde{\vx}-\vx{||}_\infty \leq \epsilon$ in both training and evaluations. The images of all datasets are normalized into [0,1].
%In this section, we justify the efficacy of our MM attack. In the experiments, we consider $L_{\infty}$-norm bounded perturbation (i.e., $||x'-x{||}_\infty \leq \epsilon$) in both training and evaluations. 

\textbf{Baselines.}
\label{exp:baselines}
We evaluate 3 versions of our MM attack and compare them with 8 baselines. MM3 (or MM5) is the version of MM attack with maximum step number $K = 20$ and targets selection number $K_s=3$ (or $K_s = 5$); MM+ is the version of MM attack with maximum step number $K = 100$ and targets selection number $K_s=9$. Baselines consist of PGD-20 (PGD) \citep{Madry18PGD}, CW-20 (CW) \citep{carlini2017towards}, APGD-CE-100 (A-CE), APGD-DLR-100 (A-DLR), FAB, Square, AutoAttack (AA) and Targeted-AutoAttack (T-AA) \citep{croce2020reliable}. %Although \citet{brendel2019accurate} consider the Accurate, reliable and fast robustness evaluation, its attack performance is far below the start-of-the art benchmark Autoattack. 

\textbf{Experimental Setup.}
\label{exp:setup}
We verify our methods on the ResNet-18 \citep{he2016deep} and the Wide-ResNet-34 (WRN-34) \citep{zagoruyko2016wide} using three benchmark datasets: CIFAR-10, CIFAR-100, SVHN. The \emph{adversarial training}~(AT) method follows \citet{Madry18PGD}. The training setup follows previous works \citep{Madry18PGD,ZhangYJXGJ19TRADES} that all networks are trained for $100$ epochs using SGD with $0.9$ momentum. The initial learning rate is $0.1$ (0.01 for SVHN), and is divided by $10$ at epoch $60$ and $90$, respectively. The weight decay is $0.0002$ ($0.0035$ for SVHN). The previous work \citep{rice2020overfitting} observed that overfitting in robust adversarial training hurts test set performance. Thus, following \citet{rice2020overfitting}, we compare different methods based on the performance of the best checkpoint model (results at epoch $60$). For generating the adversarial data for updating the network, we set the $L_{\infty}$-norm bounded perturbation $\epsilon_{train} = 8/255$; the maximum number of PGD steps is $K = 10$; step size $\alpha = \epsilon_{train} / 10$. In testing, unless otherwise specified, we set $L_{\infty}$-norm bounded perturbation $\epsilon_{test} = 8/255$. 

%We verify our methods on different neural networks including ResNet-18 \citep{he2016deep} and Wide-ResNet-34 (WRN-34) \citep{zagoruyko2016wide} using different datasets: CIFAR-10, CIFAR-100, SVHN. The \emph{adversarial training}~(AT) method in this paper follows \citep{Madry18PGD}. The training setup follows previous works \citep{Madry18PGD,ZhangYJXGJ19TRADES} that all networks are trained for 100 epochs using SGD with $0.9$ momentum. The initial learning rate is $0.1$ (0.01 for SVHN), and is divided by $10$ at epoch $60$ and $90$, respectively. The weight decay is 0.0002 (0.0035 for SVHN). The previous work \citep{rice2020overfitting} observed that overfitting in robust adversarial training hurts test set performance. Thus, following \citet{rice2020overfitting}, we compare different methods based on the performance of the best checkpoint model (the early stopping results at epoch 60). For generating the adversarial data for updating the network, we set the $L_{\infty}$-norm bounded perturbation $\epsilon_{train} = 8/255$; the maximum number of PGD steps is $K = 10$; step size $\alpha = \epsilon_{train} / 10$. In testing, unless otherwise specified, we set $L_{\infty}$-norm bounded perturbation $\epsilon_{test} = 8/255$. 
For PGD and CW, we follow the setting in \citep{zhang2020attacks}: the maximum number of steps $K = 20$, and the step size $\alpha = \epsilon / 4$. For APGD-CE, APGD-DLR, FAB and their corresponding targeted version, we follow the setting in \citep{croce2020reliable}: the maximum  number of steps $K = 100$. There is a random start in training and testing, i.e., uniformly random perturbations ($[-\epsilon_{train}, +\epsilon_{train}]$ and $[-\epsilon_{test}, +\epsilon_{test}]$) are added to natural instances.

%other hyper-parameters are the same as that of PGD above.

\textbf{Performance Evaluation.}
In Figure~\ref{fig:moti} and Figure~\ref{main_exp}, we compare our proposed MM attack with 10 baselines. We report the performance of our MM attack and all baselines on CIFAR-10, CIFAR-100, SVHN with the model structure chosen as ResNet-18. Following \citet{Madry18PGD}, the threat model is trained by PGD-10. As shown in Figure~\ref{fig:moti} and Figure~\ref{main_exp}, first, our MM attack can perform better than any single attack of PGD, CW, A-DLR, A-CE and FAB; second, compared with the ensemble of diverse attacks AA and T-AA, our MM attack achieves comparable performance but only incurs a very small amount of computational time. 

In Appendix~\ref{app:exptable}, experiments on the large-capacity network WRN-34 are provided. Besides the defense by \citet{Madry18PGD}, we conduct extensive experiments between our MM attack with baselines on 12 defenses of RobustBench in Figure~\ref{robustbench_exp1}, Figure~\ref{robustbench_exp2} and Figure~\ref{robustbench_exp3} (Appendix~\ref{app:robustbench}) \citep{sehwag2021improving, rade2021helper, rebuffi2021fixing, andriushchenko2020understanding, gowal2020uncovering, sridhar2021robust, wong2020fast, robustness, carmon2019unlabeled, wang2019improving, wu2020adversarial, zhang2020geometry}. For $L_{2}$-norm bounded perturbation that $||\tilde{x}-x{||}_2 \leq \epsilon$, we also compare the performance on the 6 defenses of RobustBench in Figure~\ref{robustbench_exp4} and Figure~\ref{robustbench_exp5} (Appendix~\ref{app:robustbench}) \citep{sehwag2021improving, rade2021helper, rice2020overfitting, rebuffi2021fixing, robustness, augustin2020adversarial}. We test MM attack and all baselines on the well-trained models which are available for download in Robustbench. Experimental results verify the effectiveness of MM attack on diverse trained model in RobustBench. To the best of our knowledge, our MM attack is current \emph{SOTA benchmark}, which provides a new direction of evaluating adversarial robustness. The code of our MM attack is available at \httpsurl{github.com/Sjtubrian/MM-attack}.

%Compared with AA, our proposed MM attack achieves \emph{comparable performance} but only costs \emph{very little of the computational time}. 
%In the field of adversarial robustness, the \emph{AutoAttack} (AA) is widely regarded as the most authoritative evaluation of adversarial robustness, e.g., nearly all papers in ICML/NeurIPS/ICLR in the \emph{RobustBench} leaderboard. 

\textbf{Adversarial Training with MM Attack.}
By injecting adversarial examples into the training data, \emph{adversarial training}~(AT) methods seek to train an adversarial-robust deep neural network whose predictions are locally invariant in a small neighborhood of its inputs. Existing empirical defense methods formulate the adversarial training as a min-max optimization problem~\citep{Madry18PGD}. Since a large number of adversarial examples need to be generated during training, practitioners pay great attention to the time cost of adversarial example generation. Compared with AA, MM attack contributes a feasible and reliable method to generate high-quality adversarial examples in AT. Table~\ref{table:exp_AT} in Appendix~\ref{app:advtraining} reports the performance. MM3-F10 (MM3-F20) denotes AT using adversarial examples generated by MM attack with 10 (20) fixed steps. MM3 denotes AT using adversarial examples generated by MM attack with 20 adaptive steps. We choose $11$ testing methods to evaluate the robustness. As shown in Table~\ref{table:exp_AT} (Appendix~\ref{app:advtraining}), by replacing PGD-generated adversarial examples with MM-attack-generated adversarial examples, the robustness of AT model is significantly improved. %Hence, our method contributes a feasible and reliable method to generate high-quality adversarial examples in adversarial training.

%To the best of our knowledge, our MM attack is current \emph{SOTA benchmark}, which provides a new direction of evaluating adversarial robustness and contributes a feasible and reliable method to generate high-quality adversarial examples in adversarial training.

\section{Conclusion}
%Fast and reliable evaluation of adversarial robustness is of great significance for practitioners with limited computational resources.
In this work, we proposed MM attack, which can reliably and efficiently evaluate adversarial robustness. For its reliability, we identified \emph{minimum margin} as the key evaluation criterion for the most adversarial example. For its computational efficiency, we proposed an effective STARS method to ensure that its computational time is independent of the number of classes. Our experiments showed that MM attack achieves comparable performance compared with AA, but only costs 3\% of the computational time. Its reliability and efficiency further allow us to extend MM attack into AT, which significantly improves the quality of adversarial examples in AT and thus boosts the performance of AT.

\section*{Acknowledgements}
RZG, JXW, KWZ, BHX and JC were supported by GRF 14208318 from the RGC of HKSAR. BH was supported by the RGC Early Career Scheme No. 22200720, NSFC Young Scientists Fund No. 62006202, and Guangdong Basic and Applied Basic Research Foundation No. 2022A1515011652. GN were supported by JST AIP Acceleration Research Grant Number JPMJCR20U3, Japan. 
%would like to thank for productive discussions.

\clearpage

\bibliography{iclr2022_conference}
\bibliographystyle{icml2022}
\clearpage

%%%%%%%%%%%%%%%%%%%%%%%%%%%%%%%%%%%%%%%%%%%%%%%%%%%%%%%%%%%%%%%%%%%%%%%%%%%%%%%
%%%%%%%%%%%%%%%%%%%%%%%%%%%%%%%%%%%%%%%%%%%%%%%%%%%%%%%%%%%%%%%%%%%%%%%%%%%%%%%
% APPENDIX
%%%%%%%%%%%%%%%%%%%%%%%%%%%%%%%%%%%%%%%%%%%%%%%%%%%%%%%%%%%%%%%%%%%%%%%%%%%%%%%
%%%%%%%%%%%%%%%%%%%%%%%%%%%%%%%%%%%%%%%%%%%%%%%%%%%%%%%%%%%%%%%%%%%%%%%%%%%%%%%
\newpage
\appendix
\onecolumn

\section{The Computational Time in the Worst Case}
\label{app_bound}

In this section, we discuss the computational time in the worst case. According to their attack mechanism as an ensemble of diverse attacks, AA and T-AA consider one attack first. If the attack succeeds, stop other attacks on the current example; else, continue to consider the next attack in the ensemble. According to the strategy of our STARS method, MM attack considers the false target with the largest predicted probability first, if the attack succeeds, stop attacks on other false targets; else, continue to consider the next target in the ranking of the predicted probability. The computational time of these methods is influenced by different datasets and models. Hence, in the worst case that all attacks inside fail to succeed, the computational time is the sum of the individual time of each attack. Hence, the computational cost of AA (or T-AA) is 109 times (or 440 times ) more than PGD, and 34 times (or 139 times) more than MM3 in this case.

%The reliable evaluation of adversarial robustness is of great help to the development of defense strategies. However, the field is currently in a delimma that reliability and efficiency cannot be achieved simultaneously, e.g., AutoAttack has good performance but high computational cost; PGD has low computational cost but poor performance. This paper provides a new perspective that Minimum Probabilistic-margin attack to achieve a reliable evaluation with a handful of computational costs. Experimental results show that our proposed MP-attack not only means a lot for the evaluation of adversarial robustness but also contributes to defense strategies. In a long term, Minimum Probabilistic-margin provides insights on how to design the reliable evaluation rather than sacrificing computational cost in pursuit of diverse attacks.

\section{The Realization of Adversarial Training of MM Attack}
We summarize the adversarial training of MM Attack in Algorithm~\ref{alg:MM-AT}. We use MM3 attack to generate adversarial examples, and the computational time is about 2 times as much as PGD \citep{Madry18PGD}, which can be acceptable for most practitioners.

\section{Potential Benefits of Diverse Rescalings}
\label{app:diverse}

We investigate the difference among different successful sets of seven rescaling methods mentioned above. In Table~\ref{table:normform_new}, the setting follows \citep{Madry18PGD} (with $20$ fixed steps). The non-empty difference sets $A \cup B_i - A$ and $A \cup B_i - B_i$ suggest that diverse rescaling methods can complement each other. Hence, when considerable computational resources are available, we recommend practitioners to consider diverse logits rescaling on a strong attack (e.g., our MM attack) rather than diverse weak attacks. Note that we do not argue that diverse weak attacks is unnecessary but rather that when a reliable enough attack exists, most relatively weak attacks have limited benefits other than increased computational cost.

\section{The Replacement of Natural Data for the Ranking in STARS}
\label{app:ranking}
In our STARS method, we also investigate the difference of replacing the natural input $x$ with adversarial examples. Table~\ref{table:advpres} shows that the replacement has limited improvements.

\section{Detailed Experimental Results}
\label{app:exptable}
To verify the rationality of minimum margin, we conduct experiments on different step size, different step number and different $\epsball[x]$ in Table~\ref{table:exp_step} and Table~\ref{table:exp_epsilon}. We compare the reliability and the computational time between MM attacks and baselines. In Table~\ref{app:exp_acc} and Table~\ref{app:exp_time}, unless specified, the model structure is ResNet-18. The experiments verify that our MM attack achieves comparable performance but only incurs a very small amount of computational time.

\section{Experimental Results on Diverse Trained Model in RobustBench}
\label{app:robustbench}
\begin{figure*}[!h]
    \begin{center}
        \subfigure[Evaluation on \citet{sehwag2021improving}]
        {\includegraphics[width=0.495\textwidth]{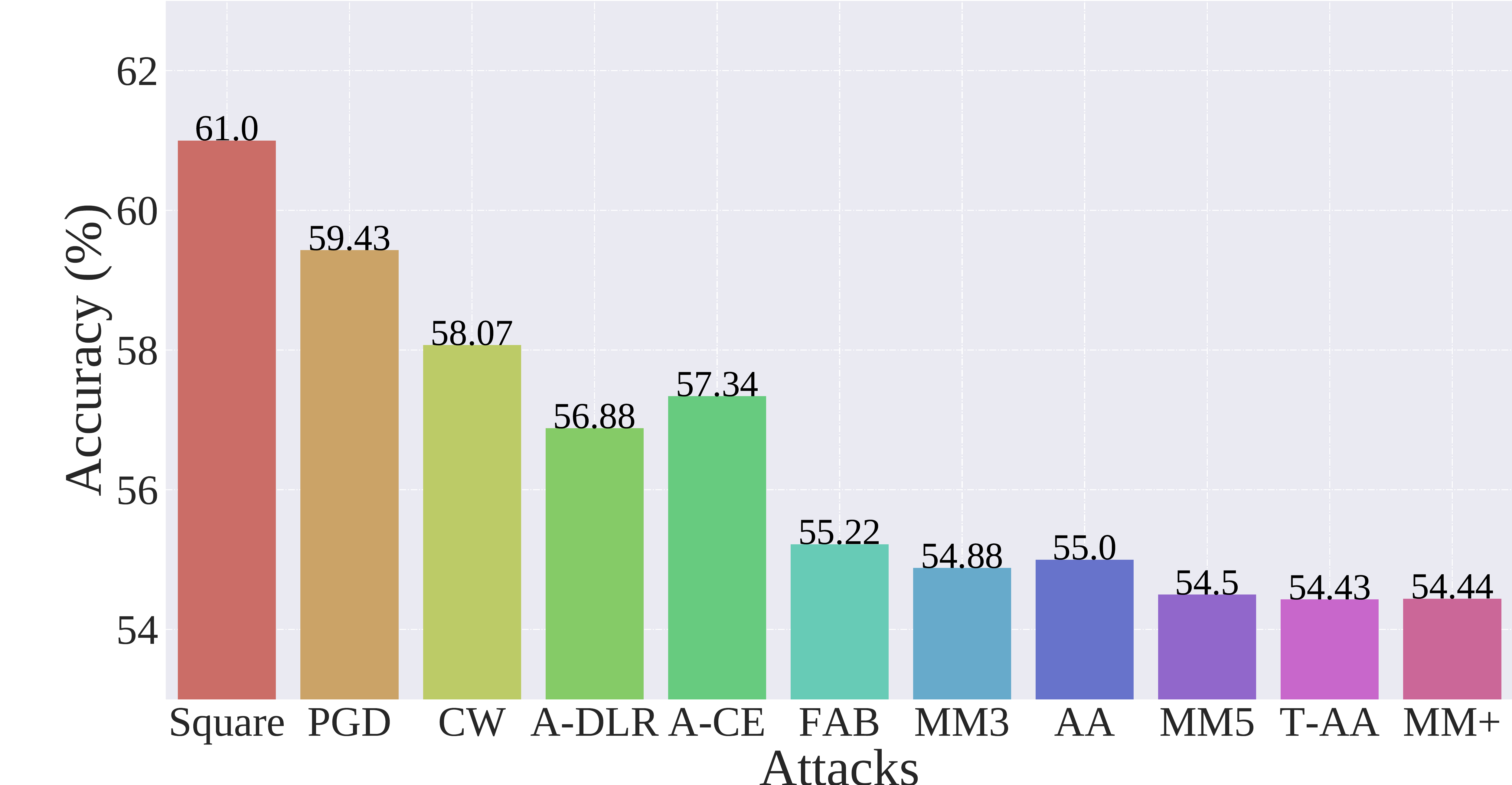}}
        \subfigure[Computational time on \citet{sehwag2021improving}]
        {\includegraphics[width=0.495\textwidth]{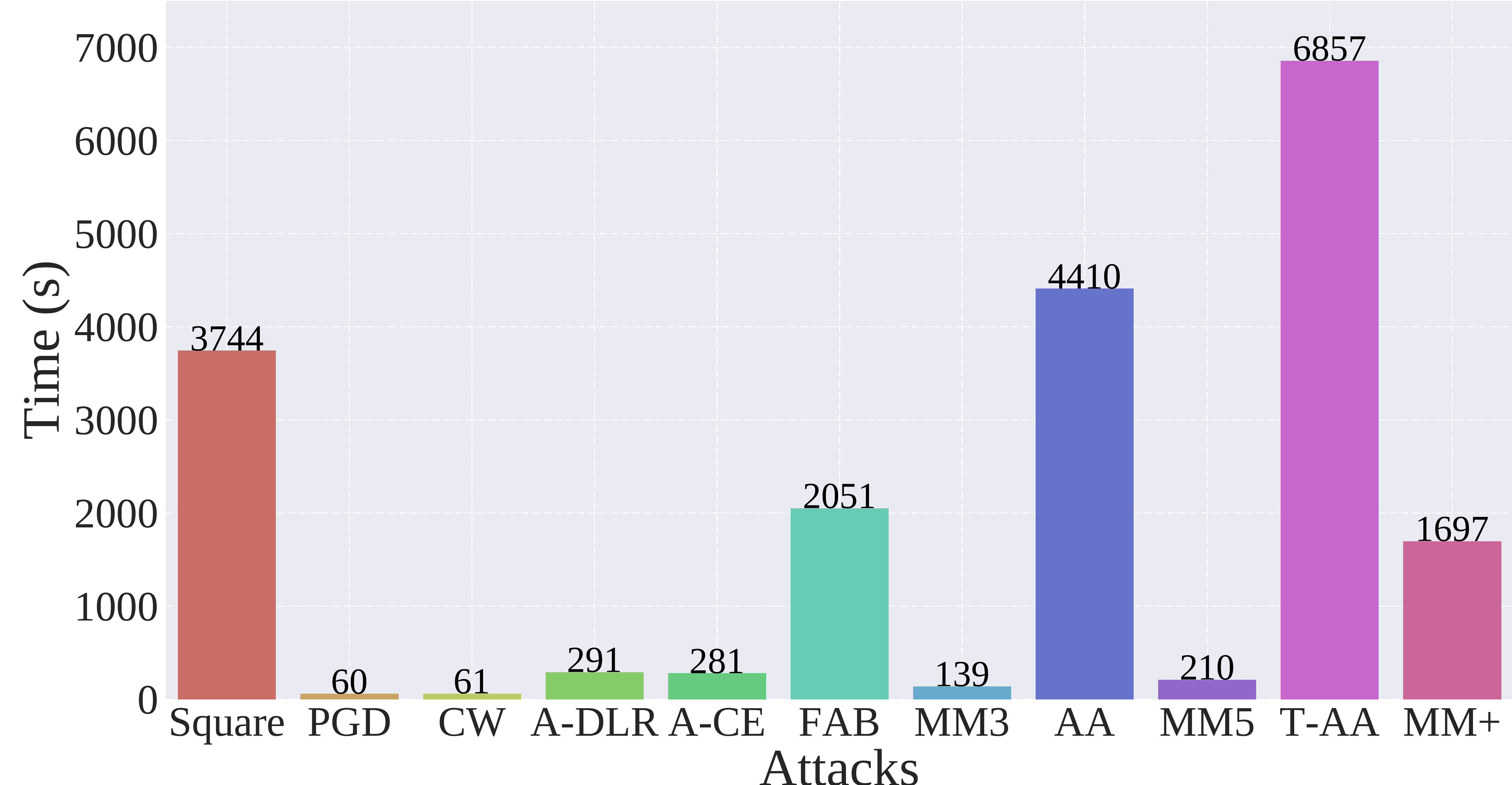}}
        \subfigure[Evaluation on \citet{rade2021helper}]
        {\includegraphics[width=0.495\textwidth]{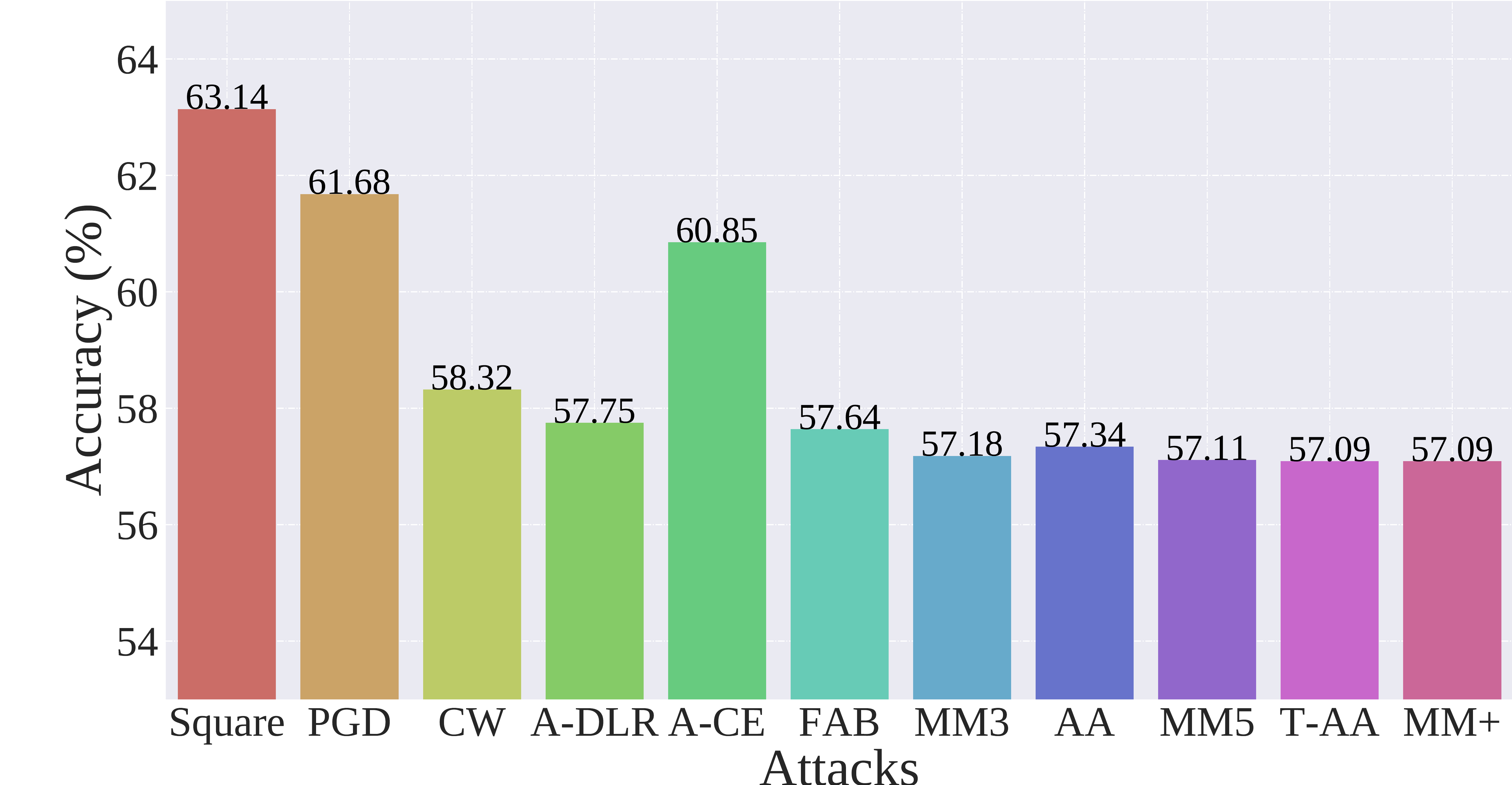}}
        \subfigure[Computational time on \citet{rade2021helper}]
        {\includegraphics[width=0.495\textwidth]{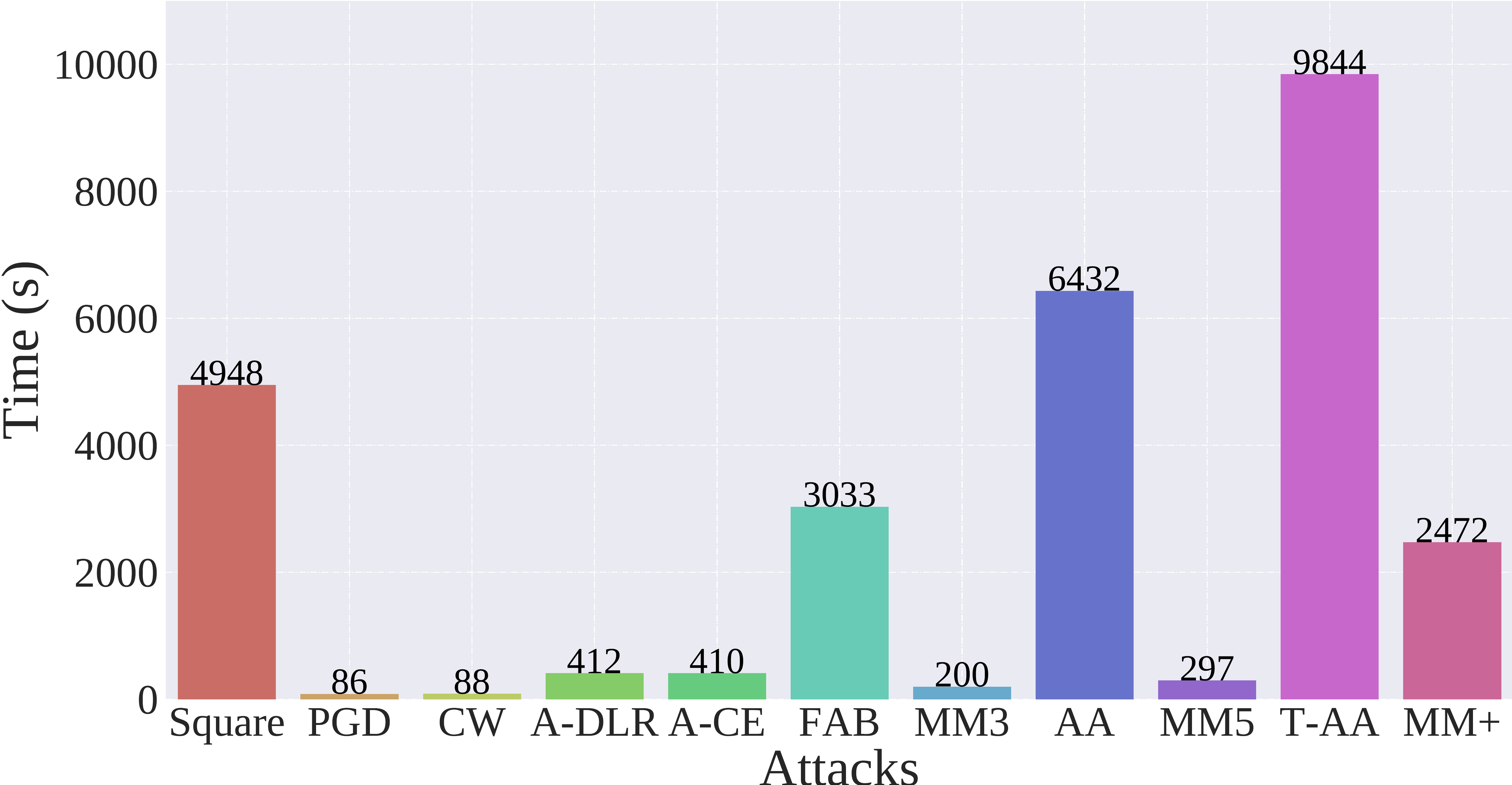}}
        \subfigure[Evaluation on \citet{rebuffi2021fixing}]
        {\includegraphics[width=0.495\textwidth]{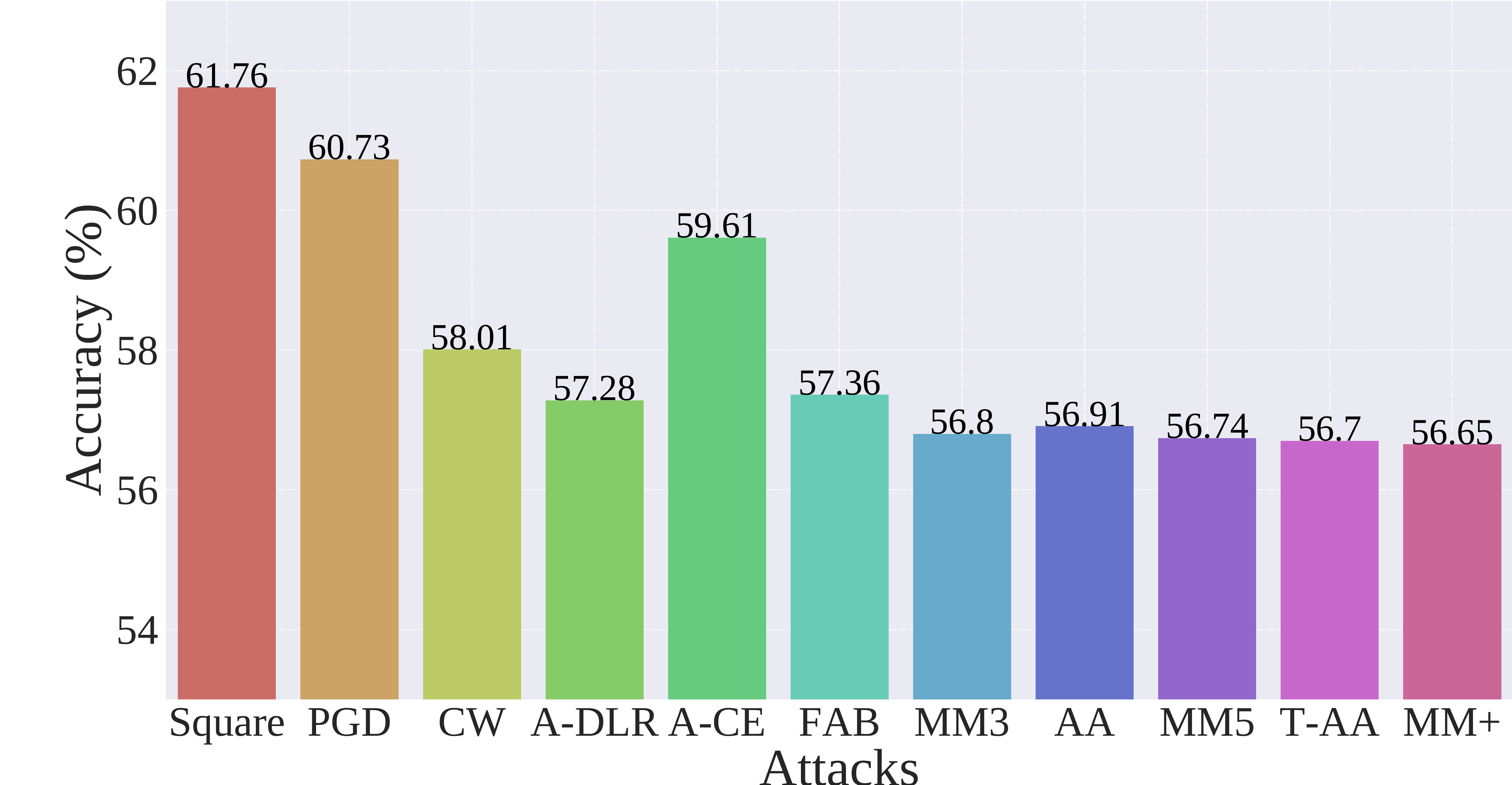}}
        \subfigure[Computational time on \citet{rebuffi2021fixing}]
        {\includegraphics[width=0.495\textwidth]{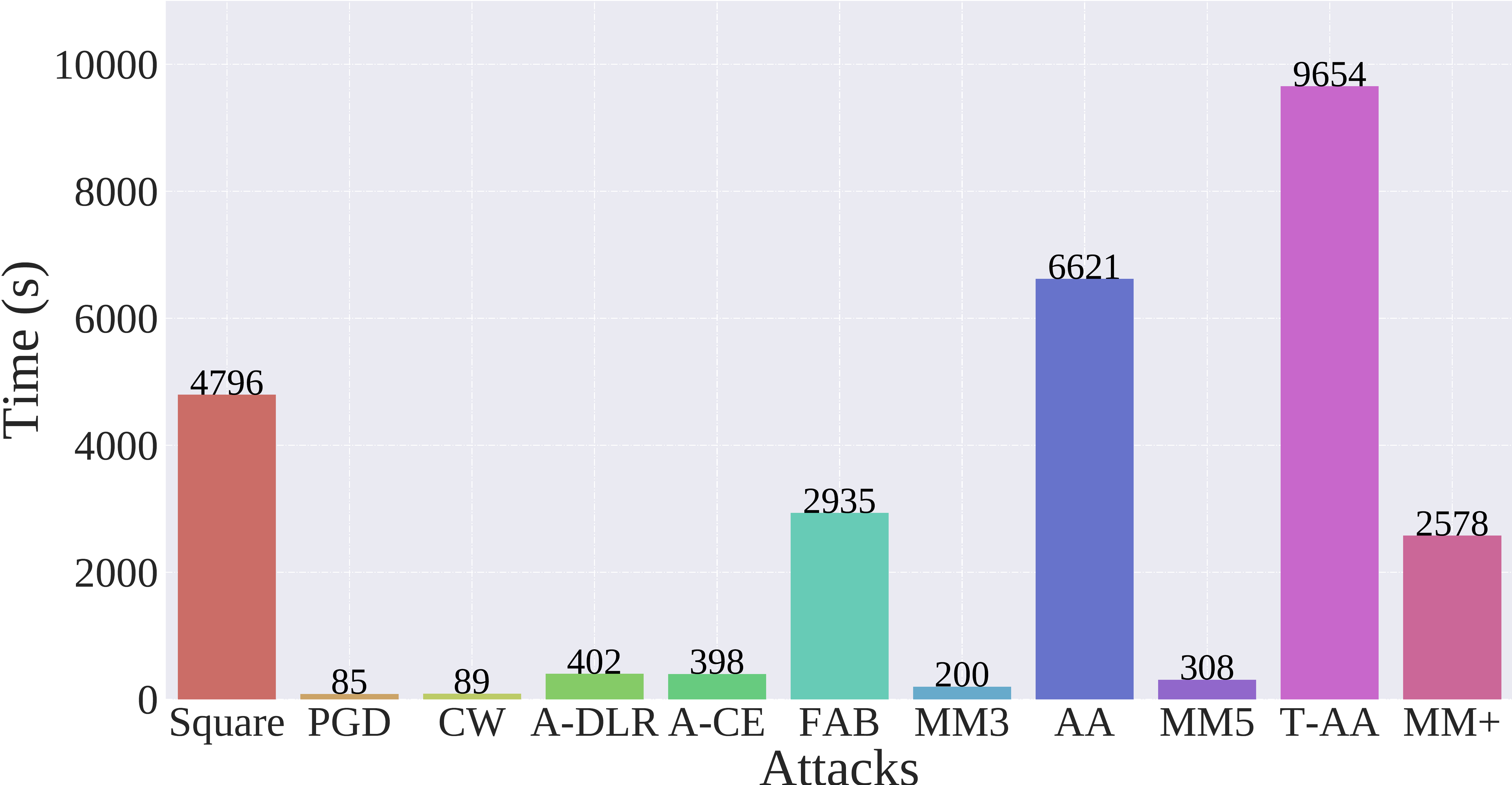}}
        \subfigure[Evaluation on \citet{andriushchenko2020understanding}]
        {\includegraphics[width=0.495\textwidth]{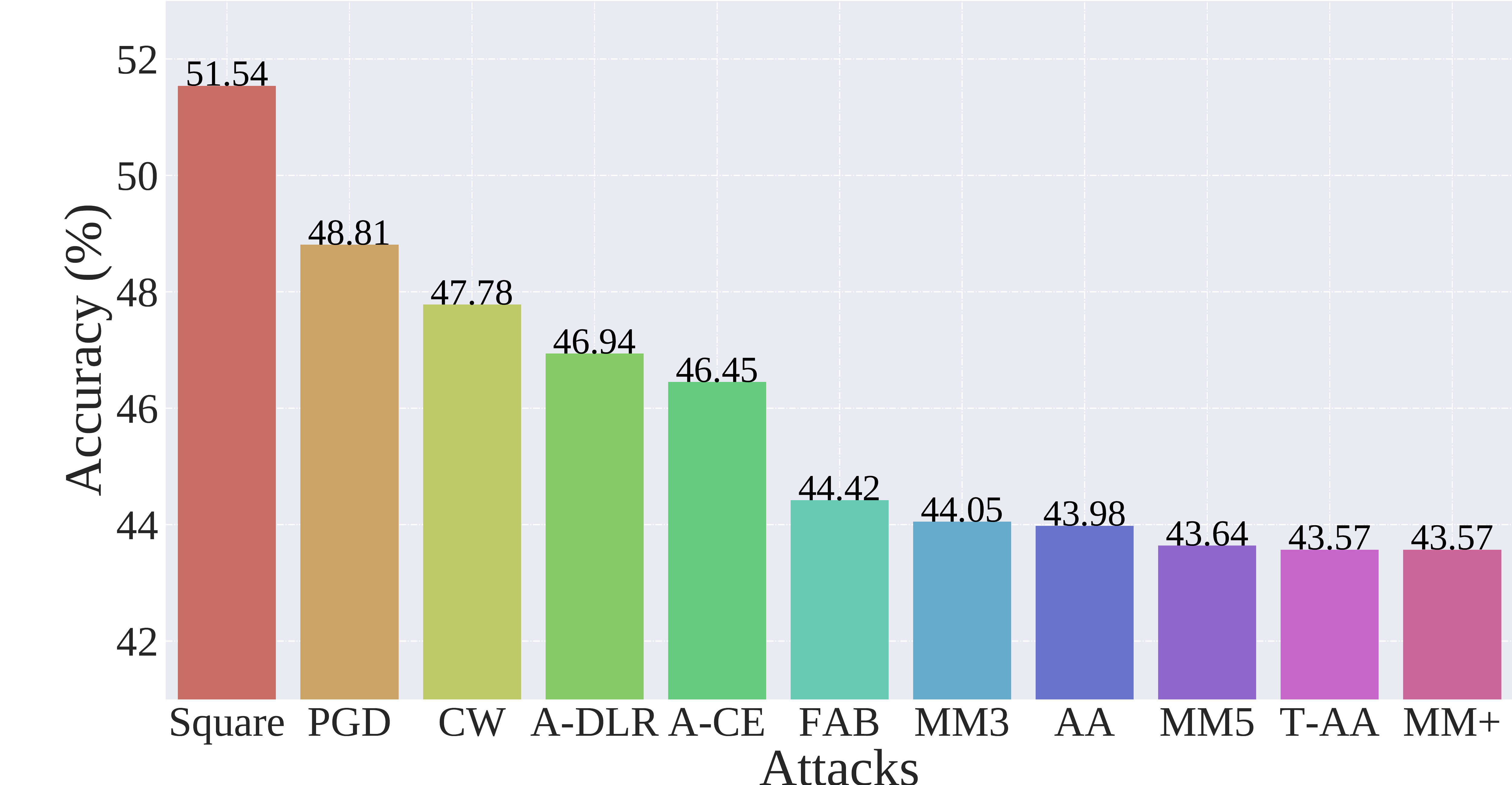}}
        \subfigure[Computational time on \citet{andriushchenko2020understanding}]
        {\includegraphics[width=0.495\textwidth]{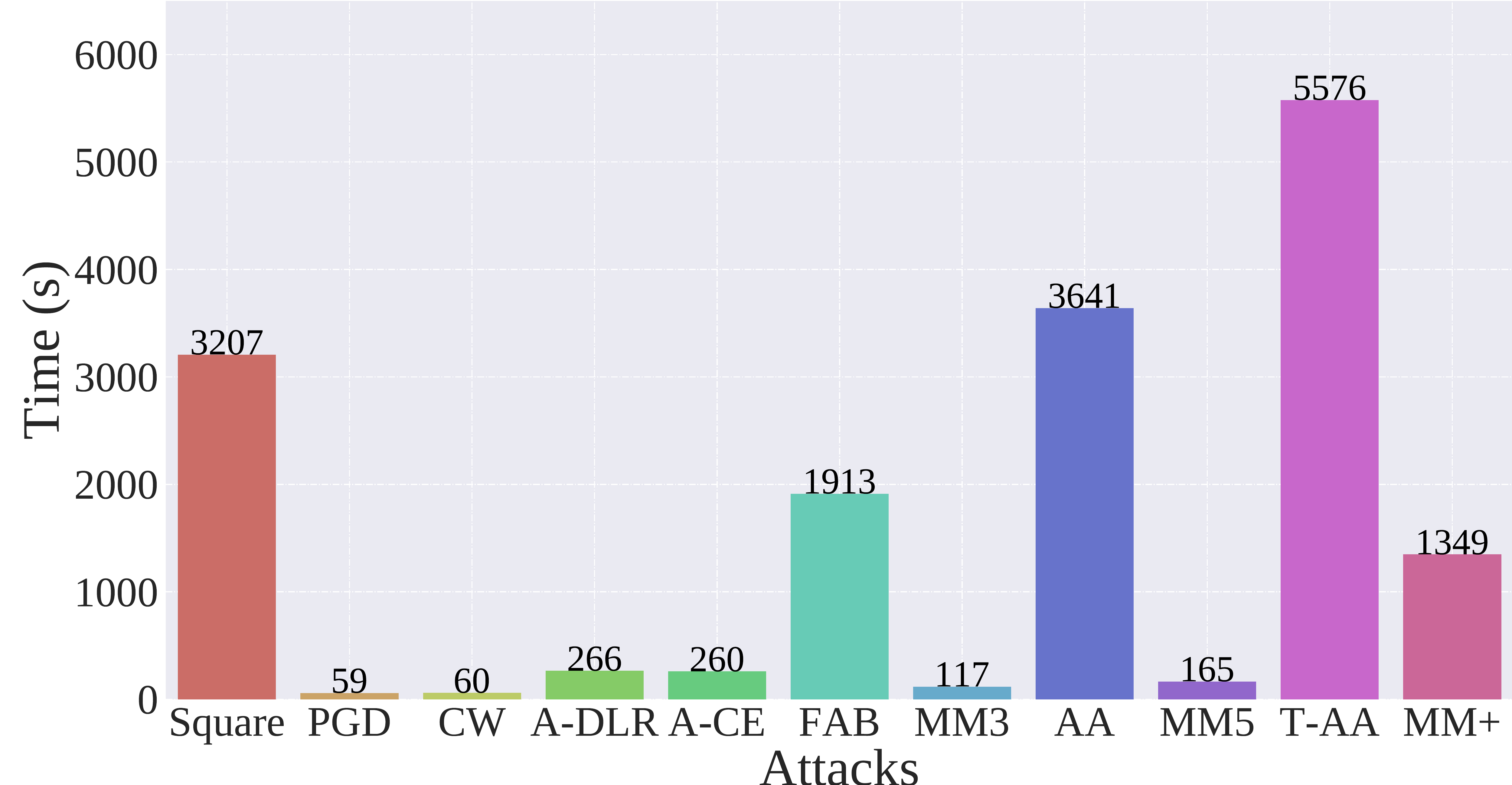}}
        % \vspace{-1em}
        \caption{\footnotesize Comparison of reliability and computational cost on different defense in RobustBench. We compare three versions of our MM attack (MM3, MM5 and MM+ mentioned in Section~\ref{exp:baselines}) with 8 baselines. In subfigure (a), (c),(e) and (g), the Y-axis is the accuracy of the attacked model, which means that the lower the accuracy, the stronger the attack (or to say the better evaluation). In subfigure (b), (d), (f) and (h), the Y-axis is computational time, which means the less the time, the higher the computational efficiency. Experiments are on CIFAR-10 with $L_{\infty}$-norm bounded perturbation.
        %In the subfigure (b), the gray shape is a hypothetical distribution of all adversarial variants maps within the bounded perturbation epsilon on a natural example; $\mathbb{P}_t$ and $\mathbb{P}_y$ are the predicted probability on a targeted false label $t$ and the true label $y$; The orange area ($\mathbb{P}_t > \mathbb{P}_y$) indicates that the adversarial variants inside can be misclassified, or to say attack successfully, while the blue area ($\mathbb{P}_t < \mathbb{P}_y$) indicates that the adversarial variants inside cannot attack successfully.
        }
        % \vspace{-2em}
    \label{robustbench_exp1}
    \end{center}
    \vspace{-1em}
\end{figure*}

\begin{figure*}[!h]
    \begin{center}
        \subfigure[Evaluation on \citet{gowal2020uncovering}]
        {\includegraphics[width=0.495\textwidth]{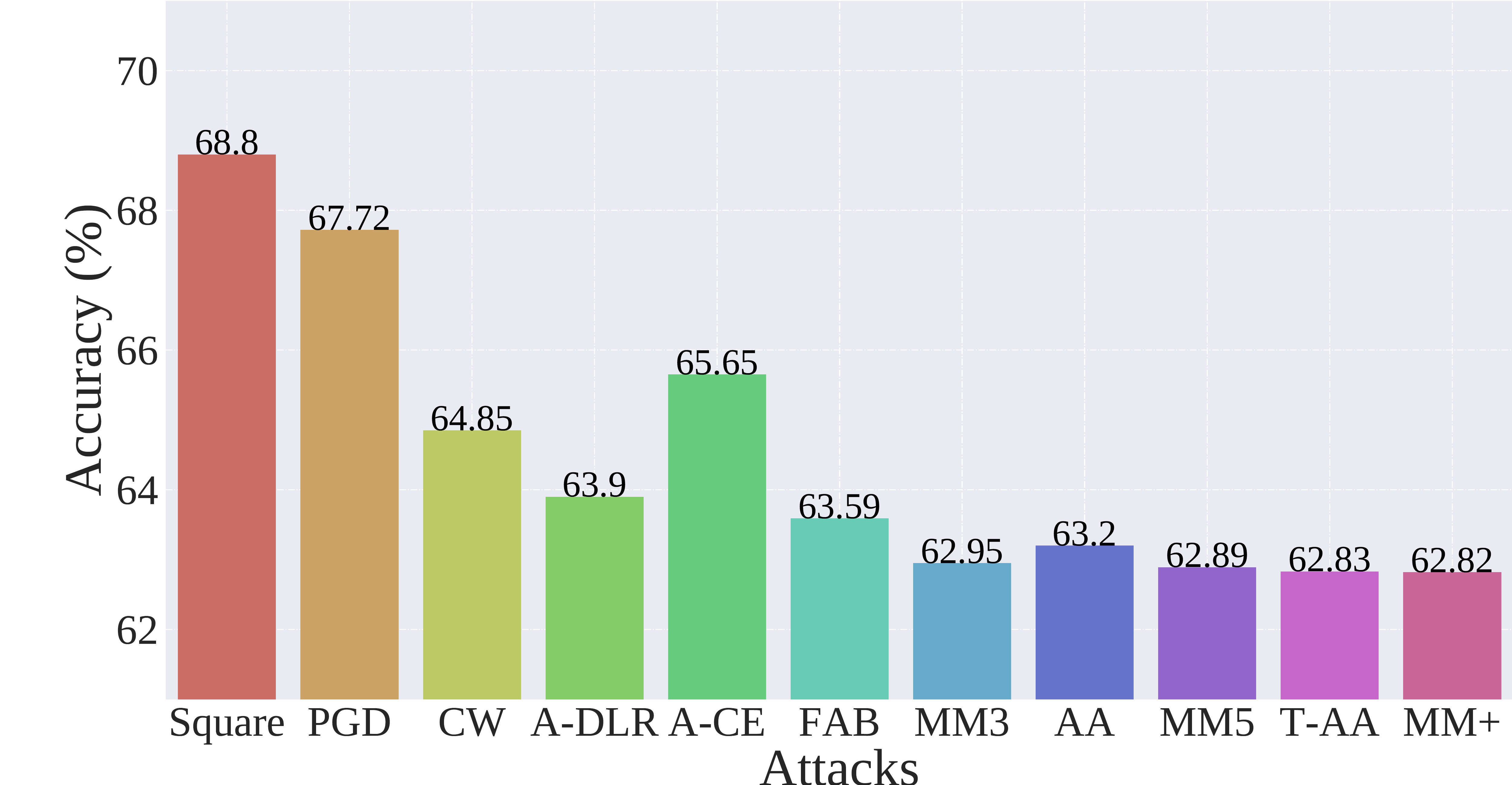}}
        \subfigure[Computational time on \citet{gowal2020uncovering}]
        {\includegraphics[width=0.495\textwidth]{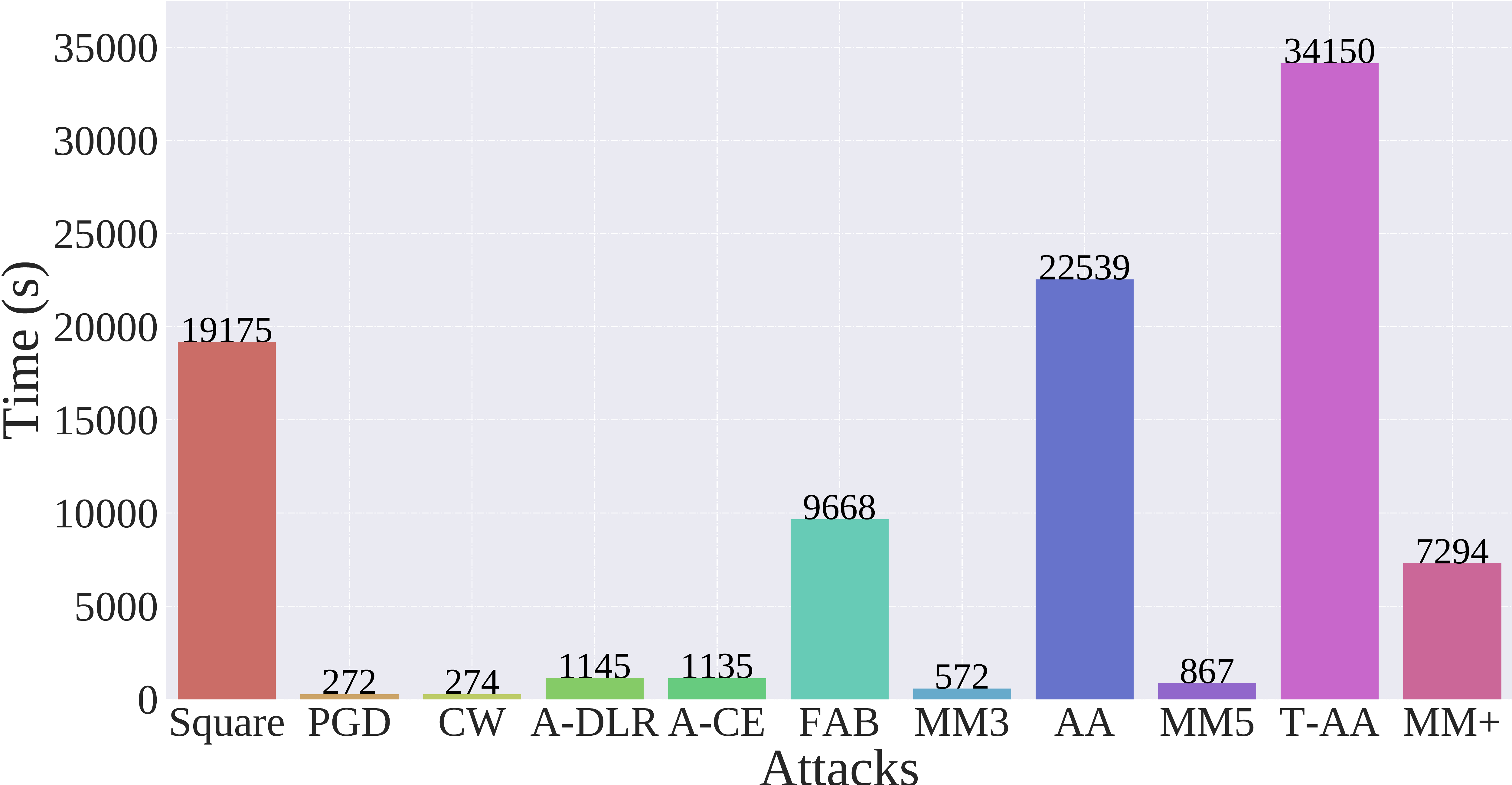}}
        \subfigure[Evaluation on \citet{sridhar2021robust}]
        {\includegraphics[width=0.495\textwidth]{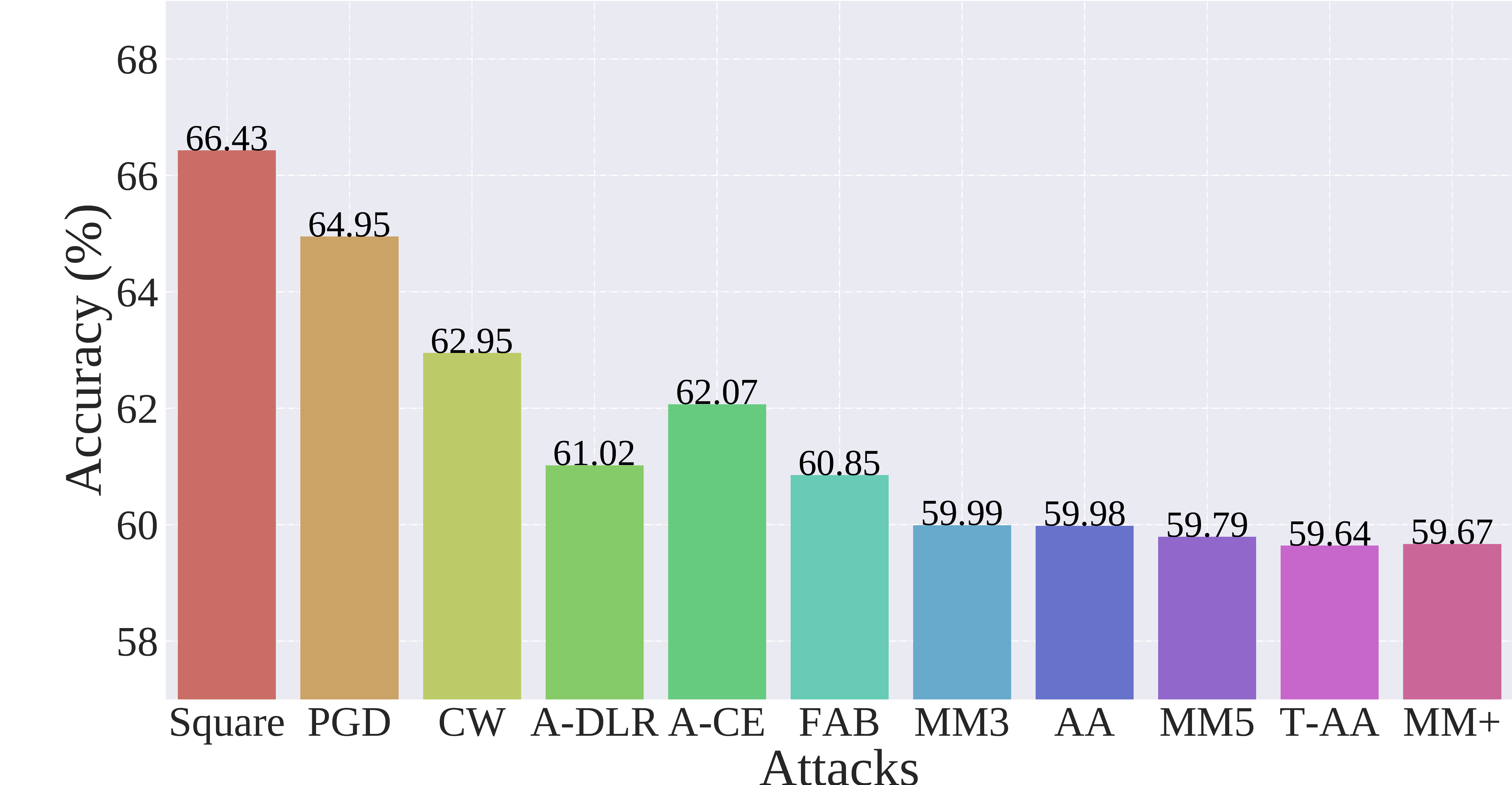}}
        \subfigure[Computational time on \citet{sridhar2021robust}]
        {\includegraphics[width=0.495\textwidth]{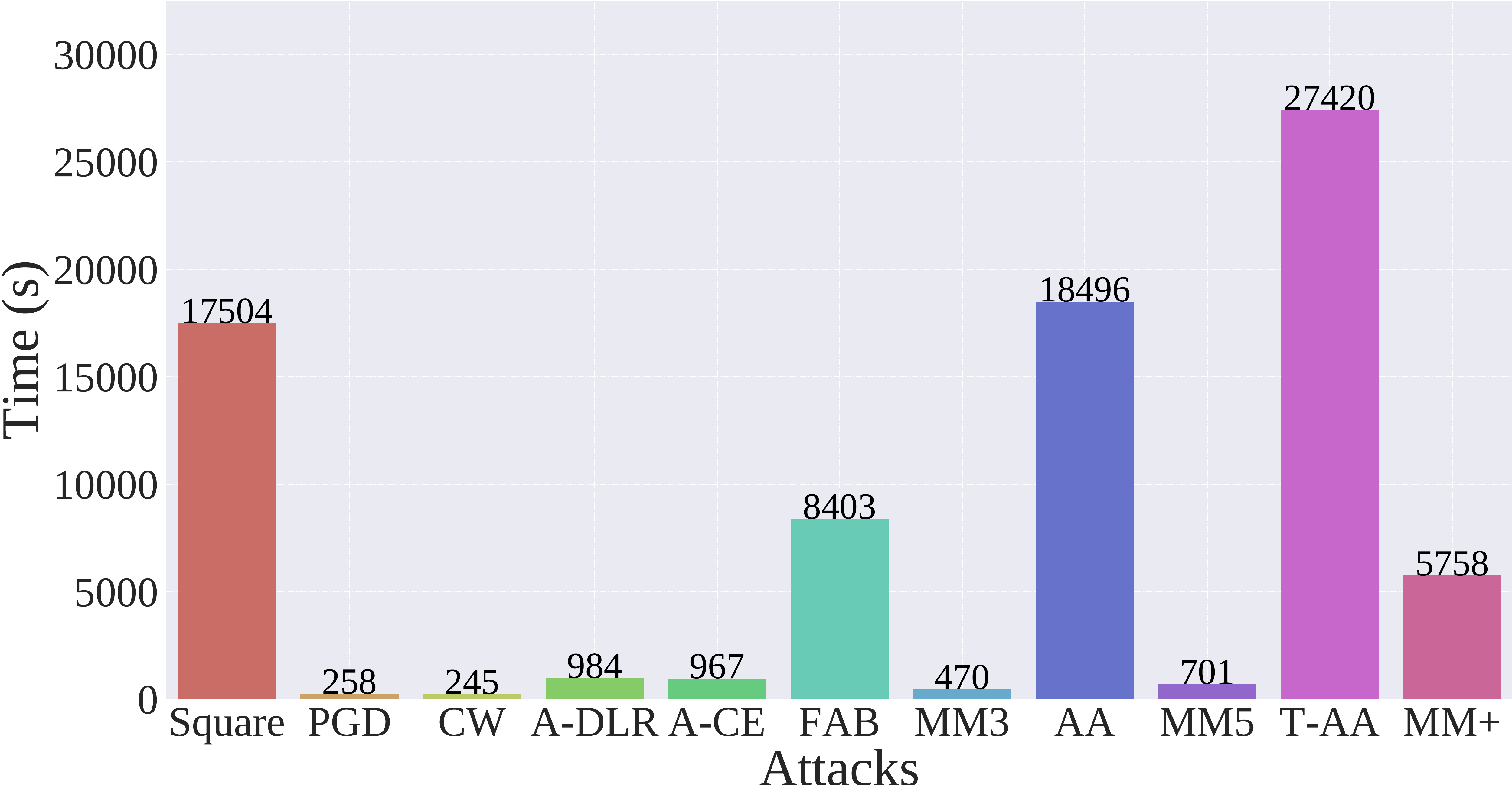}}
        \subfigure[Evaluation on \citet{wong2020fast}]
        {\includegraphics[width=0.495\textwidth]{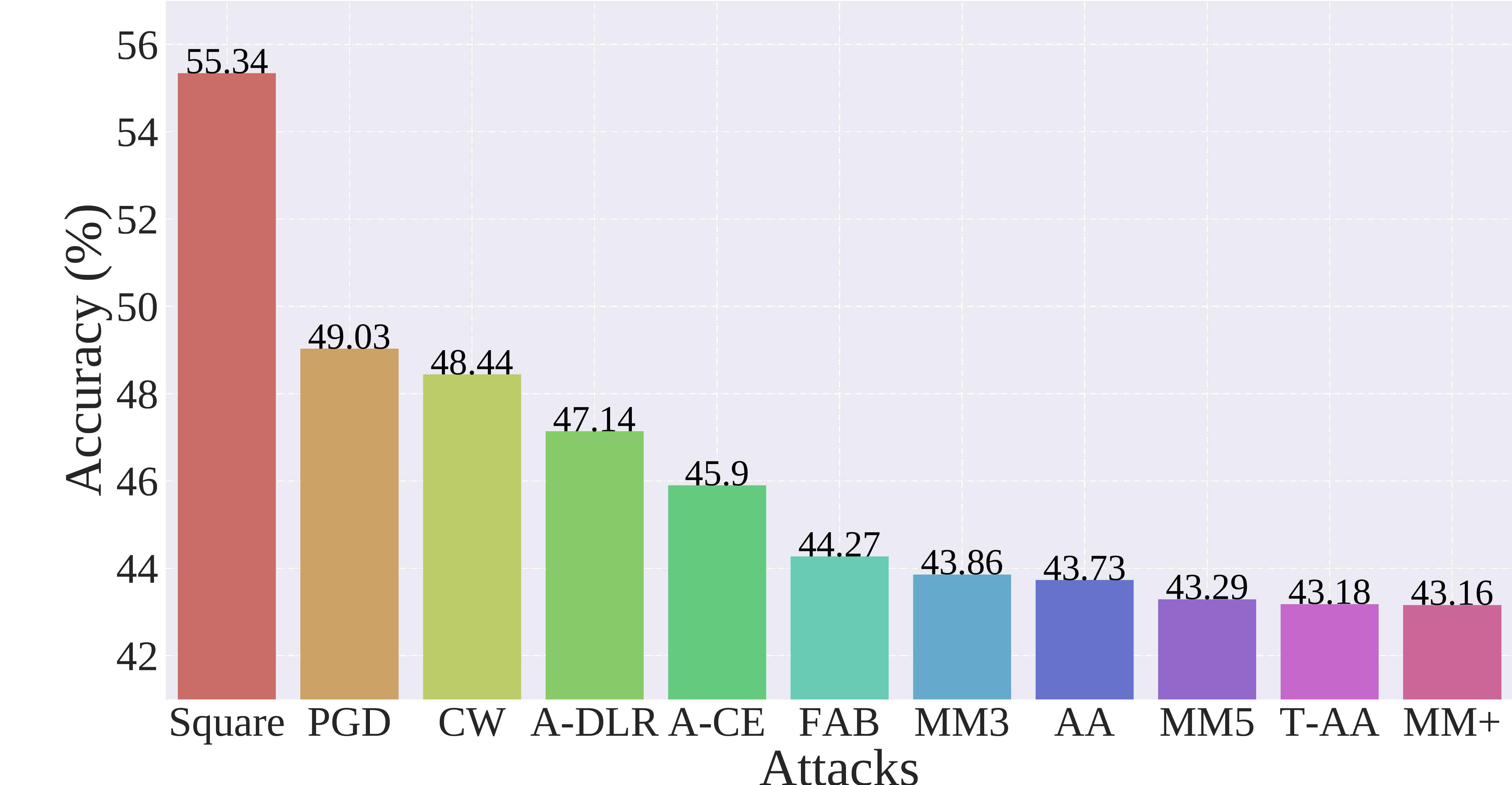}}
        \subfigure[Computational time on \citet{wong2020fast}]
        {\includegraphics[width=0.495\textwidth]{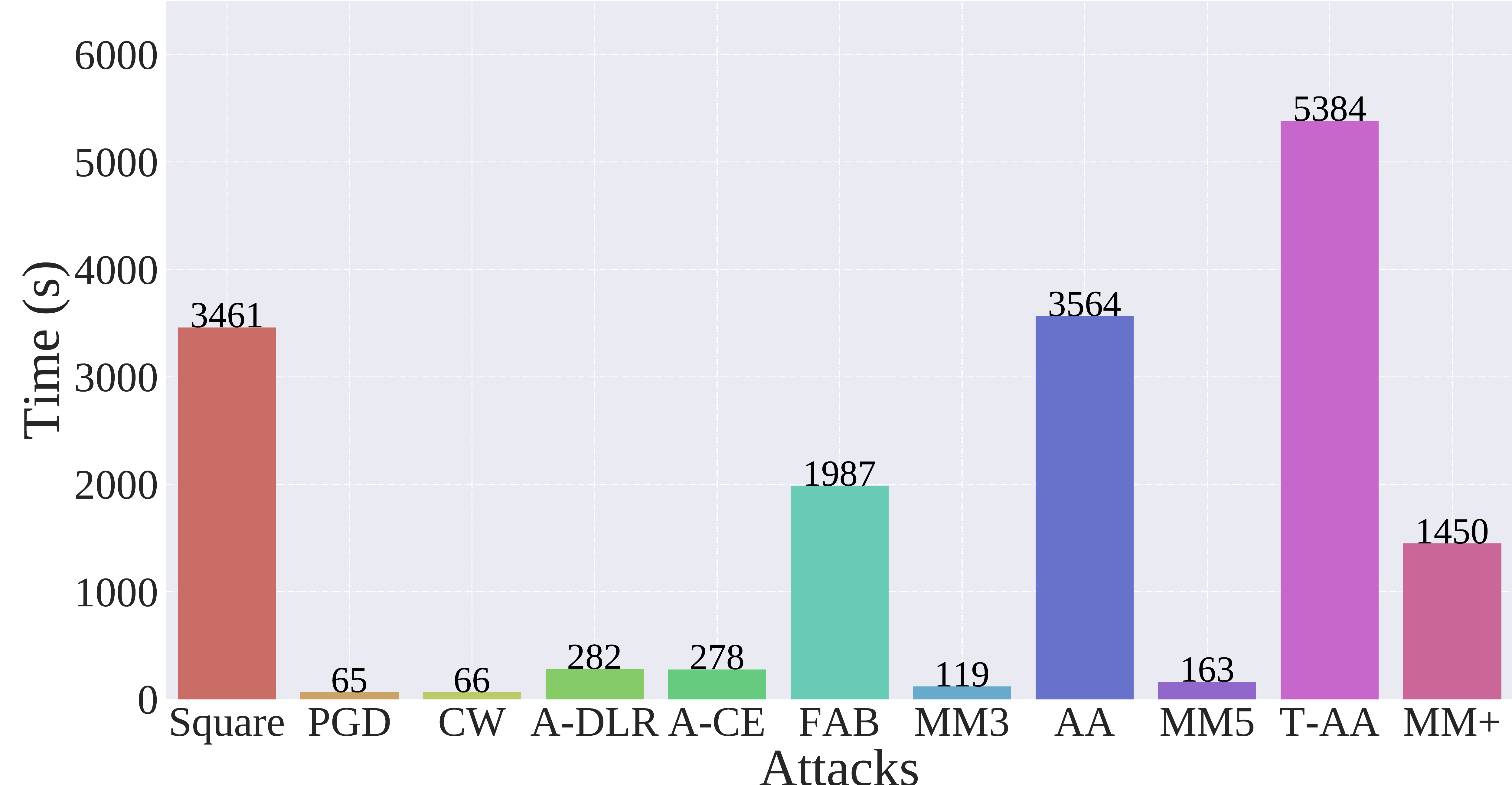}}
        \subfigure[Evaluation on \citet{robustness}]
        {\includegraphics[width=0.495\textwidth]{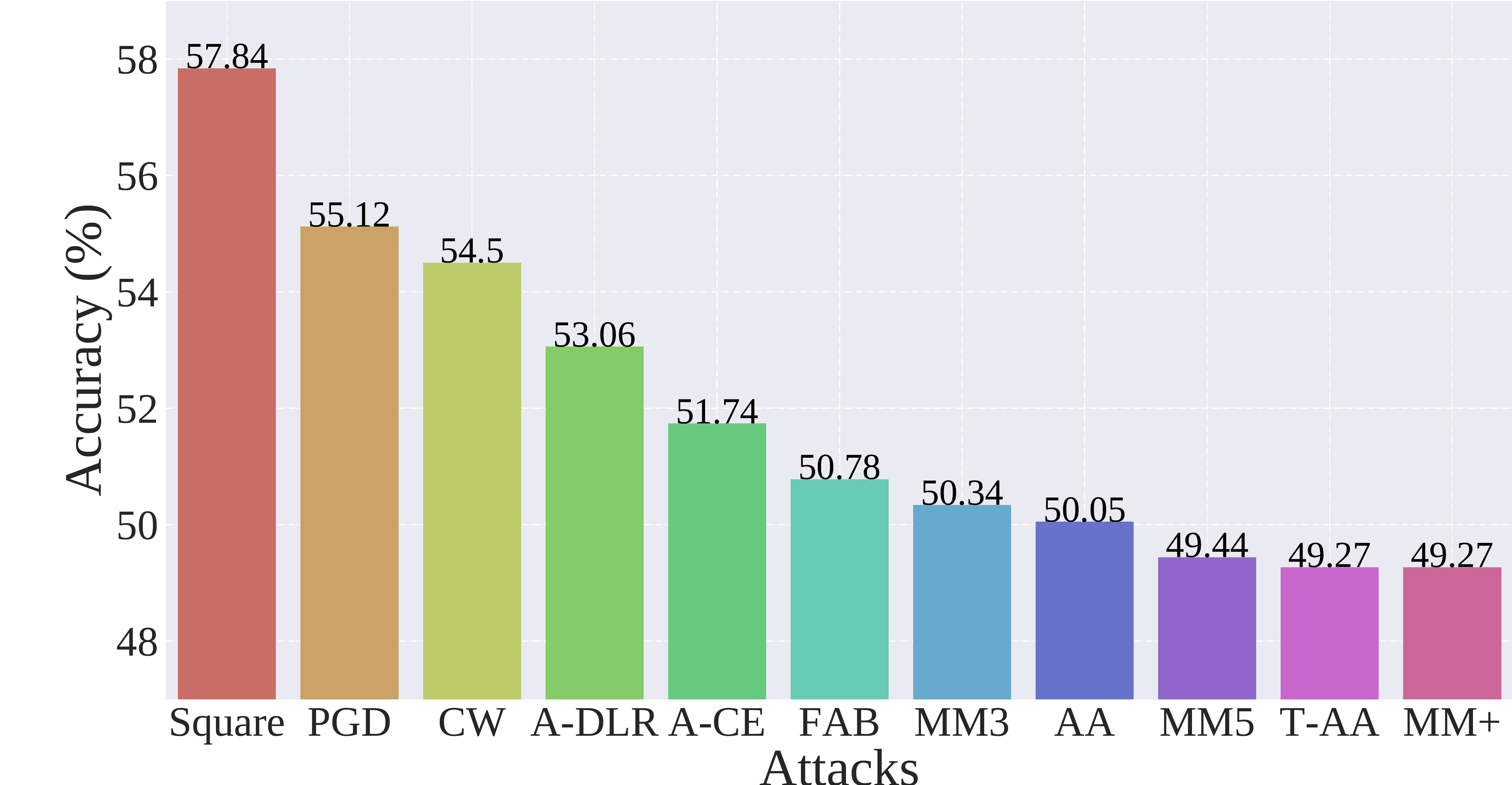}}
        \subfigure[Computational time on \citet{robustness}]
        {\includegraphics[width=0.495\textwidth]{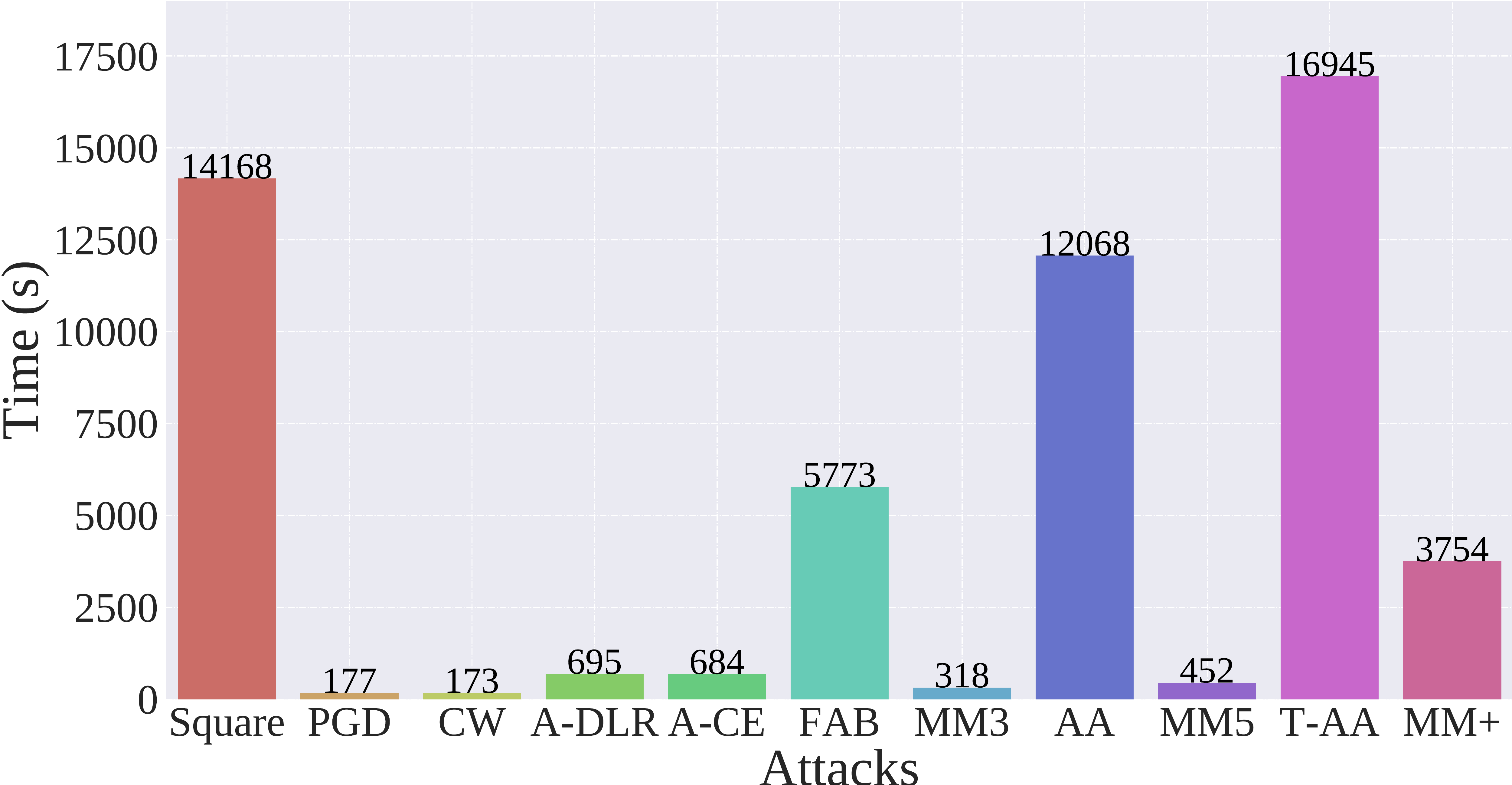}}
        % \vspace{-1em}
        \caption{\footnotesize Comparison of reliability and computational cost on different defense in RobustBench. We compare three versions of our MM attack (MM3, MM5 and MM+ mentioned in Section~\ref{exp:baselines}) with 8 baselines. In subfigure (a), (c),(e) and (g), the Y-axis is the accuracy of the attacked model, which means that the lower the accuracy, the stronger the attack (or to say the better evaluation). In subfigure (b), (d), (f) and (h), the Y-axis is computational time, which means the less the time, the higher the computational efficiency. Experiments are on CIFAR-10 with $L_{\infty}$-norm bounded perturbation. 
        %In the subfigure (b), the gray shape is a hypothetical distribution of all adversarial variants maps within the bounded perturbation epsilon on a natural example; $\mathbb{P}_t$ and $\mathbb{P}_y$ are the predicted probability on a targeted false label $t$ and the true label $y$; The orange area ($\mathbb{P}_t > \mathbb{P}_y$) indicates that the adversarial variants inside can be misclassified, or to say attack successfully, while the blue area ($\mathbb{P}_t < \mathbb{P}_y$) indicates that the adversarial variants inside cannot attack successfully.
        }
        % \vspace{-2em}
    \label{robustbench_exp2}
    \end{center}
    \vspace{-1em}
\end{figure*}

\begin{figure*}[!h]
    \begin{center}
        \subfigure[Evaluation on \citet{carmon2019unlabeled}]
        {\includegraphics[width=0.495\textwidth]{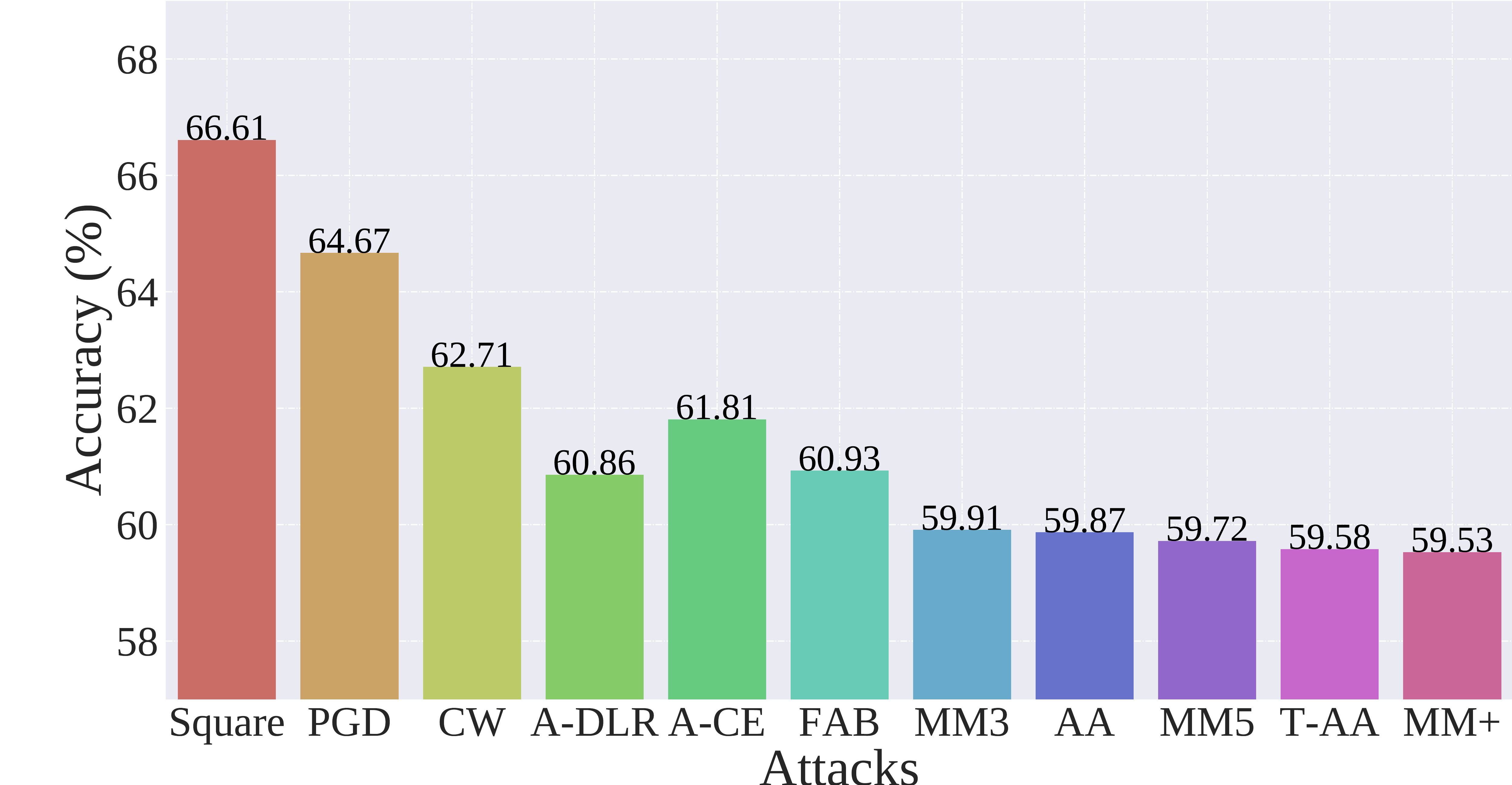}}
        \subfigure[Computational time on \citet{carmon2019unlabeled}]
        {\includegraphics[width=0.495\textwidth]{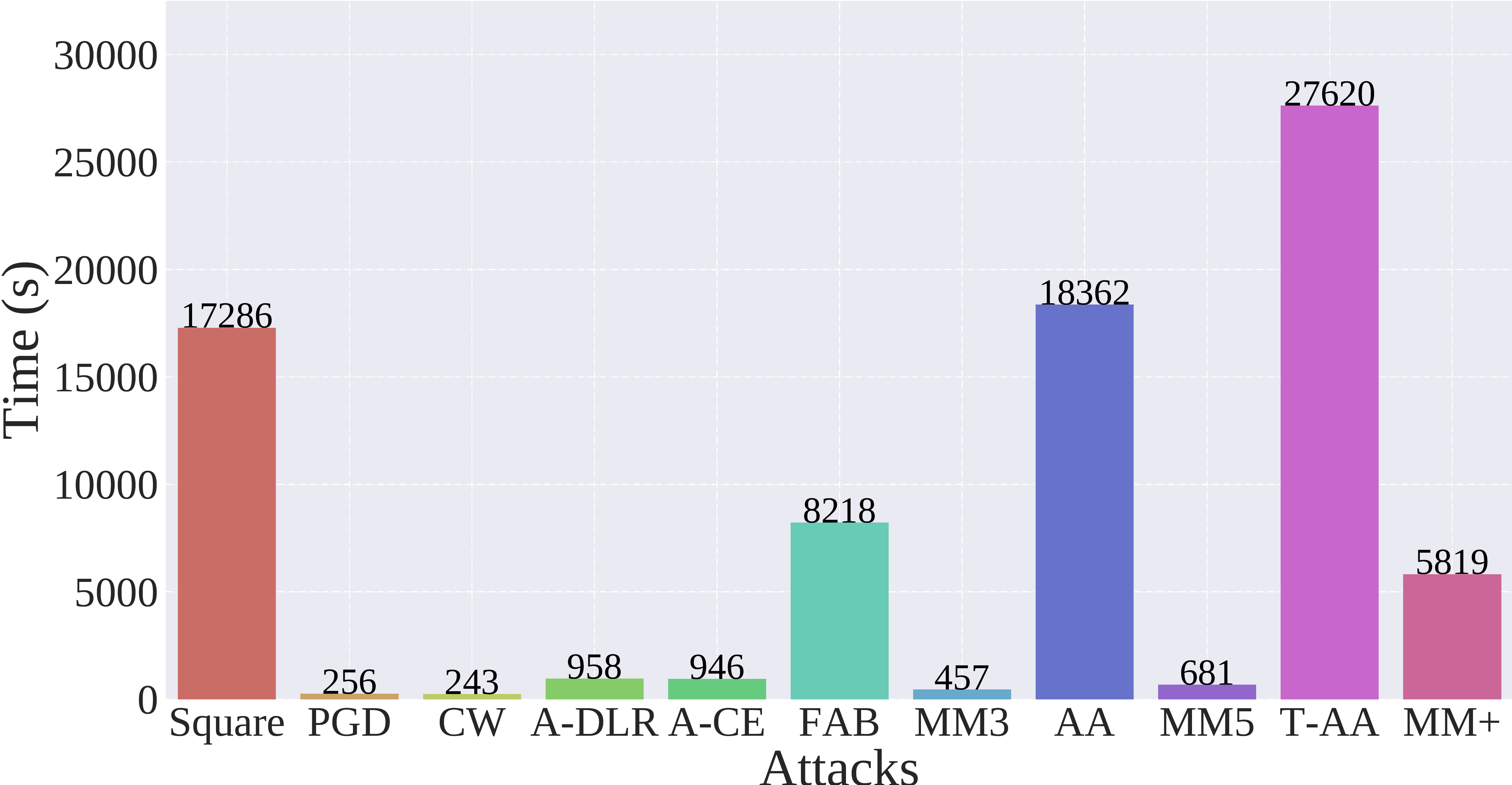}}
        \subfigure[Evaluation on \citet{wang2019improving}]
        {\includegraphics[width=0.495\textwidth]{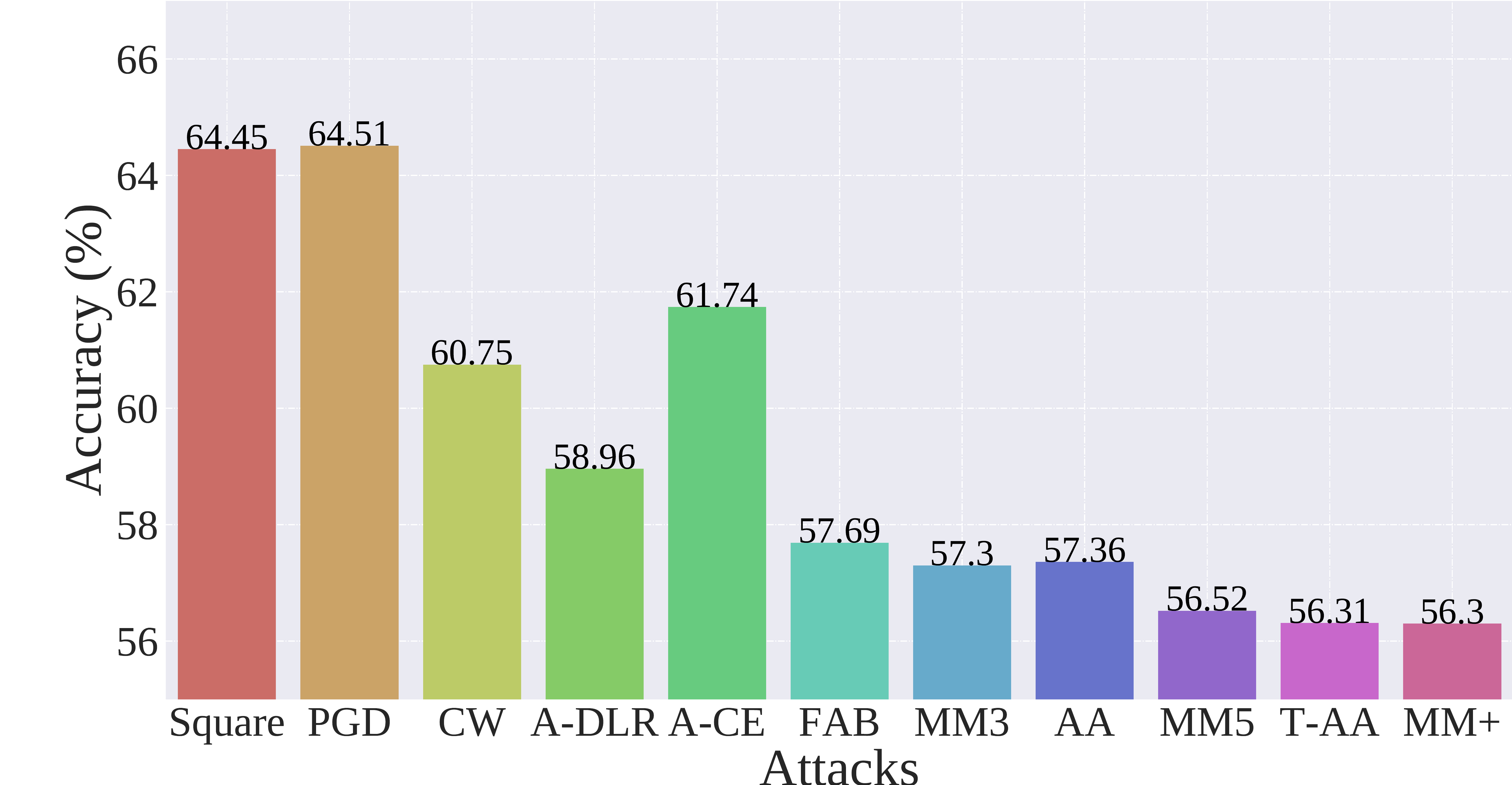}}
        \subfigure[Computational time on \citet{wang2019improving}]
        {\includegraphics[width=0.495\textwidth]{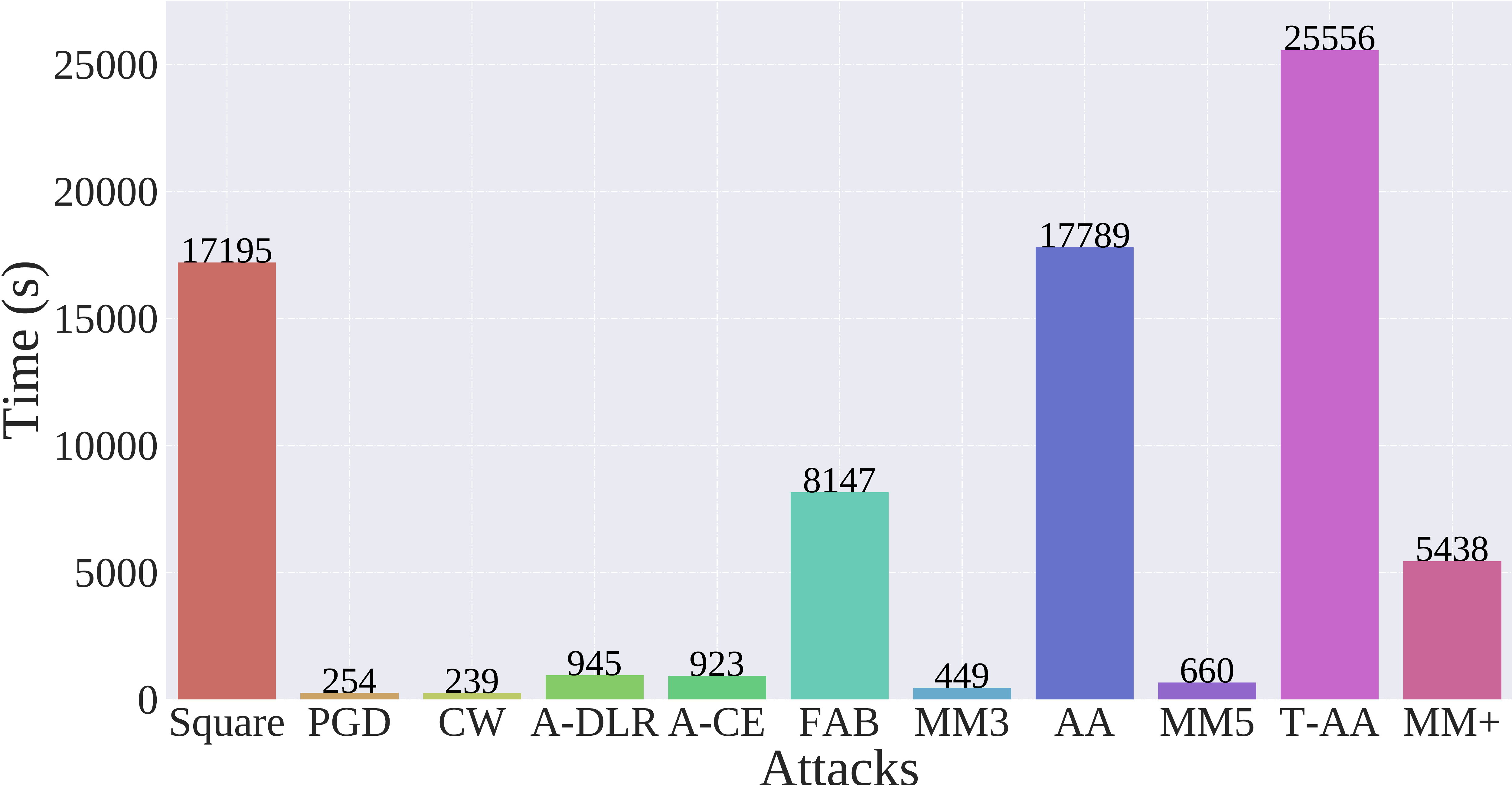}}
        \subfigure[Evaluation on \citet{wu2020adversarial}]
        {\includegraphics[width=0.495\textwidth]{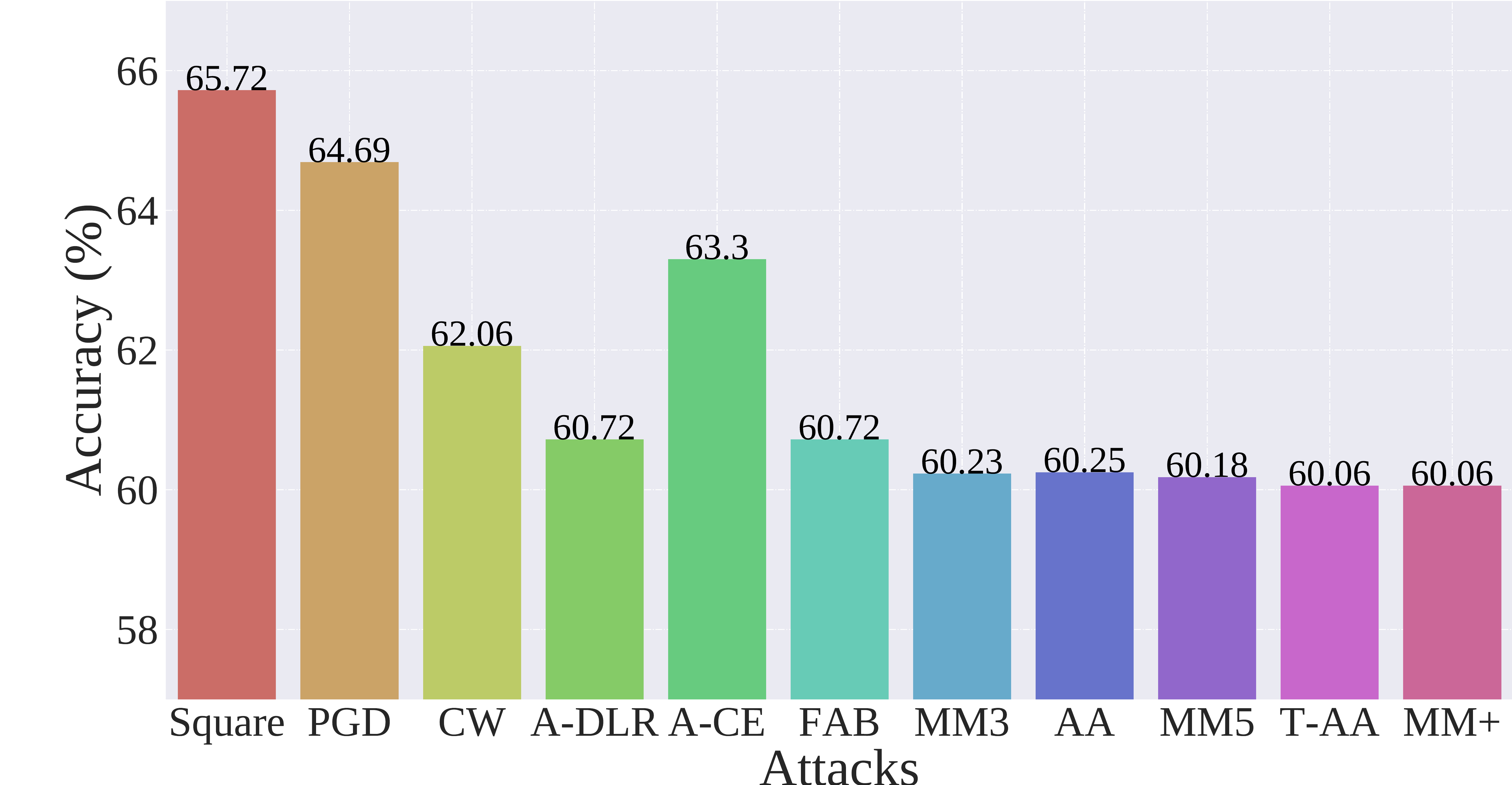}}
        \subfigure[Computational time on \citet{wu2020adversarial}]
        {\includegraphics[width=0.495\textwidth]{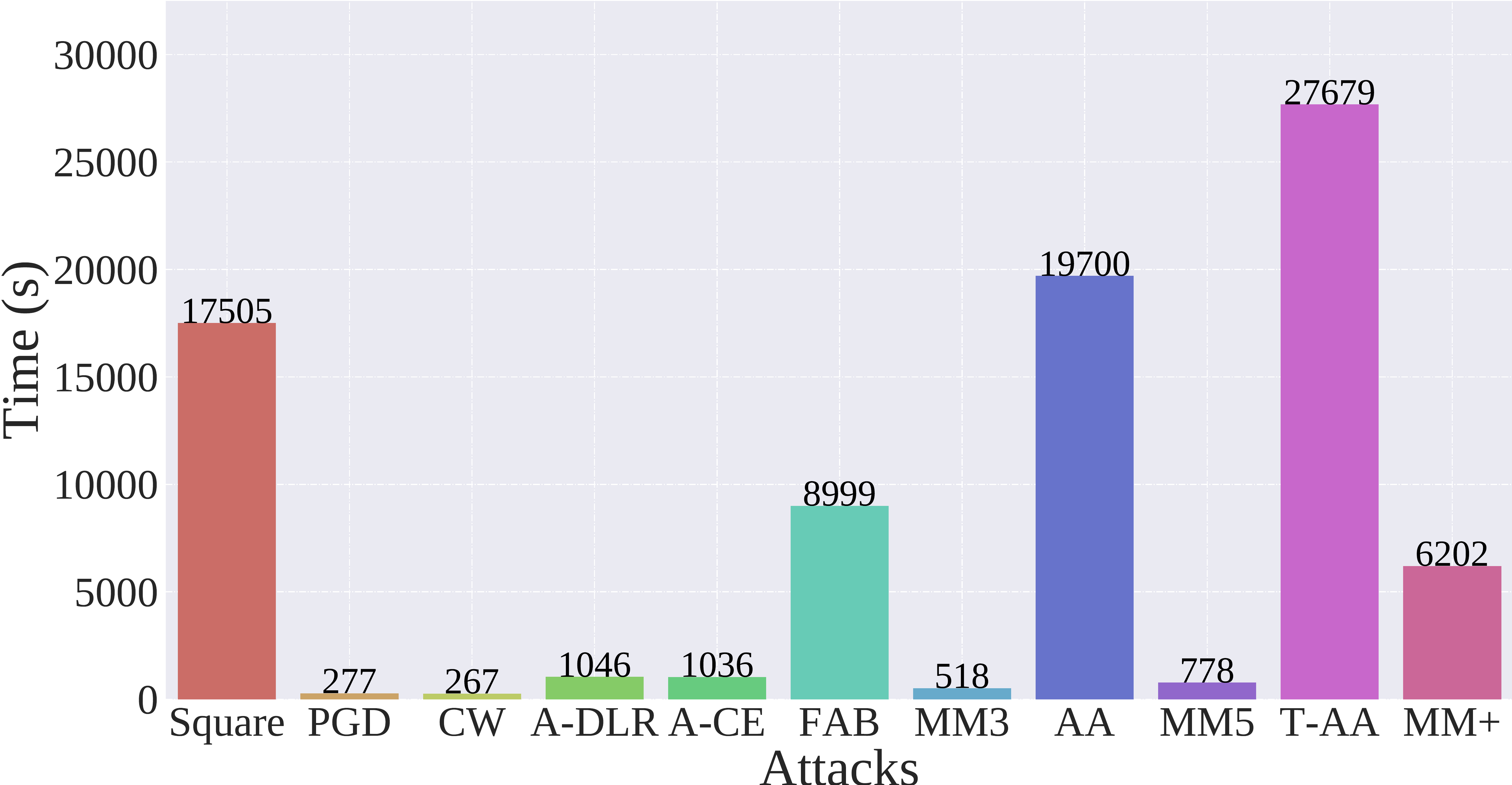}}
        \subfigure[Evaluation on \citet{zhang2020geometry}]
        {\includegraphics[width=0.495\textwidth]{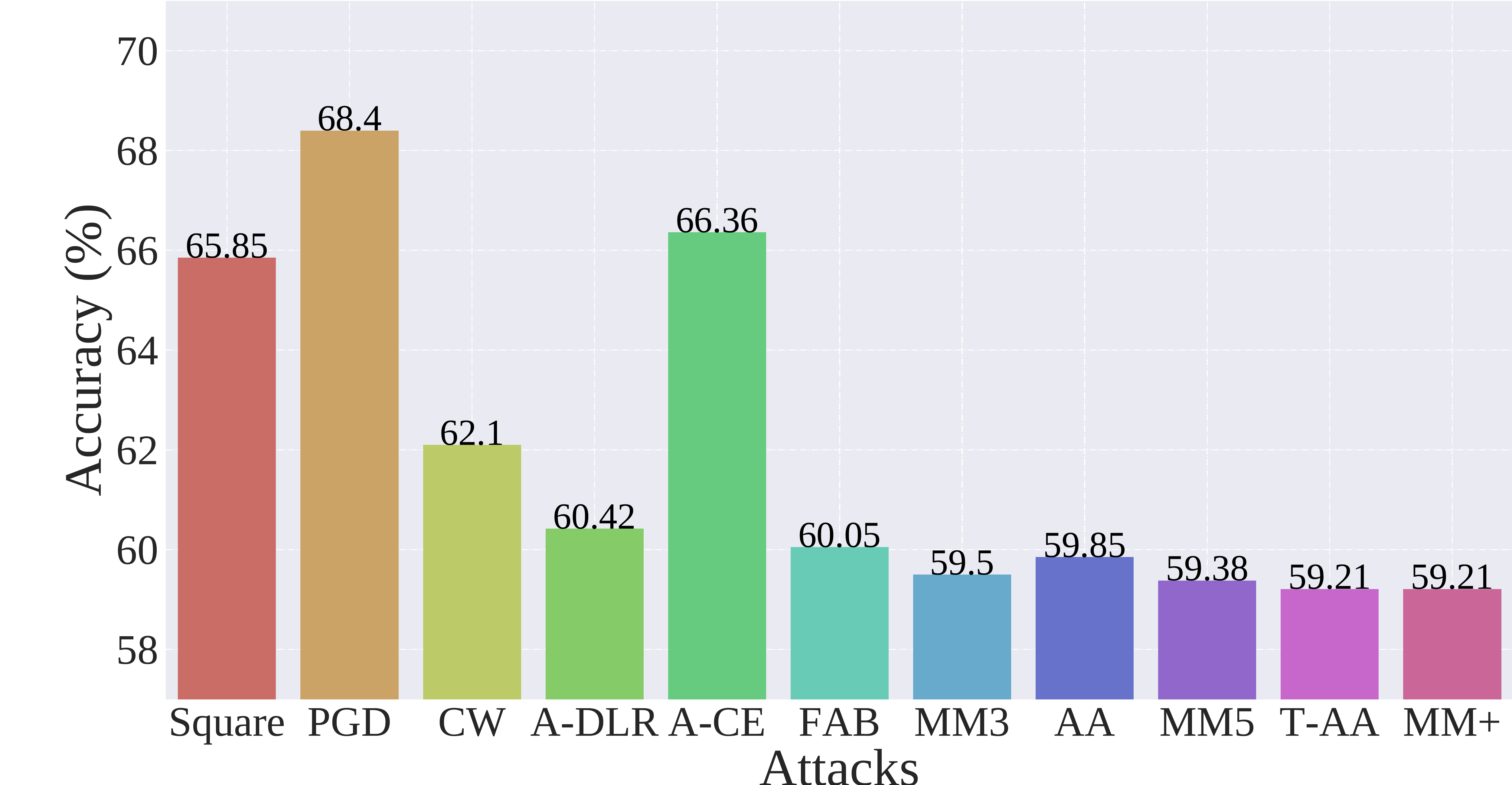}}
        \subfigure[Computational time on \citet{zhang2020geometry}]
        {\includegraphics[width=0.495\textwidth]{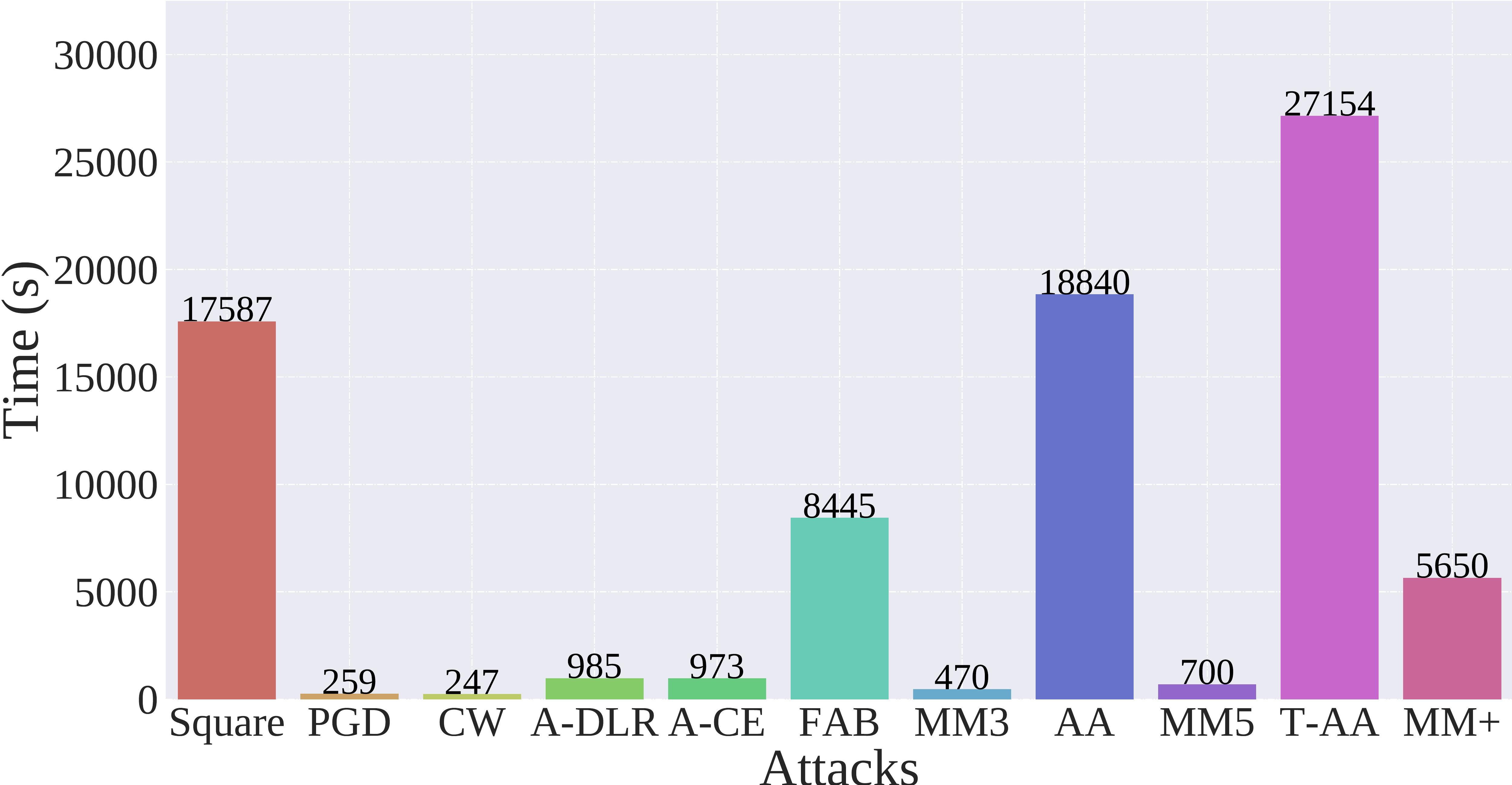}}
        % \vspace{-1em}
        \caption{\footnotesize Comparison of reliability and computational cost on different defense in RobustBench. We compare three versions of our MM attack (MM3, MM5 and MM+ mentioned in Section~\ref{exp:baselines}) with 8 baselines. In subfigure (a), (c),(e) and (g), the Y-axis is the accuracy of the attacked model, which means that the lower the accuracy, the stronger the attack (or to say the better evaluation). In subfigure (b), (d), (f) and (h), the Y-axis is computational time, which means the less the time, the higher the computational efficiency. Experiments are on CIFAR-10 with $L_{\infty}$-norm bounded perturbation.
        %In the subfigure (b), the gray shape is a hypothetical distribution of all adversarial variants maps within the bounded perturbation epsilon on a natural example; $\mathbb{P}_t$ and $\mathbb{P}_y$ are the predicted probability on a targeted false label $t$ and the true label $y$; The orange area ($\mathbb{P}_t > \mathbb{P}_y$) indicates that the adversarial variants inside can be misclassified, or to say attack successfully, while the blue area ($\mathbb{P}_t < \mathbb{P}_y$) indicates that the adversarial variants inside cannot attack successfully.
        }
        % \vspace{-2em}
    \label{robustbench_exp3}
    \end{center}
    \vspace{-1em}
\end{figure*}

\begin{figure*}[!h]
    \begin{center}
        \subfigure[Evaluation on \citet{sehwag2021improving}]
        {\includegraphics[width=0.495\textwidth]{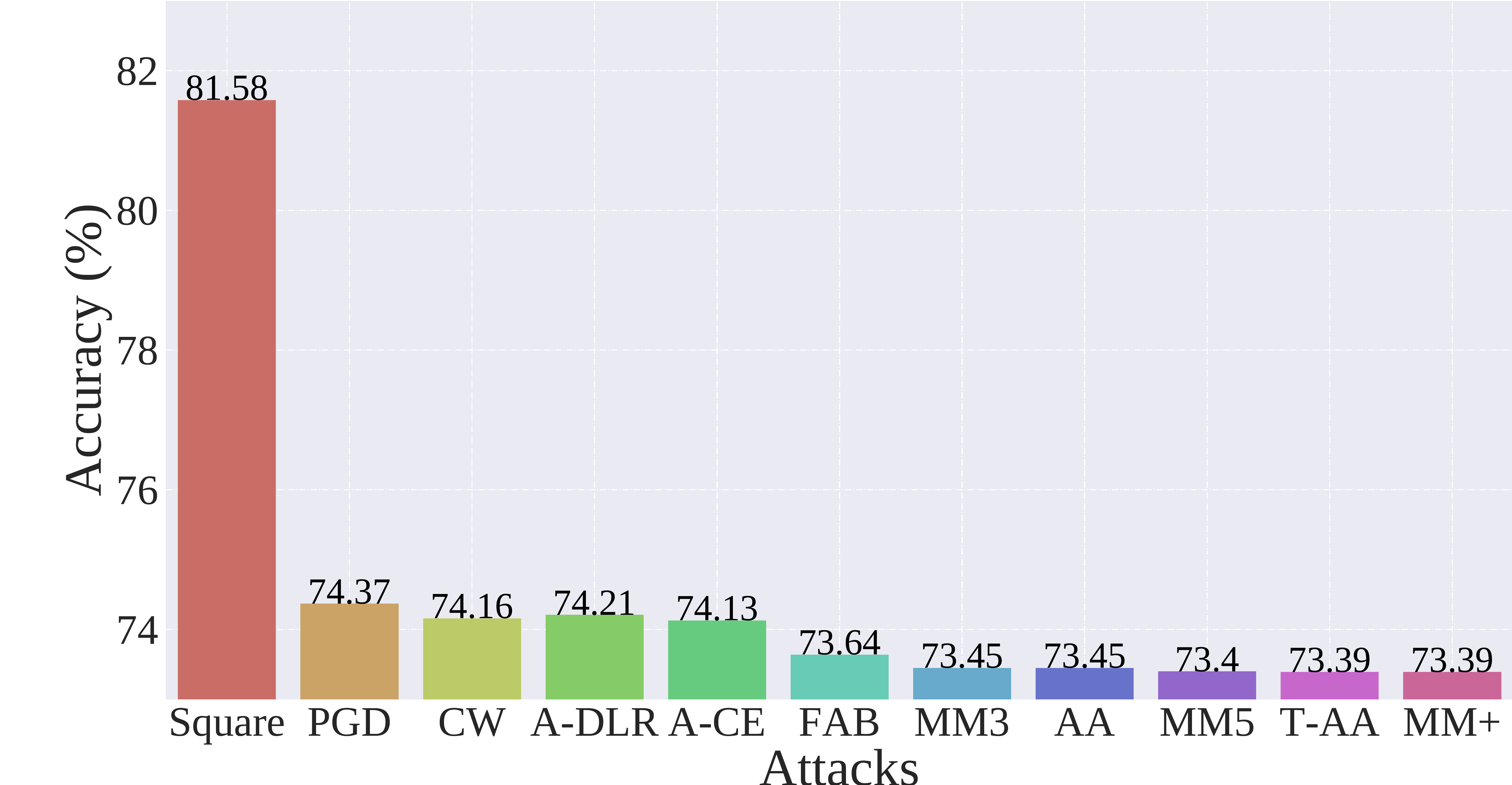}}
        \subfigure[Computational time on \citet{sehwag2021improving}]
        {\includegraphics[width=0.495\textwidth]{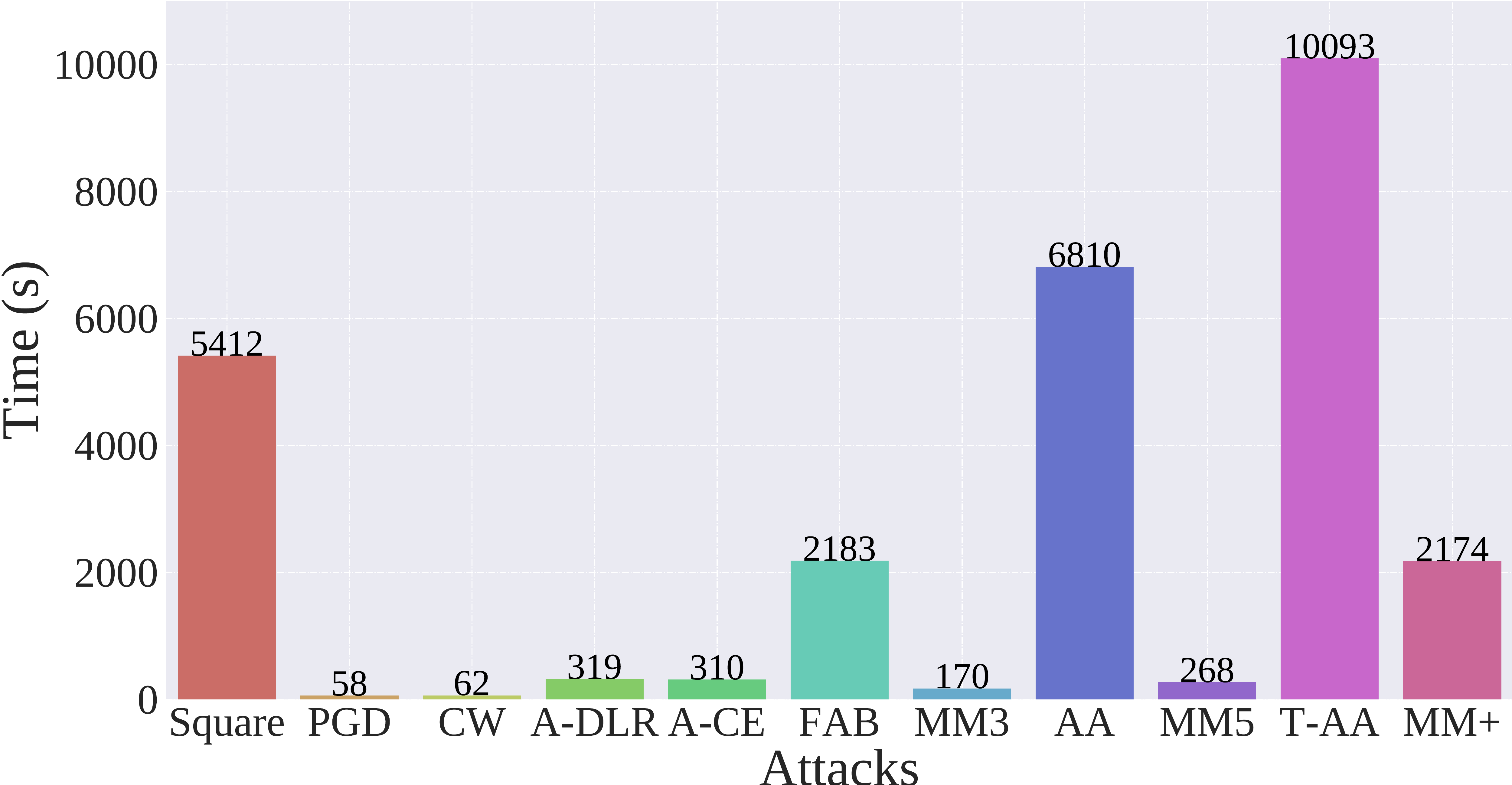}}
        \subfigure[Evaluation on \citet{rade2021helper}]
        {\includegraphics[width=0.495\textwidth]{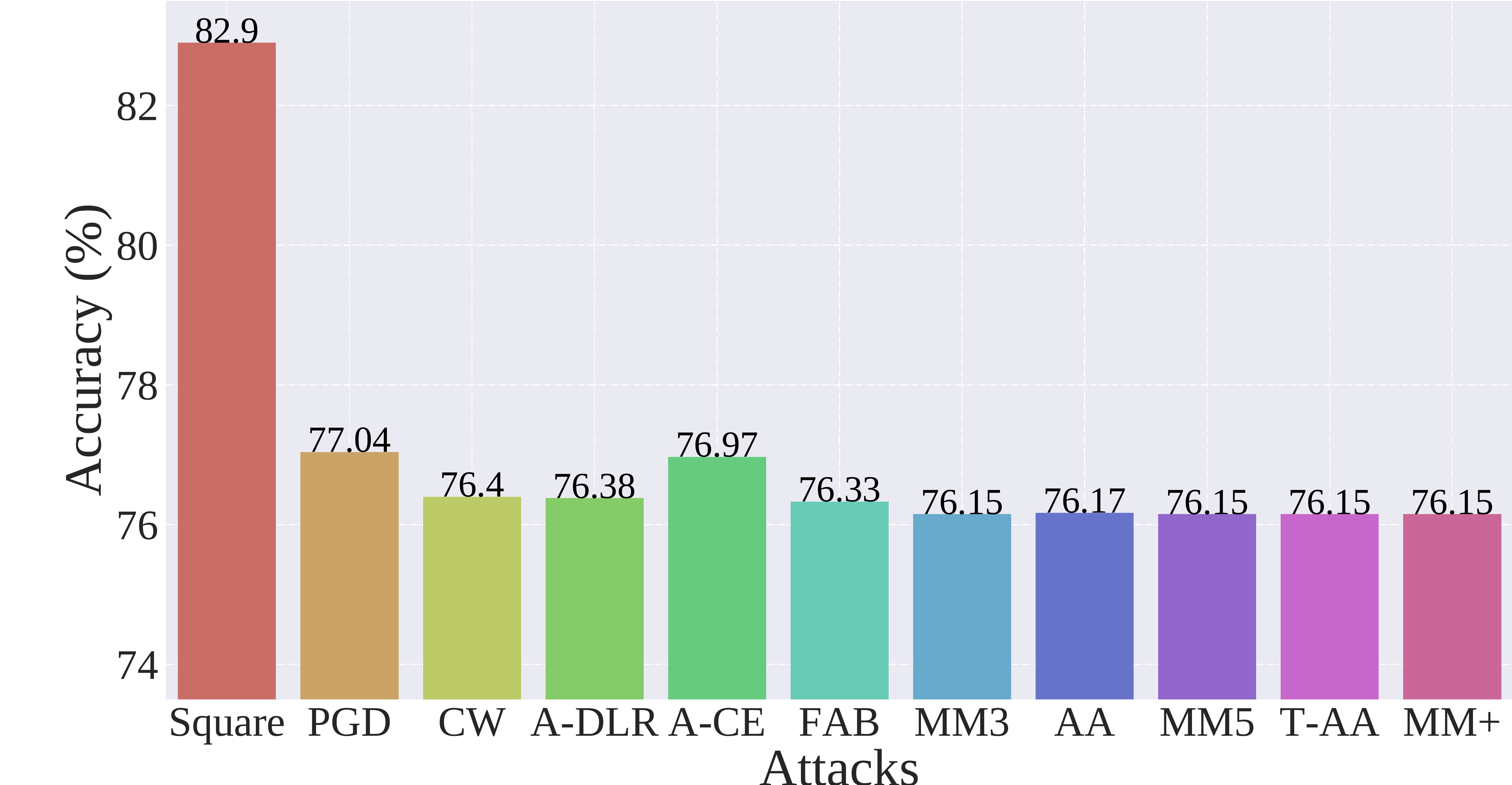}}
        \subfigure[Computational time on \citet{rice2020overfitting}]
        {\includegraphics[width=0.495\textwidth]{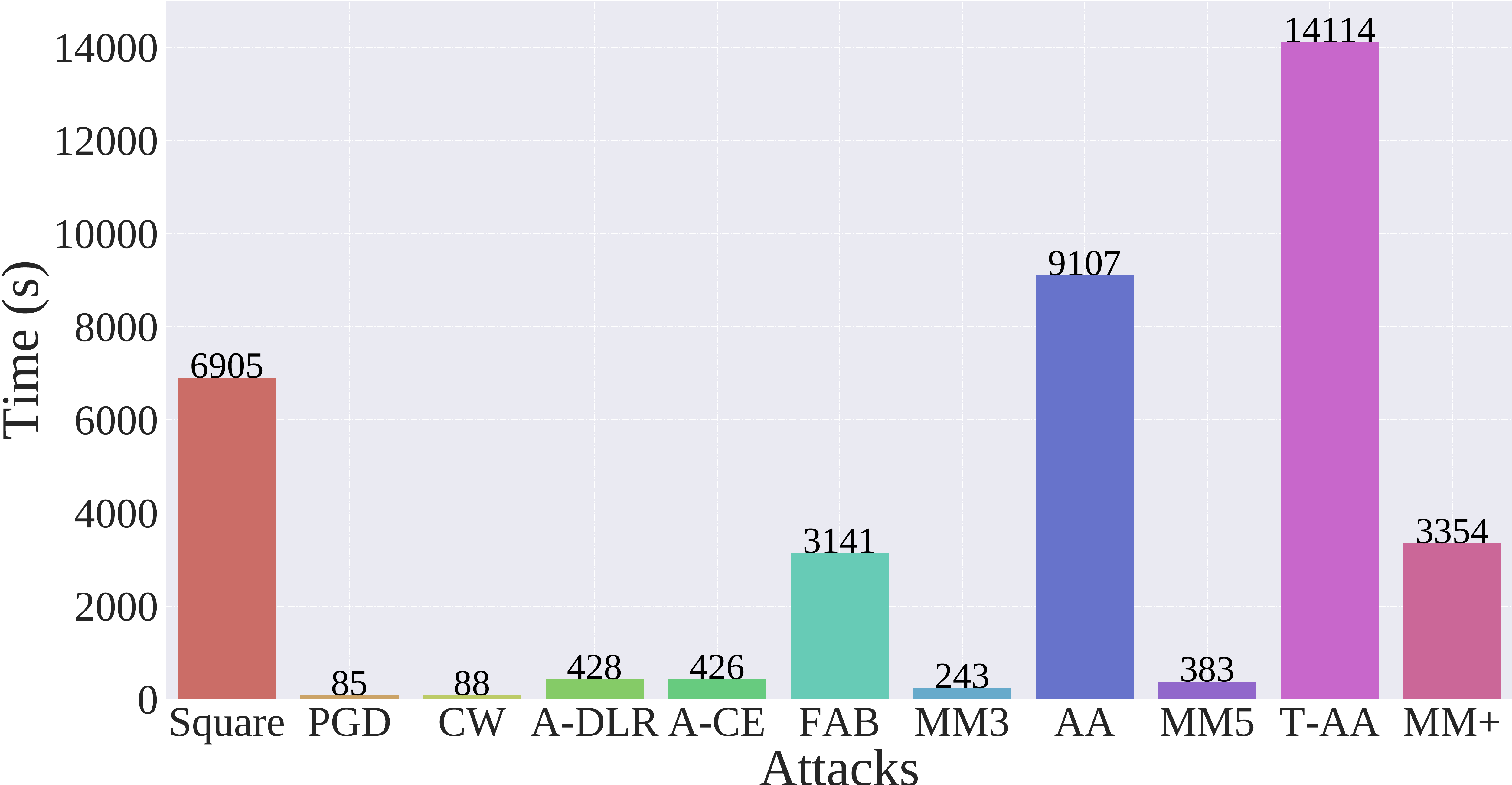}}
        \subfigure[Evaluation on \citet{rice2020overfitting}]
        {\includegraphics[width=0.495\textwidth]{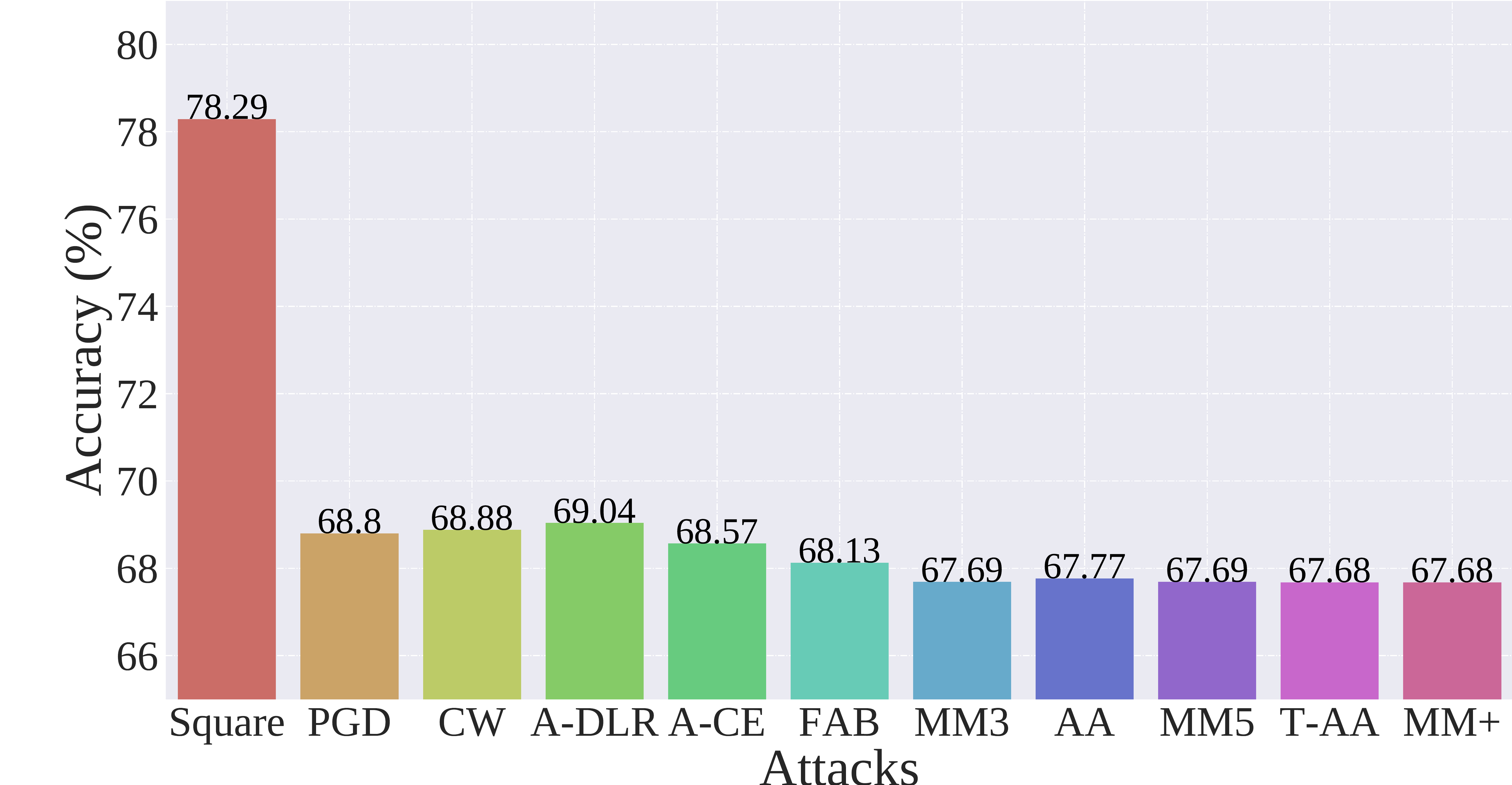}}
        \subfigure[Computational time on \citet{rice2020overfitting}]
        {\includegraphics[width=0.495\textwidth]{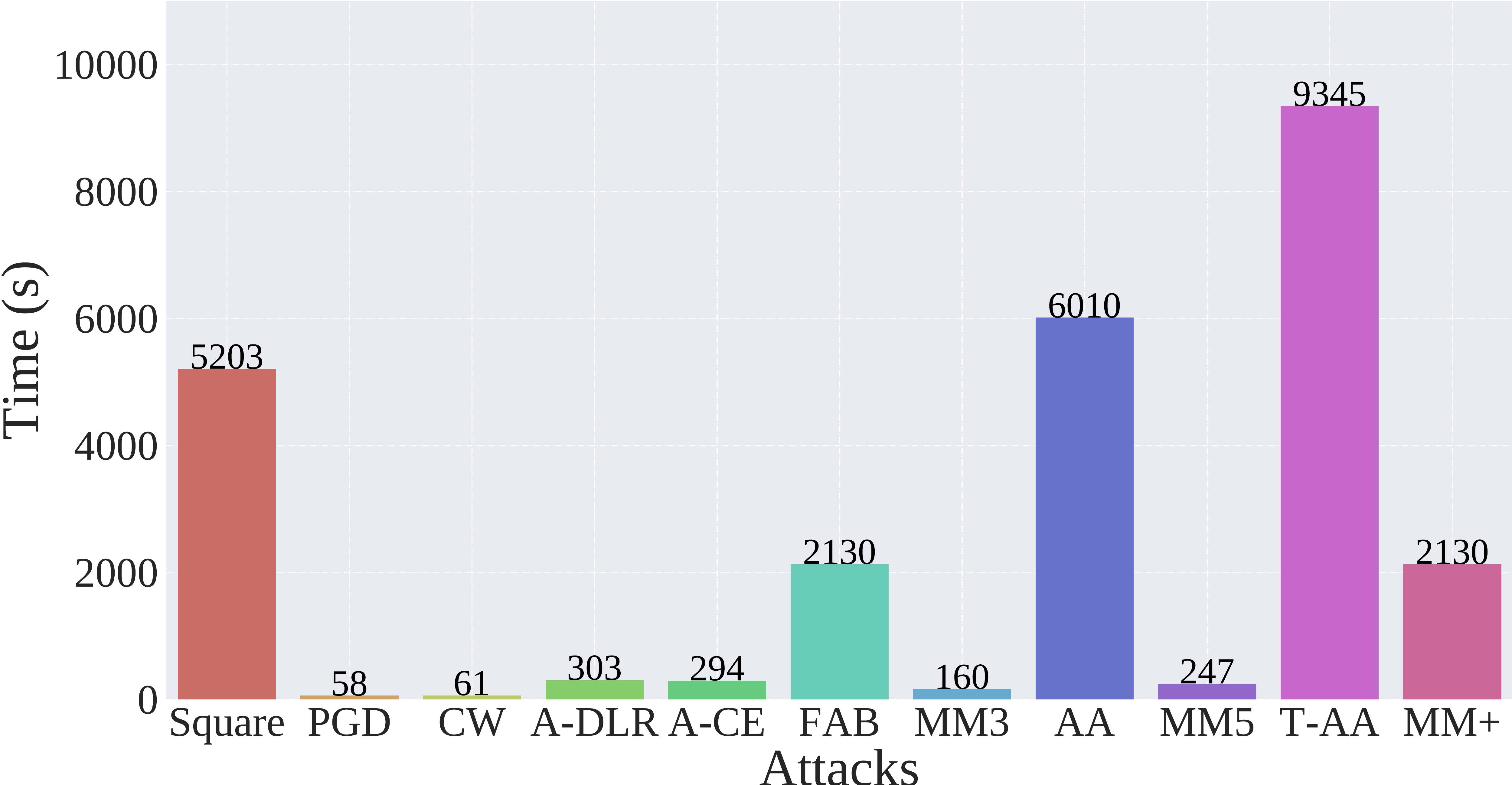}}
        \subfigure[Evaluation on \citet{rebuffi2021fixing}]
        {\includegraphics[width=0.495\textwidth]{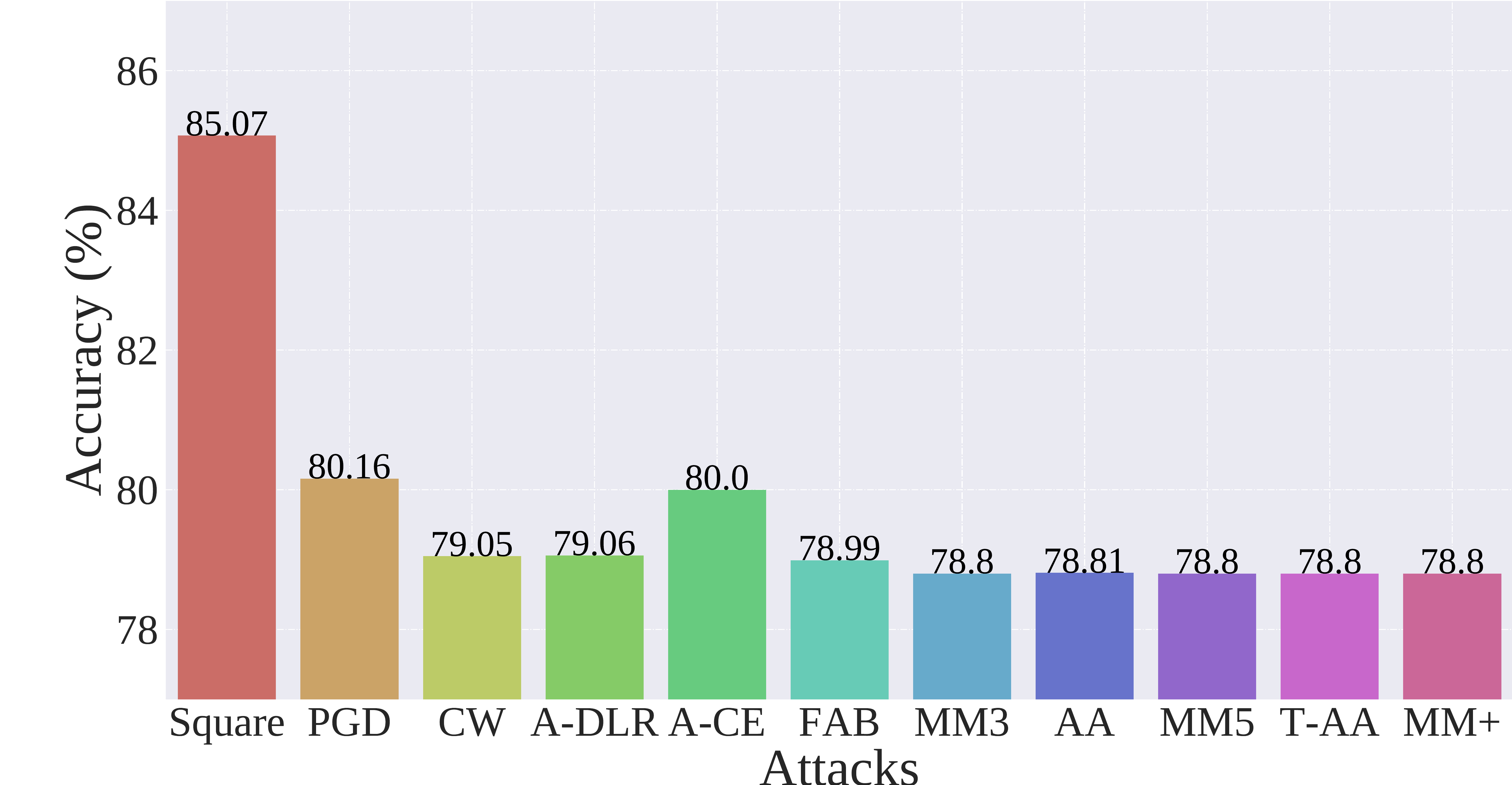}}
        \subfigure[Computational time on \citet{rebuffi2021fixing}]
        {\includegraphics[width=0.495\textwidth]{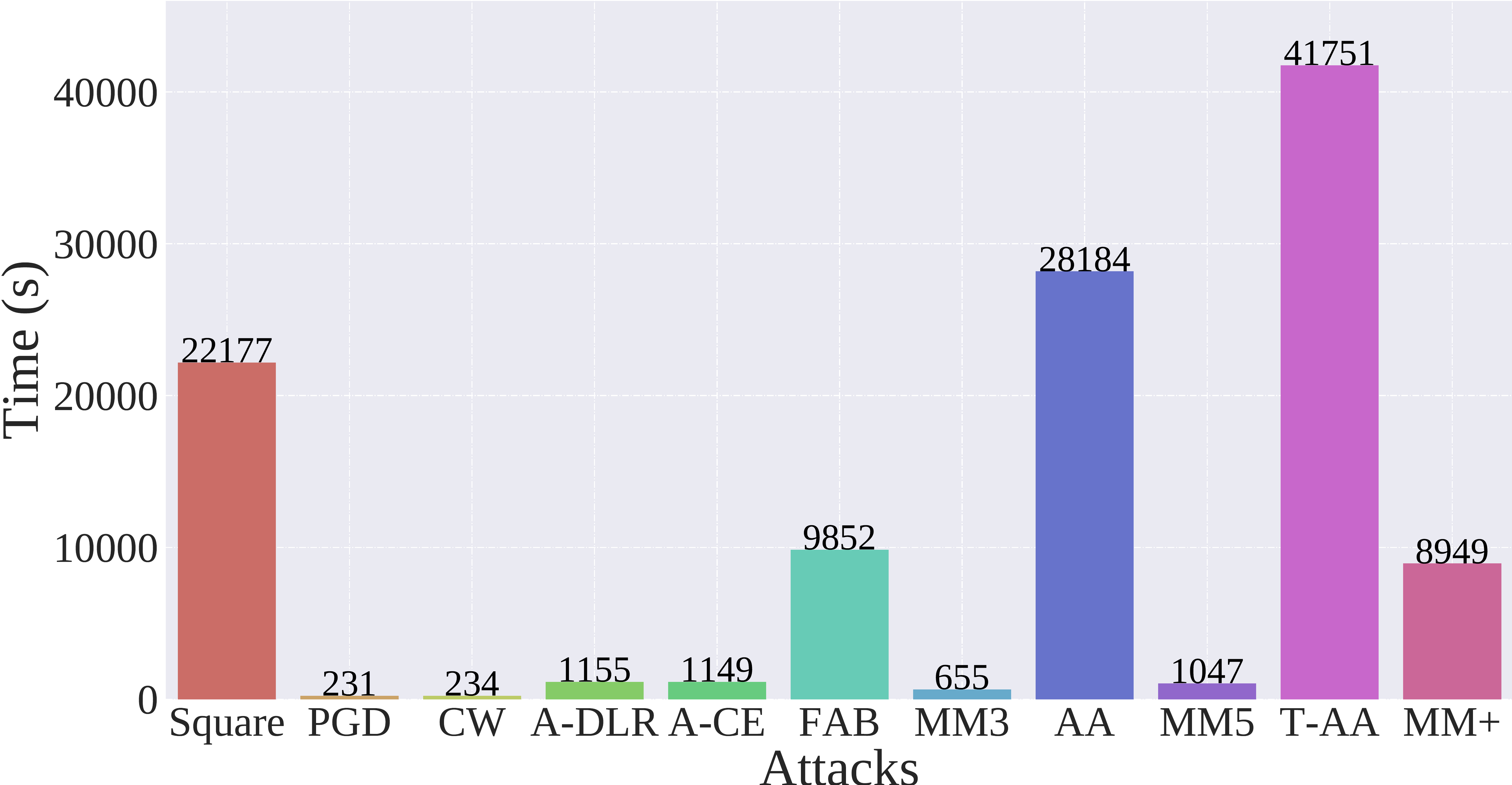}}
        % \vspace{-1em}
        \caption{\footnotesize Comparison of reliability and computational cost on different defense in RobustBench. We compare three versions of our MM attack (MM3, MM5 and MM+ mentioned in Section~\ref{exp:baselines}) with 8 baselines. In subfigure (a), (c),(e) and (g), the Y-axis is the accuracy of the attacked model, which means that the lower the accuracy, the stronger the attack (or to say the better evaluation). In subfigure (b), (d), (f) and (h), the Y-axis is computational time, which means the less the time, the higher the computational efficiency. Experiments are on CIFAR-10 with $L_{2}$-norm bounded perturbation. 
        %In the subfigure (b), the gray shape is a hypothetical distribution of all adversarial variants maps within the bounded perturbation epsilon on a natural example; $\mathbb{P}_t$ and $\mathbb{P}_y$ are the predicted probability on a targeted false label $t$ and the true label $y$; The orange area ($\mathbb{P}_t > \mathbb{P}_y$) indicates that the adversarial variants inside can be misclassified, or to say attack successfully, while the blue area ($\mathbb{P}_t < \mathbb{P}_y$) indicates that the adversarial variants inside cannot attack successfully.
        }
        % \vspace{-2em}
    \label{robustbench_exp4}
    \end{center}
    \vspace{-1em}
\end{figure*}

\begin{figure*}[!h]
    \begin{center}
        \subfigure[Evaluation on \citet{robustness}]
        {\includegraphics[width=0.495\textwidth]{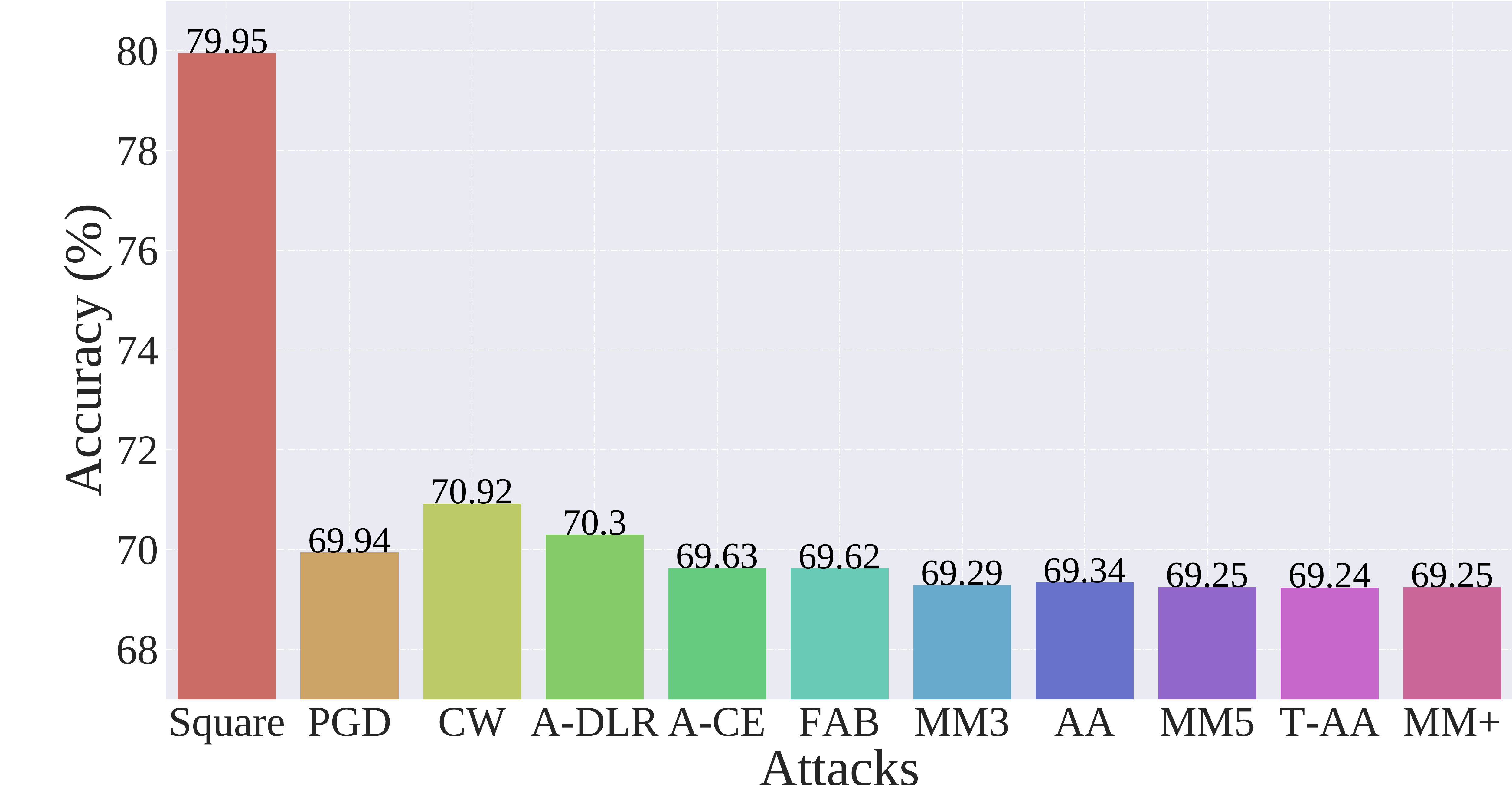}}
        \subfigure[Computational time on \citet{robustness}]
        {\includegraphics[width=0.495\textwidth]{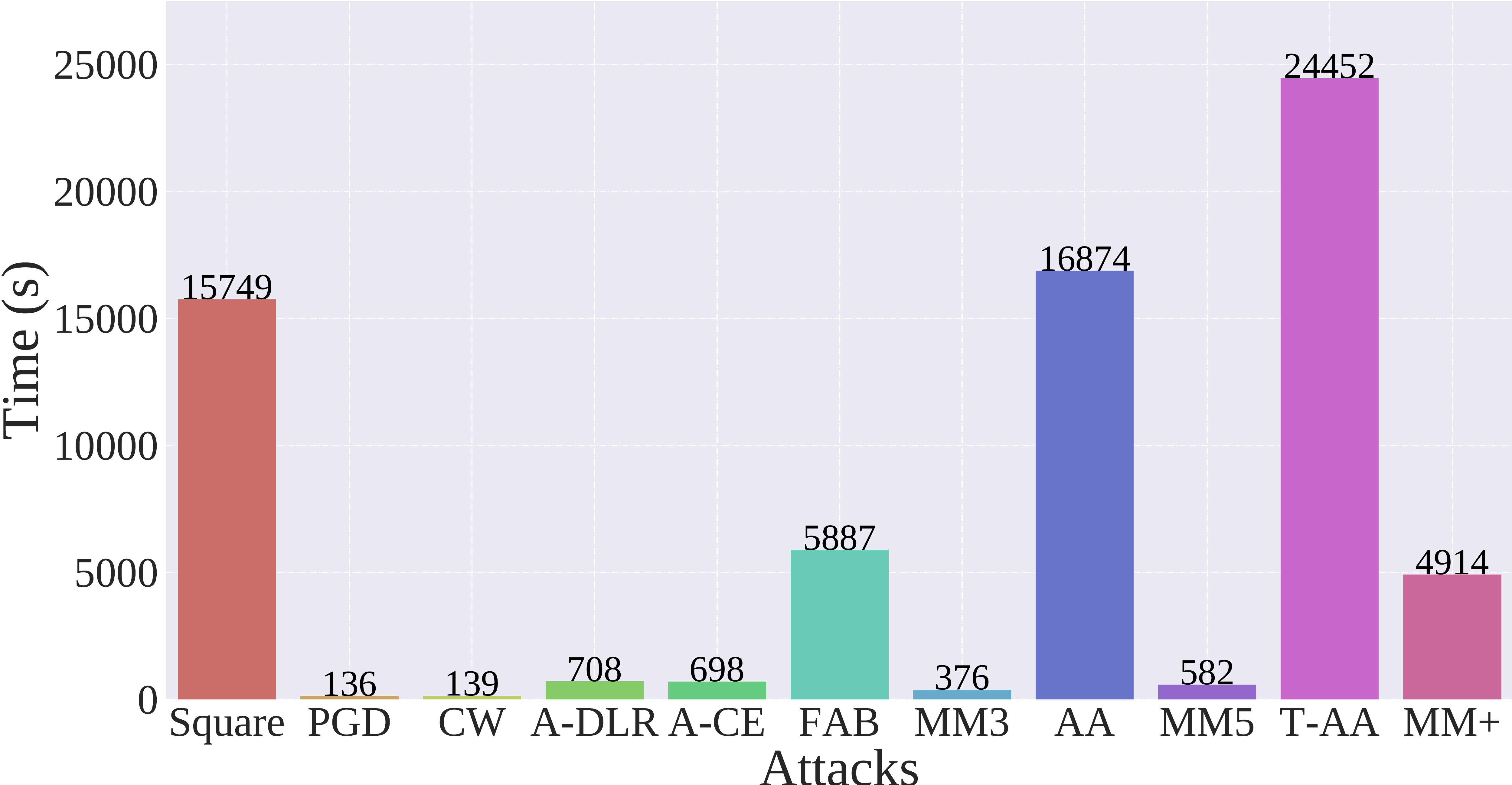}}
        \subfigure[Evaluation on \citet{augustin2020adversarial}]
        {\includegraphics[width=0.495\textwidth]{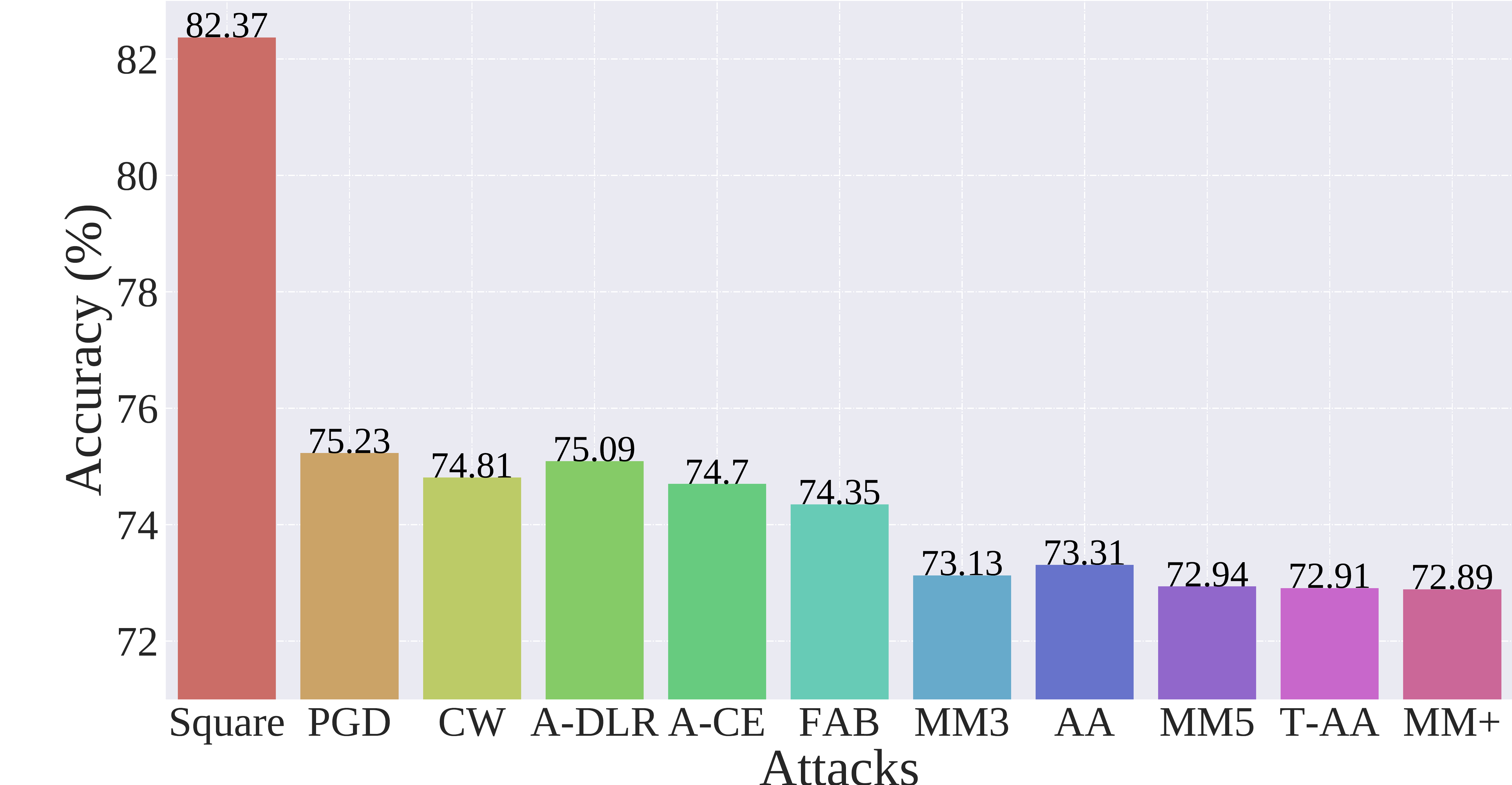}}
        \subfigure[Computational time on \citet{augustin2020adversarial}]
        {\includegraphics[width=0.495\textwidth]{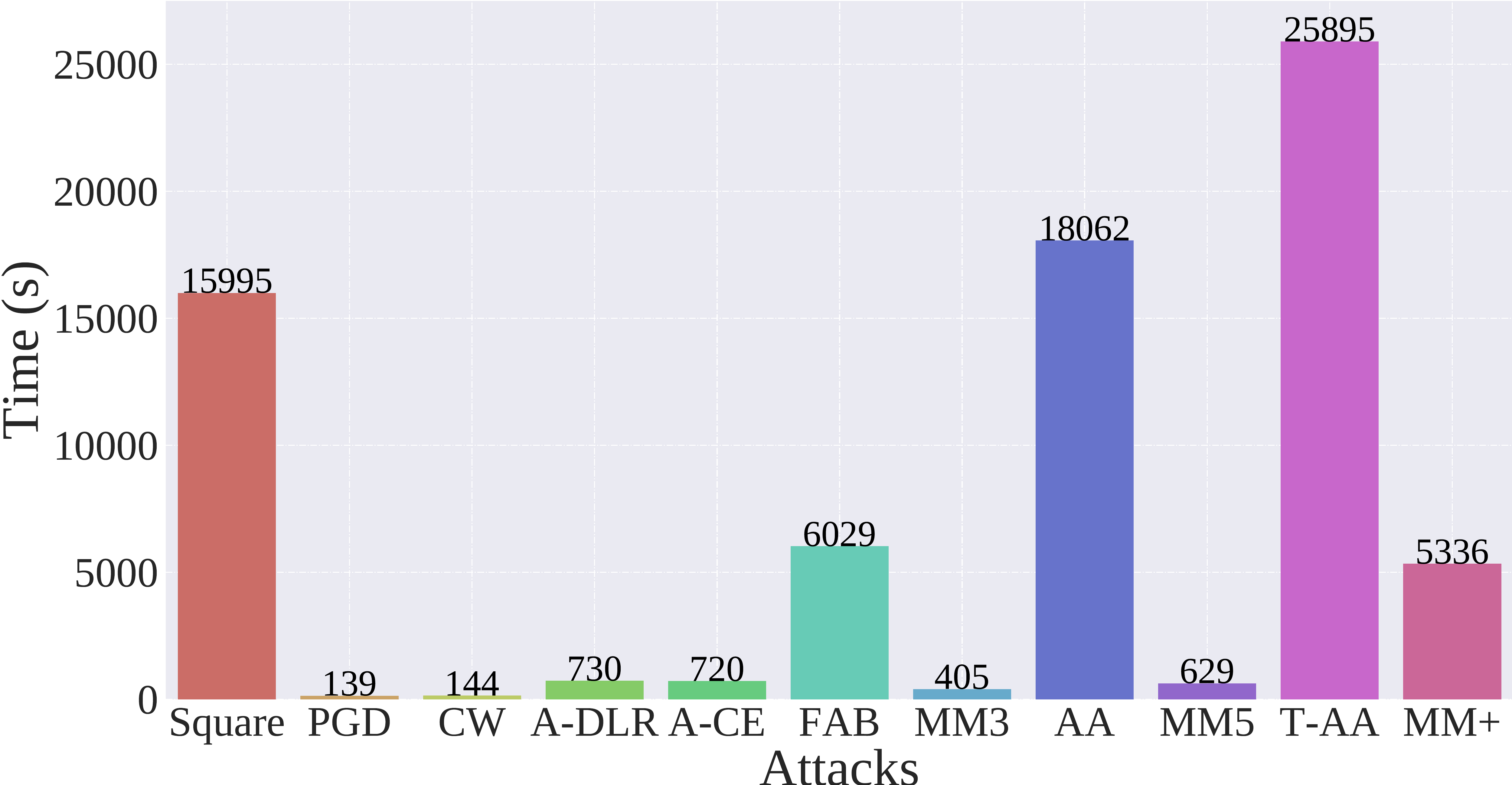}}
        % \vspace{-1em}
        \caption{\footnotesize Comparison of reliability and computational cost on different defense in RobustBench. We compare three versions of our MM attack (MM3, MM5 and MM+ mentioned in Section~\ref{exp:baselines}) with 8 baselines. In subfigure (a) and (c), the Y-axis is the accuracy of the attacked model, which means that the lower the accuracy, the stronger the attack (or to say the better evaluation). In subfigure (b) and (d), the Y-axis is computational time, which means the less the time, the higher the computational efficiency. Experiments are on CIFAR-10 with $L_{2}$-norm bounded perturbation.  
        %In the subfigure (b), the gray shape is a hypothetical distribution of all adversarial variants maps within the bounded perturbation epsilon on a natural example; $\mathbb{P}_t$ and $\mathbb{P}_y$ are the predicted probability on a targeted false label $t$ and the true label $y$; The orange area ($\mathbb{P}_t > \mathbb{P}_y$) indicates that the adversarial variants inside can be misclassified, or to say attack successfully, while the blue area ($\mathbb{P}_t < \mathbb{P}_y$) indicates that the adversarial variants inside cannot attack successfully.
        }
        % \vspace{-2em}
    \label{robustbench_exp5}
    \end{center}
    \vspace{-1em}
\end{figure*}

We conduct extensive experiments between our MM attack with baselines on 12 defenses of RobustBench in Figure~\ref{robustbench_exp1}, Figure~\ref{robustbench_exp2} and Figure~\ref{robustbench_exp3} (Appendix~\ref{app:robustbench}) \citep{sehwag2021improving, rade2021helper, rebuffi2021fixing, andriushchenko2020understanding, gowal2020uncovering, sridhar2021robust, wong2020fast, robustness, carmon2019unlabeled, wang2019improving, wu2020adversarial, zhang2020geometry}. For $L_{2}$-norm bounded perturbation that $||\tilde{x}-x{||}_2 \leq \epsilon$, we also compare the performance on the 6 defenses of RobustBench in Figure~\ref{robustbench_exp4} and Figure~\ref{robustbench_exp5} \citep{sehwag2021improving, rade2021helper, rice2020overfitting, rebuffi2021fixing, robustness, augustin2020adversarial}. Experimental results verify the effectiveness of MM attack on diverse trained model in RobustBench.

\section{Experimental Details of Adversarial Training }
\label{app:advtraining}

We choose $11$ testing methods to evaluate the robustness of MM attack-based AT. Table~\ref{table:exp_AT} reports the performance. MM3-F10 (MM3-F20) denotes AT using adversarial examples generated by MM attack with 10 (20) fixed steps. MM3 denotes AT using adversarial examples generated by MM attack with 20 adaptive steps.  As shown in Table~\ref{table:exp_AT}, by replacing PGD-generated adversarial examples with MM-attack-generated adversarial examples, the robustness of AT model is significantly improved.

\section{Experimental Resources}
\label{App:resource}
We implement all methods on Python $3.7$ (Pytorch $1.7.1$) with an NVIDIA GeForce RTX 3090 GPU with AMD Ryzen Threadripper 3960X 24 Core Processor. The CIFAR-10 dataset, the SVHN and the CIFAR-100 dataset can be downloaded via Pytorch. Given the $50,000$ images from the CIFAR-10 and CIFAR-100 training set, $73,257$ digits from the SVHN training set, we conduct the adversarial training on ResNet-18 and Wide ResNet-34 for classification.
\begin{algorithm}[!t]
\footnotesize
\caption{Adversarial Training of MM attack.}
\label{alg:MM-AT}
\begin{algorithmic}[1]
\STATE \textbf{Input:} network architecture $f$ parametrized by $\theta$, training dataset $S$, loss function $l$, learning rate $\eta$, number of epochs $T$, batch size $n$;
\STATE \textbf{Output:} Adversarial robust network $f_\theta$;
\FOR{epoch = $1,2,\dots,T$}
\FOR{mini-batch = 1,2,\dots,$N$}
\STATE Sample a mini-batch $\left\{ \left ( x_i,y_i \right) \right\} ^n_{i=1}$ from $S$;
\FOR{$i = 1,2,\dots,n $}
\STATE Obtain adversarial data of MM attack ${x'}_{i}$ of ${x}_i$ by Algorithm~\ref{alg:Alg__MM};
\ENDFOR
\STATE $\theta \gets \theta - \eta  \sum_{i=1}^{n}  \nabla_{\theta} \ell \left ( f_\theta \left ( {x'}_i \right),y_i \right) /n$;
\ENDFOR
\ENDFOR
%\vspace{1mm}
\end{algorithmic}
\end{algorithm}

\begin{table}[!t]
\setlength{\tabcolsep}{8.1mm}
%\footnotesize
\scriptsize
\renewcommand\arraystretch{1.2}
\centering
\caption{Test accuracy (\%): Replacing natural data with adversarial data in STARS method. \emph{Diff.} represents the difference between the current result and the optimal result in the sub-column. The model structure of all methods is ResNet-18. Bold values represent the lowest accuracy (the highest attack success rate) in each sub-column.}
\label{table:advpres}
\begin{tabular}{c|c|c|c c|c c}
\toprule[1.5pt]
Dataset & Reference attack  & Select-$\epsilon$ & MM3 & Diff. & MM9 & Diff. \\
\midrule[0.6pt]
\midrule[0.6pt]
CIFAR-10 & None & 8/255 & 48.23 & -0.42 & 47.81 & 0.00\\
\midrule    
CIFAR-10 & FGSM & 8/255 & 48.05 & -0.24 & 47.81 & 0.00\\
\midrule    
CIFAR-10 & PGD-20 & 8/255 & \textbf{47.92} & -0.11 & 47.81 & 0.00\\
\midrule    
CIFAR-10 & PGD-20 & 6/255 & 47.98 & -0.17 & 47.81 & 0.00\\
\midrule    
CIFAR-10 & PGD-20 & 4/255 & 48.04 & -0.23 & 47.81 & 0.00\\
\midrule[0.6pt]
\midrule[0.6pt]
SVHN & None & 8/255 & 52.45 & -0.61 & 51.84 & 0.00\\
\midrule    
SVHN & FGSM & 8/255 & 52.07 & -0.23 & 51.84 & 0.00\\
\midrule    
SVHN & PGD-20 & 8/255 & \textbf{51.97} & -0.13 & 51.84 & 0.00\\
\midrule    
SVHN & PGD-20 & 6/255 & 52.00 & -0.16 & 51.84 & 0.00\\
\midrule    
SVHN & PGD-20 & 4/255 & 52.07 & -0.23 & 51.84 & 0.00\\
\midrule[0.6pt]
\midrule[0.6pt]
CIFAR-100 & None & 8/255 & 23.92 & -0.41 & 23.51 & 0.00\\
\midrule    
CIFAR-100 & FGSM & 8/255 & 23.63 & -0.12 & 23.51 & 0.00\\
\midrule    
CIFAR-100 & PGD-20 & 8/255 & \textbf{23.57} & -0.06 & 23.51 & 0.00\\
\midrule    
CIFAR-100 & PGD-20 & 6/255 & \textbf{23.57} & -0.06 & 23.51 & 0.00\\
\midrule    
CIFAR-100 & PGD-20 & 4/255 & 23.63 & -0.12 & 23.51 & 0.00\\
\bottomrule[1.5pt]
\end{tabular}
\vskip1ex%
\vskip -0ex%
\vspace{-1em}
\end{table}\

\begin{table}[!t]
\setlength{\tabcolsep}{5.1mm}
%\footnotesize
\scriptsize
\renewcommand\arraystretch{1.2}
\centering
\caption{Test accuracy (\%): the rationality of MM under different step sizes and step numbers. \emph{Diff.} represents the difference between the current result and the optimal result in the row. The model structure of all methods is ResNet-18. Bold values represent the lowest accuracy (the highest attack success rate) in each row.}
\label{table:exp_step}
\begin{tabular}{c|c|c c|c c|c c|c c}
\toprule[1.5pt]
Step size & Step num & PGD-20 & Diff. & CW & Diff. & MM3-F & Diff. & MM9-F & Diff.\\
\midrule[0.6pt]
\midrule[0.6pt]
\multicolumn{10}{c}{CIFAR-10} \\
\midrule[0.6pt]
\midrule[0.6pt]
0.003 & 20 & 51.14 & -3.33 & 49.95 & -2.14 & 48.23 & -0.42 & \textbf{47.81} & 0.00\\
\midrule    
1/255 & 40 & 50.16 & -3.15 & 49.13 & -2.12 & 47.46 & -0.45 & \textbf{47.01} & 0.00\\
\midrule    
1/255 & 20 & 50.28 & -3.22 & 49.19 & -2.13 & 47.50 & -0.44 & \textbf{47.06} & 0.00\\
\midrule    
1/255 & 40 & 49.30 & -2.92 & 48.45 & -2.07 & 46.88 & -0.50 & \textbf{46.38} & 0.00\\
\midrule    
2/255 & 10 & 50.54 & -3.26 & 49.38 & -2.10 & 46.70 & -0.42 & \textbf{47.28} & 0.00\\
\midrule    
2/255 & 20 & 49.36 & -2.93 & 48.48 & -2.05 & 46.92 & -0.49 & \textbf{46.43} & 0.00\\
\midrule    
4/255 & 10 & 49.52 & -2.97 & 48.60 & -2.05 & 47.02 & -0.47 & \textbf{46.55} & 0.00\\
\midrule[0.6pt]
\midrule[0.6pt] 
\multicolumn{10}{c}{SVHN} \\
\midrule[0.6pt]
\midrule[0.6pt]
0.003 & 20 & 57.68 & -5.84 & 54.42 & -2.58 & 52.45 & -0.61 & \textbf{51.84} & 0.00\\
\midrule    
1/255 & 40 & 56.03 & -5.78 & 52.90 & -2.65 & 50.91 & -0.66 & \textbf{50.25} & 0.00\\
\midrule    
1/255 & 20 & 56.81 & -5.74 & 53.69 & -2.62 & 51.72 & -0.65 & \textbf{51.07} & 0.00\\
\midrule    
1/255 & 40 & 55.49 & -5.50 & 52.59 & -2.60 & 50.65 & -0.66 & \textbf{49.99} & 0.00\\
\midrule    
2/255 & 10 & 57.30 & -5.71 & 54.12 & -2.53 & 52.19 & -0.60 & \textbf{51.59} & 0.00\\
\midrule    
2/255 & 20 & 55.45 & -5.32 & 52.70 & -2.57 & 50.79 & -0.66 & \textbf{50.13} & 0.00\\
\midrule    
4/255 & 10 & 56.16 & -5.13 & 53.52 & -2.49 & 51.62 & -0.59 & \textbf{51.03} & 0.00\\
\bottomrule[1.5pt]
\end{tabular}
\vskip1ex%
\vskip -0ex%
\vspace{-1em}
\end{table}\
\begin{table}[!t]
\setlength{\tabcolsep}{6.5mm}
%\footnotesize
\scriptsize
\renewcommand\arraystretch{1.2}
\centering
\caption{Test accuracy (\%): the rationality of MM under different $\epsball[x]$. \emph{Diff.} represents the difference between the current result and the optimal result in the row. The model structure of all methods is ResNet-18. Bold values represent the lowest accuracy (the highest attack success rate) in each row.}
\label{table:exp_epsilon}
\begin{tabular}{c|c c|c c|c c|c c}
\toprule[1.5pt]
$\epsilon$ & PGD-20 & Diff. & CW & Diff. & MM3-F & Diff. & MM9-F & Diff.\\
\midrule[0.6pt]
\midrule[0.6pt]
\multicolumn{9}{c}{ResNet-18} \\
\midrule[0.6pt]
\midrule[0.6pt]
4 & 67.90 & -0.70 & 68.06 & -0.86 & 67.23 & -0.03 & \textbf{67.20} & 0.00\\
\midrule    
8 & 51.14 & -3.33 & 49.95 & -2.14 & 48.23 & -0.42 & \textbf{47.81} & 0.00\\
\midrule    
12 & 45.53 & -4.62 & 43.85 & -2.94 & 41.86 & -0.95 & \textbf{40.91} & 0.00\\
\midrule[0.6pt]
\midrule[0.6pt] 
\multicolumn{9}{c}{WRN-34} \\
\midrule[0.6pt]
\midrule[0.6pt]
4 & 70.23 & -0.30 & 70.55 & -0.62 & 69.94 & -0.01 & \textbf{69.93} & 0.00\\
\midrule    
8 & 53.69 & -2.07 & 53.89 & -2.27 & 51.95 & -0.33 & \textbf{51.62} & 0.00\\
\midrule    
12 & 46.76 & -3.68 & 46.24 & -3.16 & 44.05 & -0.97 & \textbf{43.08} & 0.00\\
\bottomrule[1.5pt]
\end{tabular}
\vskip1ex%
\vskip -0ex%
\vspace{-1em}
\end{table}\

\begin{table}[!t]
\setlength{\tabcolsep}{5.1mm}
%\footnotesize
\scriptsize
\renewcommand\arraystretch{1.4}
\centering
\caption{Evaluation: test accuracy (\%) on different datasets and model structures. \emph{Diff.} represents the difference between the current result and the optimal result in the column. The model structure of the methods that are not specified is ResNet-18. Bold values represent the lowest accuracy (the highest attack success rate) in each column.}
\label{app:exp_acc}
\begin{tabular}{c|c c|c c|c c|c c}
\toprule[1.5pt]
Methods & CIFAR-10 & Diff. & CIFAR-100 & Diff. & SVHN & Diff. & [WRN34]~CIFAR-10  & Diff. \\
\midrule[0.6pt]
\midrule[0.6pt]
PGD & 51.14 & -5.03 & 26.45 & -3.92 & 57.68 & -10.39 & 53.70 & -3.88 \\
\midrule
CW & 49.95 & -3.84 & 25.60 & -3.07 & 54.50 & -7.21 & 53.90 & -4.08 \\
\midrule
A-CE & 48.58 & -2.47 & 24.71 & -2.18 & 51.55 & -4.26 & 51.00 & -1.18 \\
\midrule
A-DLR & 48.85 & -2.74 & 24.85 & -2.32 & 50.64 & -3.35 & 52.24 & -2.42 \\
\midrule
FAB & 47.28 & -1.17 & 23.16 & -0.63 & 52.19 & -4.90 & 51.04 & -1.22 \\
\midrule
Square & 54.46 & -8.35 & 27.94 & -5.41 & 53.80 & -6.51 & 58.04 & -8.22 \\
\midrule
AA & 46.43 & -0.32 & 23.07 & -0.54 & 48.44 & -1.15 & 50.21 & -0.39 \\
\midrule
T-AA & 46.12 & -0.01 & \textbf{22.53} & 0.00 & 47.36 & -0.07 & \textbf{49.82} & 0.00 \\
\midrule
MM3 & 46.69 & -0.58 & 22.98 & -0.45 & 49.15 & -1.86 & 50.26 & -0.44 \\
\midrule
MM5 & 46.34 & -0.23 & 22.72 & -0.19 & 48.69 & -1.40 & 49.99 & -0.17 \\
\midrule
MM+ & \textbf{46.11} & 0.00 & \textbf{22.53} & 0.00 & \textbf{47.29} & 0.00 & \textbf{49.82} & 0.00 \\
\bottomrule[1.5pt]
\end{tabular}
\vskip1ex%
\vskip -0ex%
\vspace{-1em}
\end{table}\
%51.14 & -5.03 & 26.45 & -3.92 & 57.68 & -10.39 & 53.70 & -3.88
%49.95 & -3.84 & 25.60 & -3.07 & 54.50 & -7.21 & 53.90 & -4.08
%48.58 & -2.47 & 24.71 & -2.18 & 51.55 & -4.26 & 51.00 & -1.18
%48.85 & -2.74 & 24.85 & -2.32 & 50.64 & -3.35 & 52.24 & -2.42
%47.28 & -1.17 & 23.16 & -0.63 & 52.19 & -4.90 & 51.04 & -1.22
%54.46 & -8.35 & 27.94 & -5.41 & 53.80 & -6.51 & 58.04 & -8.22
%46.43 & -0.32 & 23.07 & -0.54 & 48.44 & -1.15 & 50.21 & -0.39
%46.12 & -0.01 & 22.53 & 0.00 & 47.36 & -0.07 & 49.82 & 0.00
%46.69 & -0.58 & 22.98 & -0.45 & 49.15 & -1.86 & 50.26 & -0.44
%46.34 & -0.23 & 22.72 & -0.19 & 48.69 & -1.40 & 49.99 & -0.17
%46.11 & 0.00 & 22.53 & 0.00 & 47.29 & 0.00 & 49.82 & 0.00
\begin{table}[!t]
\setlength{\tabcolsep}{4.5mm}
%\footnotesize
\scriptsize
\renewcommand\arraystretch{1.4}
\centering
\caption{Evaluation: the computational time (s) on different datasets and model structures. \emph{Diff.} represents the difference between the current computational time and the least computational time in the column. The model structure of the methods that are not specified is ResNet-18. Bold values represent the least computational time in each column.}
\label{app:exp_time}
\begin{tabular}{c|c c|c c|c c|c c}
\toprule[1.5pt]
Methods & CIFAR-10 & Diff. & CIFAR-100 & Diff. & SVHN & Diff. & [WRN34]~CIFAR-10  & Diff. \\
\midrule[0.6pt]
\midrule[0.6pt]
PGD & \textbf{60} & 0 & \textbf{60} & 0 & 166 & -2 & 416 & -10 \\
\midrule
CW & 62 & -2 & 64 & -4 & \textbf{164} & 0 & \textbf{406} & 0 \\
\midrule
A-CE & 289 & -229 & 215 & -155 & 777 & -613 & 1910 & -1504 \\
\midrule
A-DLR & 305 & -245 & 222 & -162 & 871 & -707 & 1901 & -1495 \\
\midrule
FAB & 2181 & -2121 & 1980 & -1920 & 6178 & -6014 & 13809 & -13403 \\
\midrule
Square & 3768 & -3708 & 2528 & -2468 & 9506 & -9342 & 22593 & -22187 \\
\midrule
AA & 3885 & -3825 & 2187 & -2127 & 11146 & -10982 & 29637 & -29231 \\
\midrule
T-AA & 5970 & -5910 & 2967 & -2907 & 25116 & -24952 & 40178 & -39772 \\
\midrule
MM3 & 126 & -66 & 91 & -31 & 332 & -168 & 796 & -390 \\
\midrule
MM5 & 182 & -122 & 137 & -77 & 587 & -423 & 1342 & -936 \\
\midrule
MM+ & 1421 & -1361 & 746 & -686 & 4431 & -4267 & 10773 & -10367 \\
\bottomrule[1.5pt]
\end{tabular}
\vskip1ex%
\vskip -0ex%
\vspace{-1em}
\end{table}\
%60 & 0 & 60 & 0 & 166 & -2 & 416 & -10
%62 & -2 & 64 & -4 & 164 & 0 & 406 & 0
%289 & -229 & 215 & -155 & 777 & -613 & 1910 & -1504
%305 & -245 & 222 & -162 & 871 & -707 & 1901 & -1495
%2181 & -2121 & 1980 & -1920 & 6178 & -6014 & 13809 & -13403
%3768 & -3708 & 2528 & -2468 & 9506 & -9342 & 22593 & -22187
%3885 & -3825 & 2187 & -2127 & 11146 & -10982 & 29637 & -29231
%5970 & -5910 & 2967 & -2907 & 25116 & -24952 & 40178 & -39772
%126 & -66 & 91 & -31 & 332 & -168 & 796 & -390
%182 & -122 & 137 & -77 & 587 & -423 & 1342 & -936
%1421 & -1361 & 746 & -686 & 4431 & -4267 & 10773 & -10367

% \begin{algorithm}[!t]
% \footnotesize
% \caption{app:MM Attack}
% \label{alg:Alg__MM}
% \begin{algorithmic}
% \STATE \textbf{Input:} natural data $x$, true label $y$, set of false labels $C$, model $f$, loss funciton $\ell_{MM}$, maximum PGD step $N$, perturbation bound $\epsilon$, initial step size $\alpha$, the number classes $K$, targets selection number $K_s$;
% \STATE \textbf{Output:} adversarial data $\tilde{x}$;
% \STATE$\tilde{x} \gets x$;

% \WHILE{$K_s > 0$} 
% \STATE $N_{Iter} \gets N$;
% \STATE $c = \argmax_{i \in C} {f(x)}_i$;
% \WHILE{$N_{Iter} > 0$} 
% \STATE $\tilde{x} \gets \Pi_{\mathcal{B}_{\epsilon}[x]}(\tilde{x}+\alpha sign(\nabla_{\tilde{x}}\ell_{MM}(f(\tilde{x}),y,c))$;
% \STATE $K_s \gets K_s - 1$;
% \ENDWHILE
% \STATE $C \gets C - \{c\}$;
% \IF{$\argmax_{i \in C} {f(x)}_i \neq y$}
% \STATE $K_s \gets 0$;
% \ENDIF
% \ENDWHILE
% %\vspace{1mm}
% \end{algorithmic}
% \end{algorithm}

\end{document}